\documentclass[10pt,twocolumn,letterpaper]{article}

\usepackage{iccv}
\usepackage{times}
\usepackage{epsfig}
\usepackage{graphicx}
\usepackage{amsmath}
\usepackage{amssymb}

\usepackage{algorithmic}
\usepackage[export]{adjustbox}
\usepackage{rotating}
\usepackage{wrapfig}
\usepackage{float}
\usepackage{array}
\usepackage{subfig}
\usepackage{tabularx}
\usepackage{tikz}
\usetikzlibrary{spy}
\usepackage{enumitem}
\usepackage{colortbl}
\usepackage{tikz}
\newlength{\ww}

\usepackage[pagebackref=true,breaklinks=true,colorlinks,bookmarks=false]{hyperref}

\iccvfinalcopy 

\def\iccvPaperID{28} 

\ificcvfinal\fi

\usepackage{cleveref}

\begin{document}

\title{Blended-NeRF: Zero-Shot Object Generation and Blending in Existing Neural Radiance Fields}

\author{Ori Gordon\\
The Hebrew University\\
{\tt\small ori.gordon@mail.huji.ac.il}
\and
Omri Avrahami\\
The Hebrew University\\
{\tt\small omri.avrahami@mail.huji.ac.il}
\and
Dani Lischinski\\
The Hebrew University\\
{\tt\small danix@mail.huji.ac.il}
}
\twocolumn[{
\renewcommand\twocolumn[1][]{#1}%
\maketitle
\centering
\captionsetup{type=figure}
\setlength{\tabcolsep}{1pt}
\renewcommand{\arraystretch}{0.5}
\setlength{\ww}{0.32\columnwidth}
\vspace{-2.2em}
\begin{tabular}{cc}
    \includegraphics[height=4.7cm]{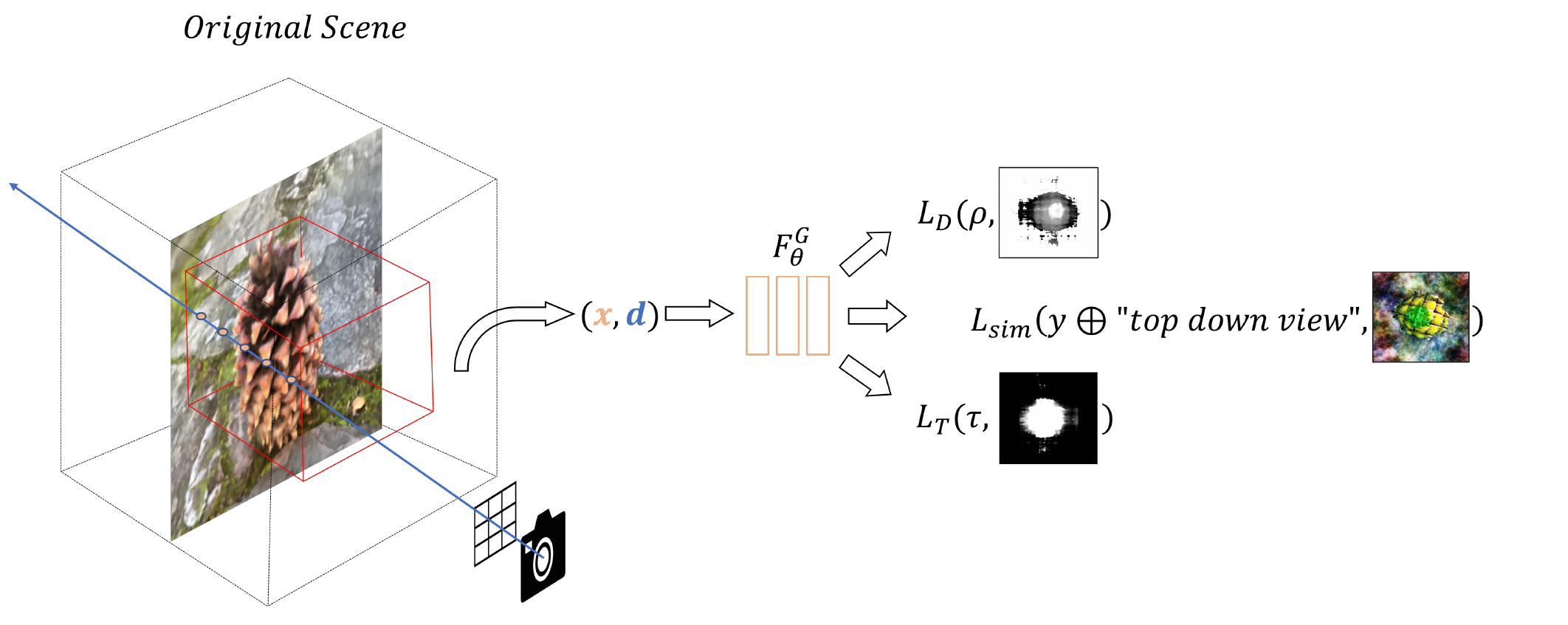} & 
    \includegraphics[height=4.7cm]{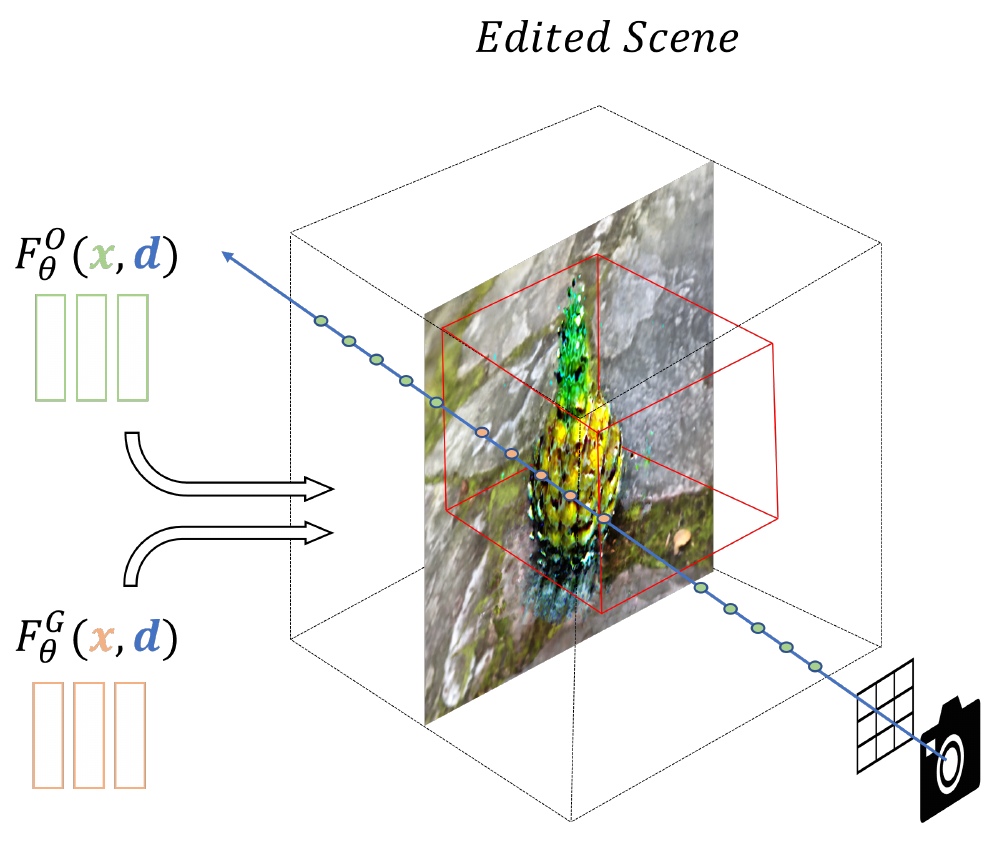} \\
    \begin{tabular}{c}
        (a)
    \end{tabular} & 
    \begin{tabular}{c}
        (b)
    \end{tabular} \\
\end{tabular}
\captionof{figure}{\textbf{Overview.} (a) \textbf{Training:} Given a NeRF scene $F_{\theta}^{O}$, our pipeline trains a NeRF generator model $F_{\theta}^{G}$, initialized with $F_{\theta}^{O}$ weights and guided by a similarity loss defined by a language-image model such as CLIP \cite{CLIP2022}, to synthesize a new object inside a user-specified ROI. This is achieved by casting rays and sampling points for the rendering process \cite{NeRF2020} only inside the ROI box. Our method introduces augmentations and priors to get more natural results. (b) \textbf{Blending process:} After training, we render the edited scene by blending the sample points generated by the two models along each view ray.}
\phantom{.}
\label{fig:teaser}
}]
\ificcvfinal\thispagestyle{empty}\fi

\newcommand{\ignorethis}[1]{}
\newcommand{\redund}[1]{#1}

\newcommand{\apriori    }     {\textit{a~priori}}
\newcommand{\aposteriori}     {\textit{a~posteriori}}
\newcommand{\perse      }     {\textit{per~se}}
\newcommand{\naive      }     {{na\"{\i}ve}}

\newcommand{\Identity   }     {\mat{I}}
\newcommand{\Zero       }     {\mathbf{0}}
\newcommand{\Reals      }     {{\textrm{I\kern-0.18em R}}}
\newcommand{\isdefined  }     {\mbox{\hspace{0.5ex}:=\hspace{0.5ex}}}
\newcommand{\texthalf   }     {\ensuremath{\textstyle\frac{1}{2}}}
\newcommand{\half       }     {\ensuremath{\frac{1}{2}}}
\newcommand{\third      }     {\ensuremath{\frac{1}{3}}}
\newcommand{\fourth     }     {\ensuremath{\frac{1}{4}}}

\newcommand{\Lone} {\ensuremath{L_1}}
\newcommand{\Ltwo} {\ensuremath{L_2}}

\newcommand{\mat        } [1] {{\text{\boldmath $\mathbit{#1}$}}}
\newcommand{\Approx     } [1] {\widetilde{#1}}
\newcommand{\change     } [1] {\mbox{{\footnotesize $\Delta$} \kern-3pt}#1}

\newcommand{\Order      } [1] {O(#1)}
\newcommand{\set        } [1] {{\lbrace #1 \rbrace}}
\newcommand{\floor      } [1] {{\lfloor #1 \rfloor}}
\newcommand{\ceil       } [1] {{\lceil  #1 \rceil }}
\newcommand{\inverse    } [1] {{#1}^{-1}}
\newcommand{\transpose  } [1] {{#1}^\mathrm{T}}
\newcommand{\invtransp  } [1] {{#1}^{-\mathrm{T}}}
\newcommand{\relu       } [1] {{\lbrack #1 \rbrack_+}}

\newcommand{\abs        } [1] {{| #1 |}}
\newcommand{\Abs        } [1] {{\left| #1 \right|}}
\newcommand{\norm       } [1] {{\| #1 \|}}
\newcommand{\Norm       } [1] {{\left\| #1 \right\|}}
\newcommand{\pnorm      } [2] {\norm{#1}_{#2}}
\newcommand{\Pnorm      } [2] {\Norm{#1}_{#2}}
\newcommand{\inner      } [2] {{\langle {#1} \, | \, {#2} \rangle}}
\newcommand{\Inner      } [2] {{\left\langle \begin{array}{@{}c|c@{}}
                               \displaystyle {#1} & \displaystyle {#2}
                               \end{array} \right\rangle}}

\newcommand{\twopartdef}[4]
{
  \left\{
  \begin{array}{ll}
    #1 & \mbox{if } #2 \\
    #3 & \mbox{if } #4
  \end{array}
  \right.
}

\newcommand{\fourpartdef}[8]
{
  \left\{
  \begin{array}{ll}
    #1 & \mbox{if } #2 \\
    #3 & \mbox{if } #4 \\
    #5 & \mbox{if } #6 \\
    #7 & \mbox{if } #8
  \end{array}
  \right.
}

\newcommand{\len}[1]{\text{len}(#1)}

\newlength{\w}
\newlength{\h}
\newlength{\x}

\definecolor{darkred}{rgb}{0.7,0.1,0.1}
\definecolor{darkgreen}{rgb}{0.1,0.6,0.1}
\definecolor{cyan}{rgb}{0.7,0.0,0.7}
\definecolor{otherblue}{rgb}{0.1,0.4,0.8}
\definecolor{maroon}{rgb}{0.76,.13,.28}
\definecolor{burntorange}{rgb}{0.81,.33,0}

\ifdefined\ShowNotes
  \newcommand{\colornote}[3]{{\color{#1}\textbf{#2} #3\normalfont}}
\else
  \newcommand{\colornote}[3]{}
\fi

\newcommand {\todo}[1]{\colornote{cyan}{TODO}{#1}}
\newcommand {\dlc}[1]{\colornote{darkred}{Dani:}{#1}}
\newcommand {\ori}[1]{\colornote{burntorange}{Ori:}{#1}}
\newcommand {\omri}[1]{\colornote{darkgreen}{Omri:}{#1}}

\newcommand {\reqs}[1]{\colornote{red}{\tiny #1}}

\newcommand {\new}[1]{{\color{red}{#1}}}

\newcommand*\rot[1]{\rotatebox{90}{#1}}

\newcommand {\newstuff}[1]{#1}

\newcommand\todosilent[1]{}

\newcommand{\woBGmask}{{w/o~bg~\&~mask}}
\newcommand{\woMask}{{w/o~mask}}

\providecommand{\keywords}[1]
{
  \textbf{\textit{Keywords---}} #1
}




\begin{abstract}
Editing a local region or a specific object in a 3D scene represented by a NeRF or consistently blending a new realistic object into the scene is challenging, mainly due to the implicit nature of the scene representation.
We present Blended-NeRF, a robust and flexible framework for editing a specific region of interest in an existing NeRF scene, based on text prompts, along with a 3D ROI box.
Our method leverages a pretrained language-image model to steer the synthesis towards a user-provided text prompt, along with a 3D MLP model initialized on an existing NeRF scene to generate the object and blend it into a specified region in the original scene. 
We allow local editing by localizing a 3D ROI box in the input scene, and blend the content synthesized inside the ROI with the existing scene using a novel volumetric blending technique. To obtain natural looking and view-consistent results, we leverage existing and new geometric priors and 3D augmentations for improving the visual fidelity of the final result.
We test our framework both qualitatively and quantitatively on a variety of real 3D scenes and text prompts, demonstrating realistic multi-view consistent results with much flexibility and diversity compared to the baselines. Finally, we show the applicability of our framework for several 3D editing applications, including adding new objects to a scene, removing/replacing/altering existing objects, and texture conversion.
\footnote{Project page: \href{https://www.vision.huji.ac.il/blended-nerf/}{\textit{www.vision.huji.ac.il/blended-nerf}}} 
\end{abstract}

    \begin{figure*}[ht]
    \centering
    \setlength{\tabcolsep}{0.5pt}
    \renewcommand{\arraystretch}{0.5}
    \setlength{\ww}{0.5\columnwidth}
  
    \begin{tabular}{cccc}        
        \includegraphics[width=\ww,frame]{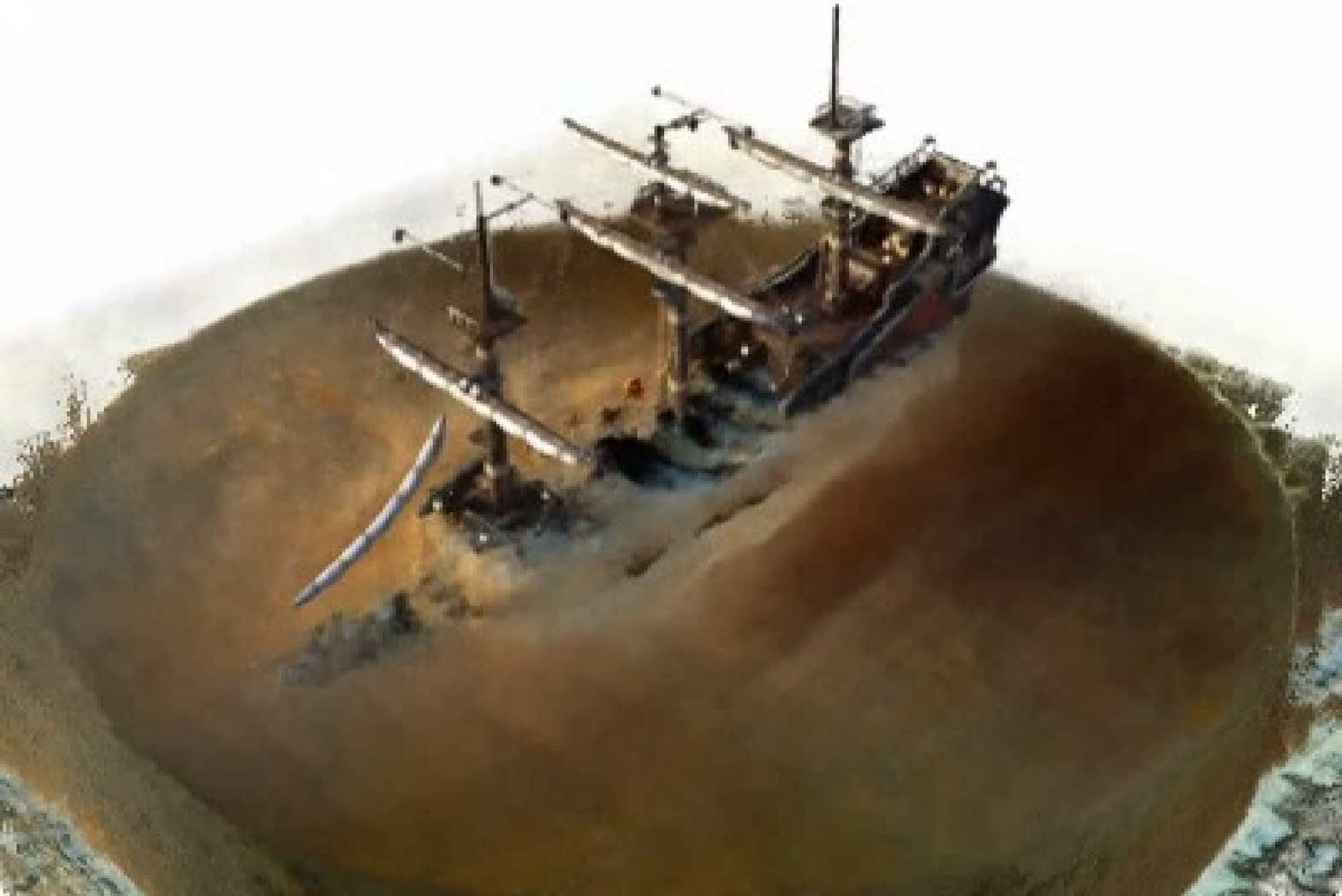}&
        \includegraphics[width=\ww,frame]{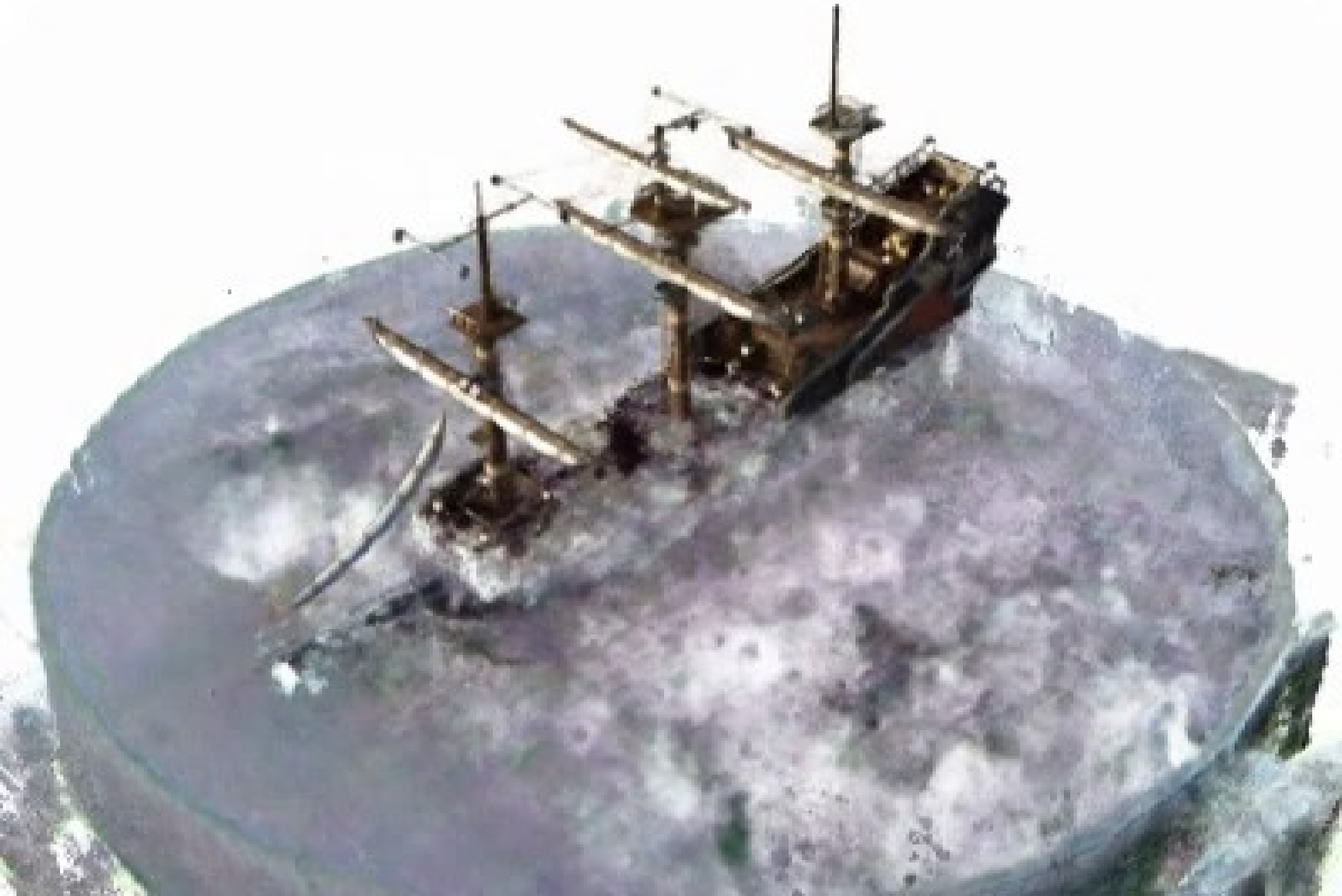}
        \hspace{0.3cm}\makebox[0pt][b]{
        \includegraphics[width=0.22\columnwidth,frame]{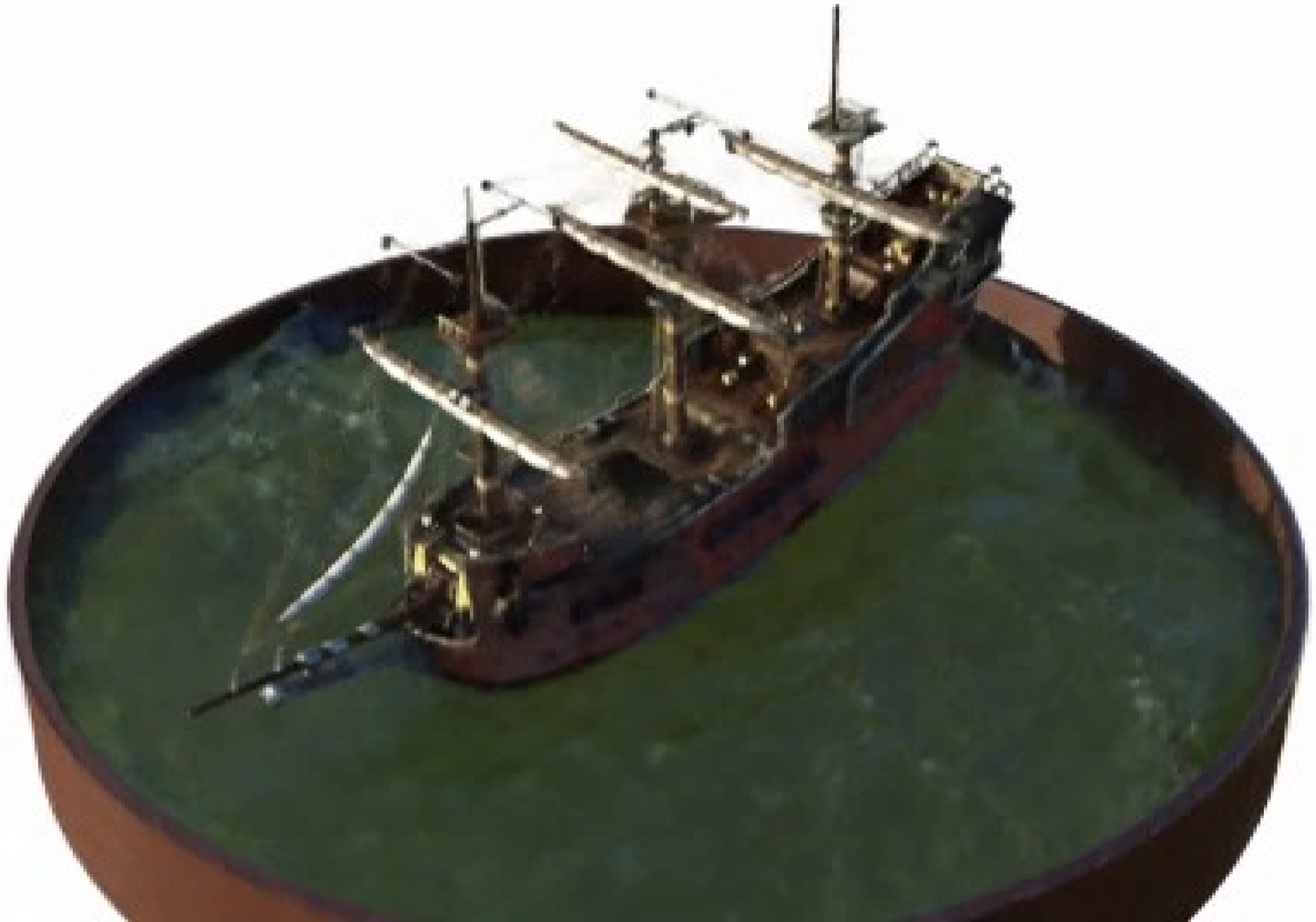}} &
        
        \includegraphics[width=\ww,frame]{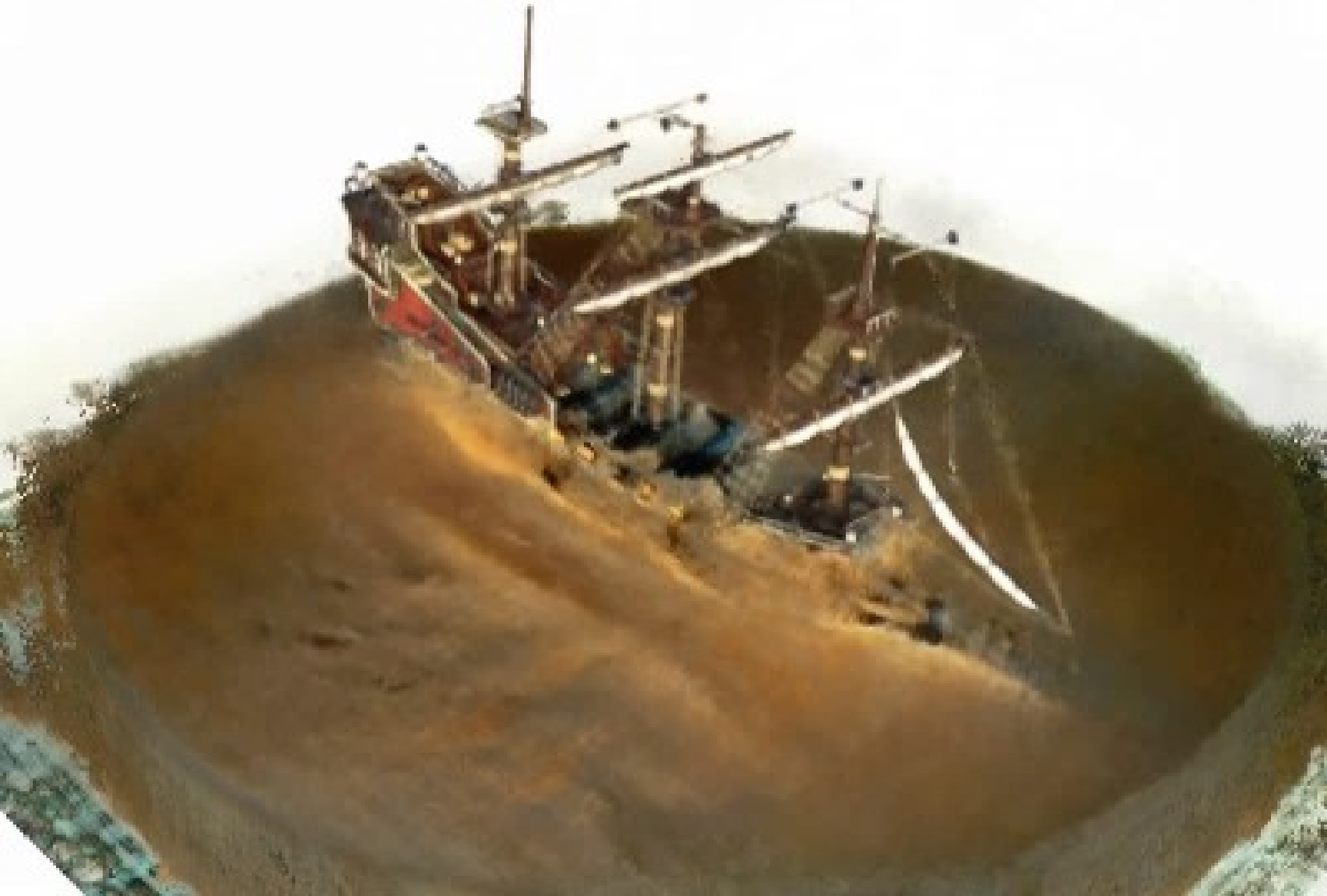}&
        \includegraphics[width=\ww,frame]{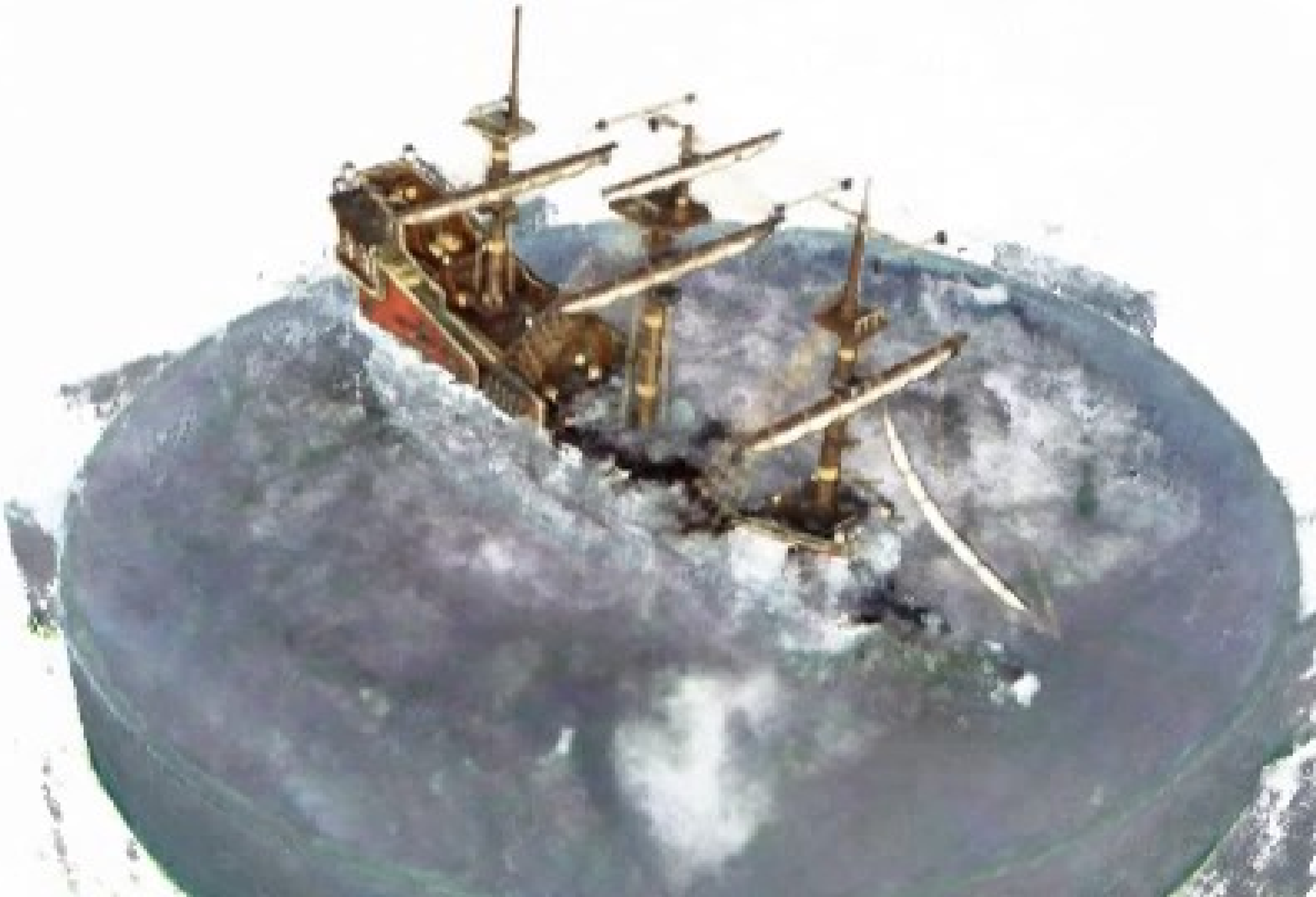}         
        \hspace{0.3cm}\makebox[0pt][b]{
        \includegraphics[width=0.22\columnwidth,frame]{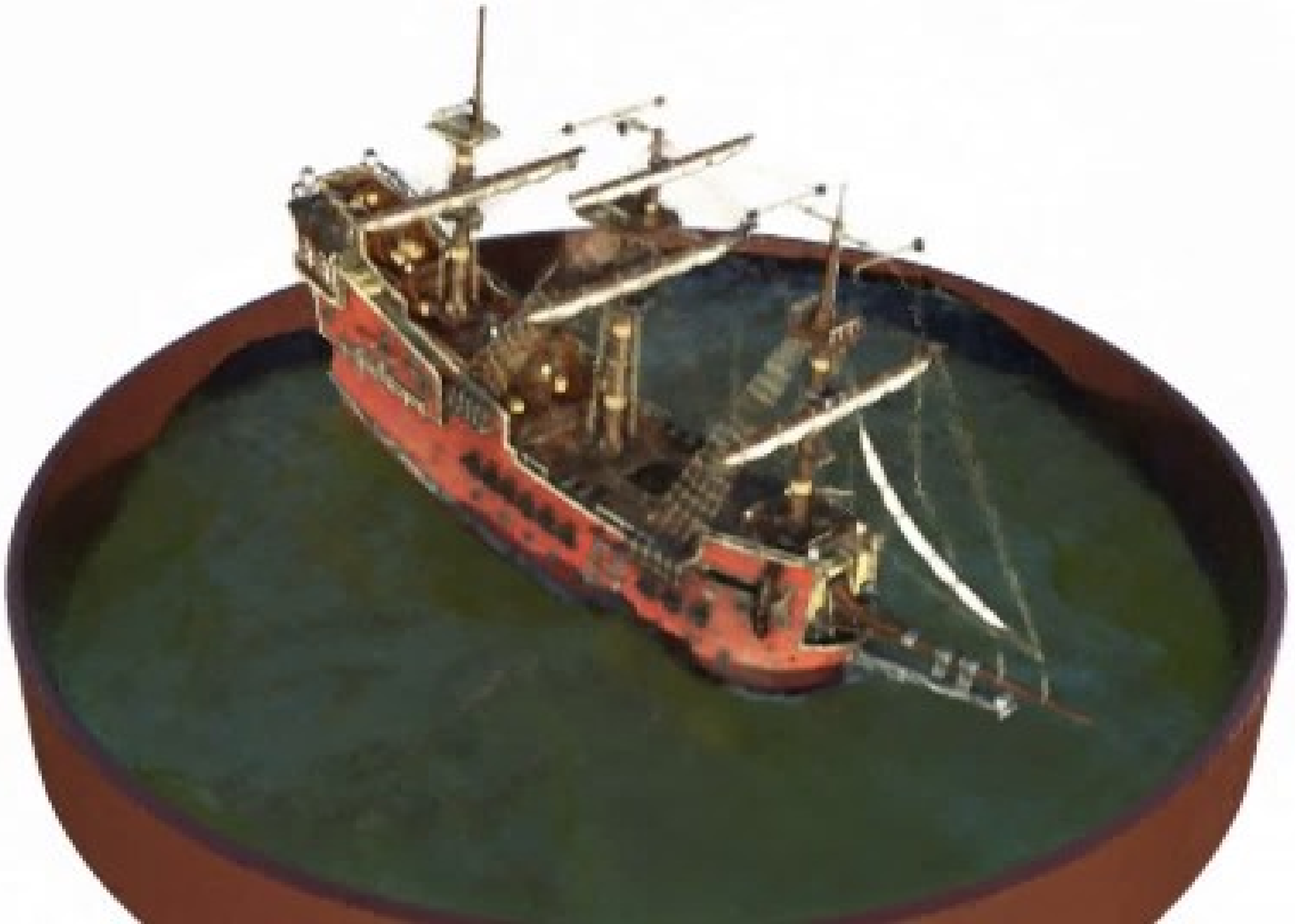}}\\
		
	\scriptsize{"A DLSR photo of dunes of sand."} & 
	\scriptsize{"A DLSR photo of ice and snow."} &
 	\scriptsize{"A DLSR photo of dunes of sand."} & 
	\scriptsize{"A DLSR photo of ice and snow."} \\
    \end{tabular}
    
    \caption{\textbf{Large object replacement.} Here we preform object replacement to the blender ship scene by localizing the ROI box to include the sea and the bottom of the ship and training our model to steer the edit towards the given text prompts.} \label{fig:blender_ship}
\end{figure*}

\section{Introduction}
\label{sec:intro}
In the last few years we have witnessed exciting developments in neural implicit representations \cite{sitzmann2019siren, strumpler2022implicit, dupont2021coin, NEURIPS2022_35d5ad98, liu2022semantic, szatkowski2022hypersound}.
In particular, implicit representations of 3D scenes \cite{sitzmann2019srns,mescheder2019occupancy, sitzmann2019metasdf, jiang2020local, park2019deepsdf,NeRF2020,barron2022mip360,barron2021mip} have enabled unprecedented quality and reliability in 3D reconstruction and novel view synthesis. The pioneering work of Mildenhall \etal~\cite{NeRF2020} introduced NeRFs, MLP-based neural models that implicitly represent a scene as a continuous volume and radiance fields from a limited number of observations, producing high-quality images from novel views via volume rendering.

However, editing a scene represented by a NeRF is non-trivial, mainly because the scene is encoded in an implicit manner by the model's weights, in contrast to explicit representations, such as meshes, voxel grids, or point clouds. NeRFs offer no explicit separation between the various components that define the object, such as shape, color, or material.
In contrast to local edits in images, \eg, \cite{avrahami2022blended, Avrahami2022BlendedLD, bau2021paint, Nichol2021GLIDETP, Ramesh2022HierarchicalTI, ho2020denoising,brooks2022instructpix2pix}, where the edit is done in pixel space with all the required information appearing in a single view, editing a NeRF-represented scene is more challenging due to the requirement for consistency across multiple views between the new and the original NeRF scenes.

The first works attempting to edit NeRF scenes focused on the removal of local parts, changing color, or shape transfer on one class of synthetic data, guided by user scribbles or latent code of another object in the class \cite{EditNeRF2021}. In CLIP-NeRF \cite{CLIPNeRF2021}, editing of the entire scene is preformed by text guidance and displacements to the latent representation of the input. They mainly focus on synthetic objects from one class, or global color changes for realistic scenes. Kobayashi \etal~\cite{DecomposeNeRF2022} perform semantic decomposition of the scene components by learning a feature field that maps each 3D coordinate to a descriptor representing a semantic feature, and allow zero-shot segmentation for local editing on a specific semantic class. Alternatively, Benaim \etal~\cite{VolumeDisentanglement2022} suggest separating the volumetric representation of a foreground object from its background using a set of 2D masks per training view. These works have limited localization abilities and focus on the separation methods. They demonstrate manipulations such as object removal, color change, and transformations such as shift, rotation, and scale.

In this work, we present our approach for ROI-based editing of NeRF scenes guided by a text prompt or an image patch that: (1) can operate on any region of a real-world scene, (2) modifies only the region of interest, while preserving the rest of the scene without learning a new feature space or requiring a set of two-dimensional masks, (3) generates natural-looking and view-consistent results that blend with the existing scene, (4) is not restricted to a specific class or domain, and (5) enables complex text guided manipulations such as object insertion/replacement, objects blending and texture conversion.

To this end, we utilize a pretrained language-image model, \eg, CLIP \cite{CLIP2022}, and a NeRF model \cite{NeRF2020} initialized on existing NeRF scene as our generator for synthesizing a new object and blend it into the scene in the region of interest (ROI). We use CLIP to steer the generation process towards the user-provided text prompt, enabling blended generation of diverse 3D objects.

To enable general local edits in any region, while preserving the rest of the scene, we localize a 3D box inside a given NeRF scene.
To blend the synthesized content inside the ROI with the base scene, we propose a novel volumetric blending approach that merges the original and the synthesized radiance fields by blending the sampled 3D points along each camera ray. 

We show that using this pipeline naively to perform the edit is insufficient, generating low quality incoherent and inconsistent results. Thus, we utilize the augmentations and priors suggested in \cite{DreamFields2022} and introduce additional priors and augmentations, such as depth regularization, pose sampling, and directional dependent prompts to get more realistic, natural-looking and 3D consistent results. Finally, we conduct extensive experiments to evaluate our framework and the effect of our additional constraints and priors. We perform an in-depth comparison with the baseline and show the applicability of our approach on a series of 3D editing applications using a variety of real 3D scenes.

\section{Related Work}
\label{sec:related_work}
\textbf{Neural Implicit Representations} have gained much popularity in the fields of computer vision and graphics in both 2D and 3D \cite{sitzmann2019siren,sitzmann2019srns,sitzmann2019metasdf,park2019deepsdf,mescheder2019occupancy,strumpler2022implicit,dupont2021coin,jiang2020local}. Among their advantages is their ability to capture complex and diverse patterns and to provide a continuous representation of the underlying scene. They are resolution independent, yet compact, compared to explicit representations of high resolution 2D images, or meshes and point clouds in 3D. NeRFs \cite{NeRF2020,barron2021mip,barron2022mip360} learn to represent a 3D scene as a continuous volume and radiance fields using the weights of a multilayer perceptron (MLP). Given a 3D position $x$ and view direction $(\theta,\phi)$, NeRF outputs the density $\sigma$ and color $c$ at $x$. Novel views of the scene can thus be rendered by accumulating the colors and densities along a view ray $\boldsymbol{r}(t)$ passing through each pixel, using an approximation to the classical volume rendering equation using the quadrature rule \cite{max1995optical}:
\begin{equation} \label{quad_eqn}
 C(\boldsymbol{r}) = \sum_{i=1}^{N}T_i(1-\exp(-\sigma_i\delta_i))c_i,\, T_i = \exp(-\sum_{j=1}^{i-1}\sigma_j\delta_j)
\end{equation}
where $\delta_i=t_{i+1}-t_i$ is the distance between adjacent samples and $T_i$ can be interpreted as the degree of transmittance at point $x_i$ along the ray.
The inputs are embedded into a high-dimensional space using a high frequency sinusoidal positional encoding $\gamma(x)$ to enable better fitting for high frequency variations in the data~\cite{SpectralBias2019,FourierHighFrequencyFunctions2020}:
\begin{equation} \label{pos_enc_eqn}
 \gamma(x) = [\cos(2^{l}x),\, \sin(2^{l}x)]_{l=0}^{L-1}
\end{equation}

\textbf{NeRF 3D Generation.} NeRFs inspired follow-up works to synthesize new NeRF objects from scratch. The first methods used NeRF combined with GANs~\cite{WassersteinGenerative2017,GenerativeAdversarial2014, ImprovedTrainingGANs2017} to design 3D-aware generators \cite{gu2022stylenerf, Pi-GAN2021, UnconstrainedSceneGeneration2021, CAMPARI2021, GIRAFFE2021CVPR, GRAF2020NEURIPS, zhou2021cips3d}. GRAF \cite{GRAF2020NEURIPS} adopts shape and appearance codes to conditionally synthesize NeRF and GIRAFF \cite{GIRAFFE2021CVPR}, StyleNeRF \cite{gu2022stylenerf} utilizes NeRF to render features instead of RGB colors and adopt a two-stage strategy, where they render low-resolution feature maps first and then up-sample the feature maps using a CNN decoder. These models are category-specific and  trained mostly on forward-facing scenes.

More recent works utilize the progress in contrastive representation learning \cite{VirTex_2021_CVPR, CLIP2022, LiT_2022_CVPR, li2022blip, li2023blip2}, which enables easy and flexible control over the content of the generated objects using textual input. In Dream Fields \cite{DreamFields2022}, frozen image-text joint embedding models from CLIP \cite{CLIP2022} are used as a guidance to a NeRF model that generates 3D objects whose renderings have high semantic similarity with the input caption. To improve the visual quality, they introduce geometric priors and augmentations to enforce transmittance sparsity, object boundaries and multi-view consistency. In this paper, we utilize some of the priors from Dream Fields \cite{DreamFields2022} and introduce improved augmentations and priors to edit existing NeRF scenes.

More recent works utilize the progress in diffusion models \cite{DenoisingDiffusion_NEURIPS2020, DenoisingImplicitModels2021_ICML, ScoreBasedNEURIPS2020} and specifically in text-conditioned diffusion models \cite{Ramesh2022HierarchicalTI, HighResolutionDiffusion_2022_CVPR, saharia2022photorealistic}. DreamFusion \cite{poole2023dreamfusion} and its follow-ups \cite{ScoreJacobianChaining2022, metzer2022latent, lin2023magic3d, raj2023dreambooth3d} optimize a NeRF model by replacing CLIP with score function losses using pretrained text-conditioned 2D diffusion-models applied on many different views of the generated scene to synthesize 3D objects aligned with the input text. These models synthesize new objects without considering how they can be inserted and blend into an existing scene.

\textbf{Editing NeRFs.} The pioneering works \cite{EditNeRF2021, CLIPNeRF2021} were the first to tackle the challenge of editing NeRF scenes. They both define a conditional NeRF, where the NeRF model is conditioned on latent shape and appearance codes, which enables separately editing the shape and the color of a 3D object. EditNeRF \cite{EditNeRF2021} only enables addition and removal of local parts or color changes guided by user scribbles and is limited to only one shape category. In ObjectNeRF \cite{yang2021objectnerf} they enable editing tasks such as moving or adding new objects to the scene by introducing a neural scene rendering system with a scene branch which encodes the scene geometry and appearance and object branch which encodes each standalone object.
CLIP-NeRF \cite{CLIPNeRF2021} leverage the joint language-image embedding space of CLIP \cite{CLIP2022} to perform text or image guided manipulation on the entire scene. During the optimization it uses two code mappers for the shape and appearance that receive the CLIP embedding and output shape and appearance codes which steer the input of the model and the model weights to apply the edit. The manipulation capabilities are demonstrated mainly on synthetic objects from one class and on global color changes for realistic scenes.

Later works focused on geometric edits \cite{yuan2022nerf-editing}, global style transfer \cite{chen2022upst, chiang2022stylizing, fan2022unified, huang2022stylizednerf}, recoloring \cite{wu2022palettenerf, gong2023recolornerf}, and disentanglement of the scene to enable local edits \cite{DecomposeNeRF2022, VolumeDisentanglement2022, zhang2022Nerflets}. Kobayashi \cite{DecomposeNeRF2022} decomposes the scene to its semantic parts by training the NeRF model to learn a 3D feature field using supervision of pre-trained 2D image feature extractors \cite{DINO2021ICCV, lSeg2022} in addition to learning of the volume density and the radiance field. After training, the model can perform zero-shot segmentation for local editing of a specific semantic class.
Benaim \etal~\cite{VolumeDisentanglement2022} disentangle the volumetric representation of a foreground object from its background using a set of 2D masks specifying the foreground object in each training view. They train two models for the full scene and the background scene, and subtract the background from the full scene in order to get the foreground. In both works the localization on the region of interest is incomplete and not flexible enough (does not enable editing parts of objects, empty regions or blending new densities into the area of existing object). They demonstrate manipulations such as object removal, transformations such as shift rotation and scale, and only basic optimization-based edits. Our work focuses on blending text generated objects with volume and color into any region of interest of an existing scene with more freedom and flexibility and without compromising on quality and visibility. For information regrading concurrent works, please refer to the supplement.

\begin{figure*}[h]
    \centering
    \setlength{\tabcolsep}{0.5pt}
    \renewcommand{\arraystretch}{0.5}
    \setlength{\ww}{0.38\columnwidth}
  
    \begin{tabular}{ccccc}        
        \includegraphics[width=\ww,frame]{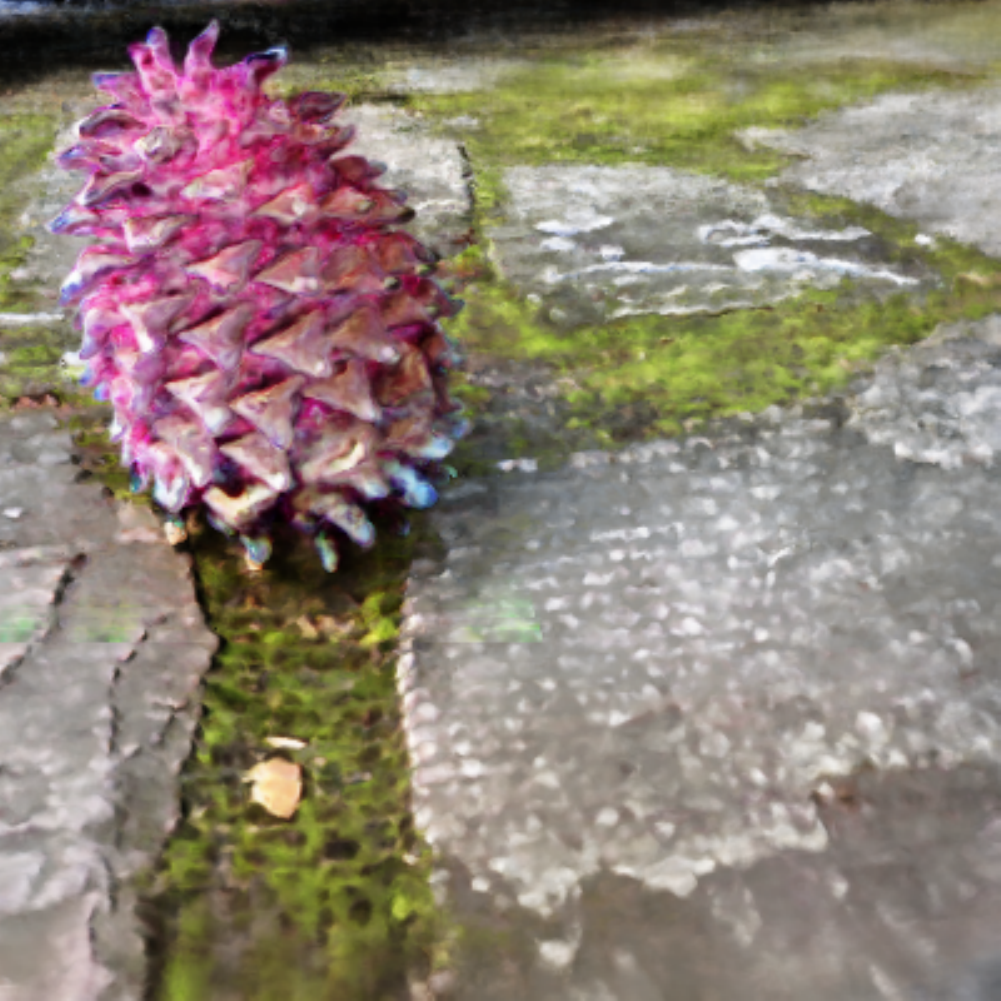} &
        \includegraphics[width=\ww,frame]{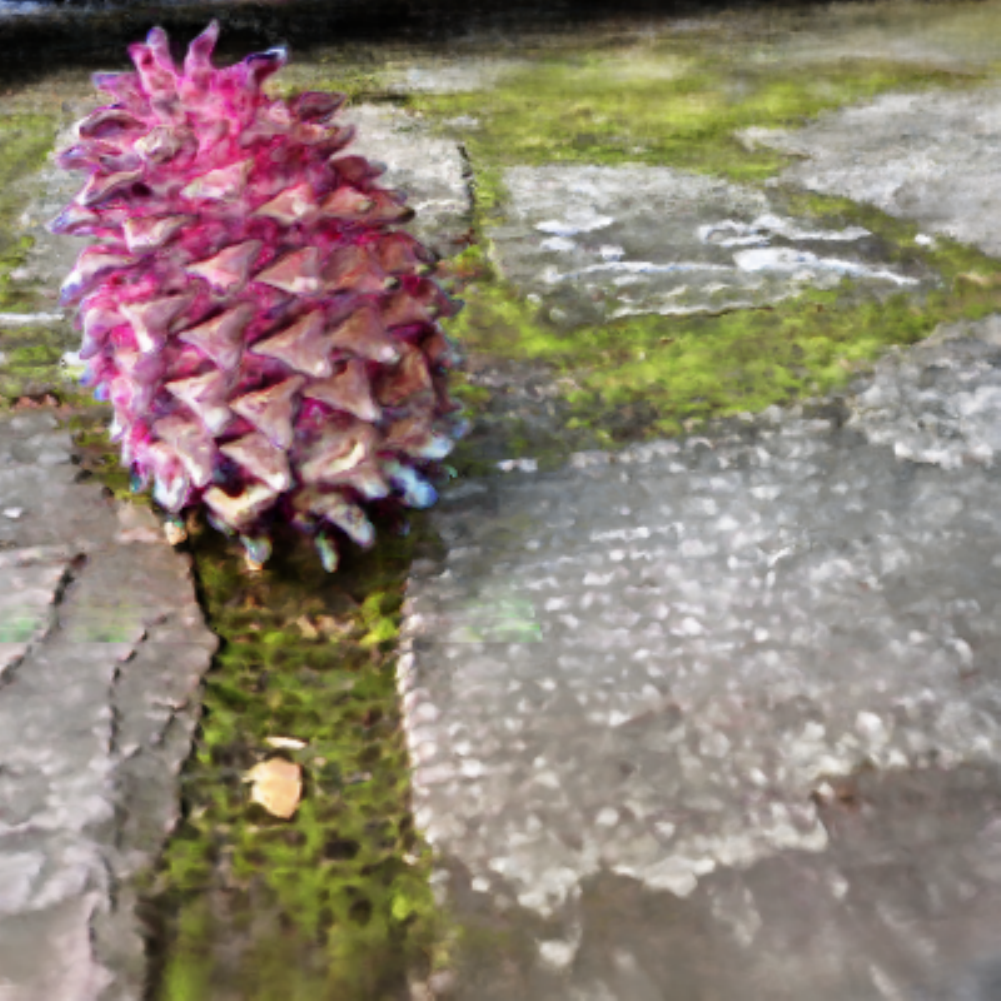} &
        \includegraphics[width=\ww,frame]{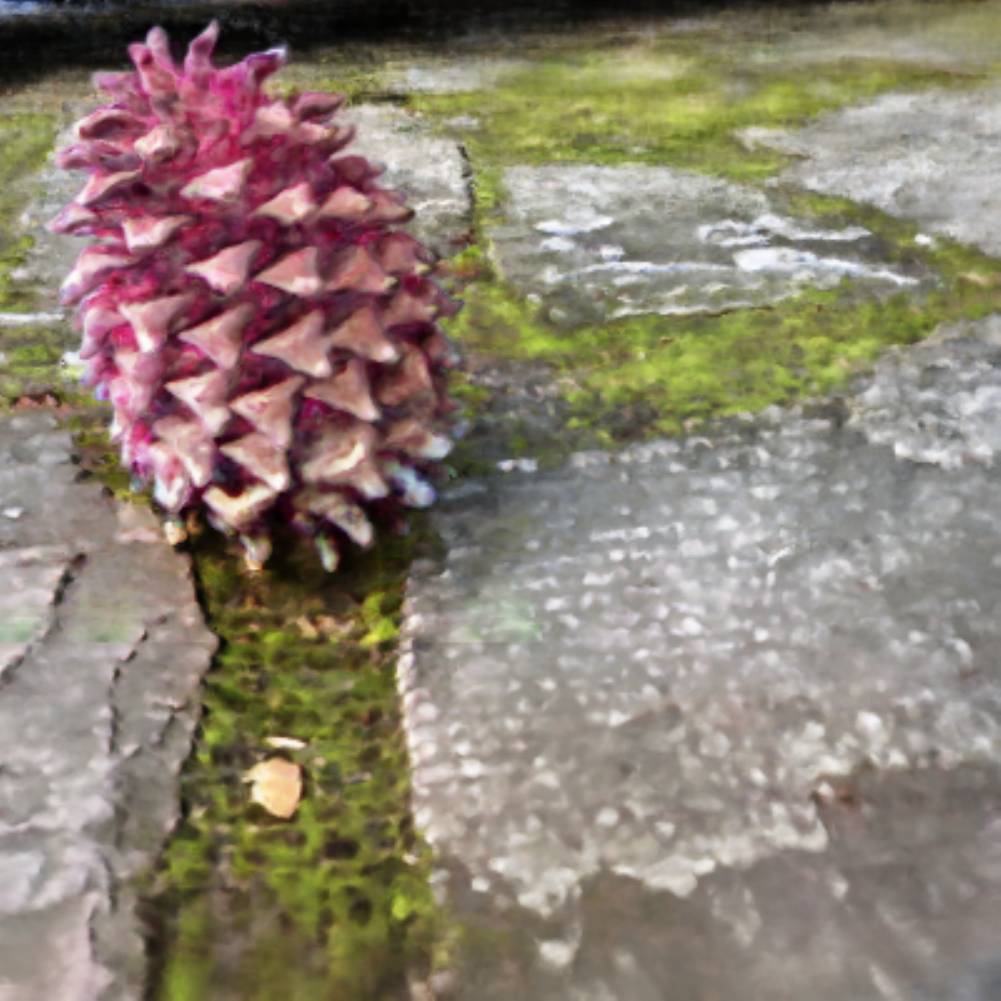} &
        \includegraphics[width=\ww,frame]{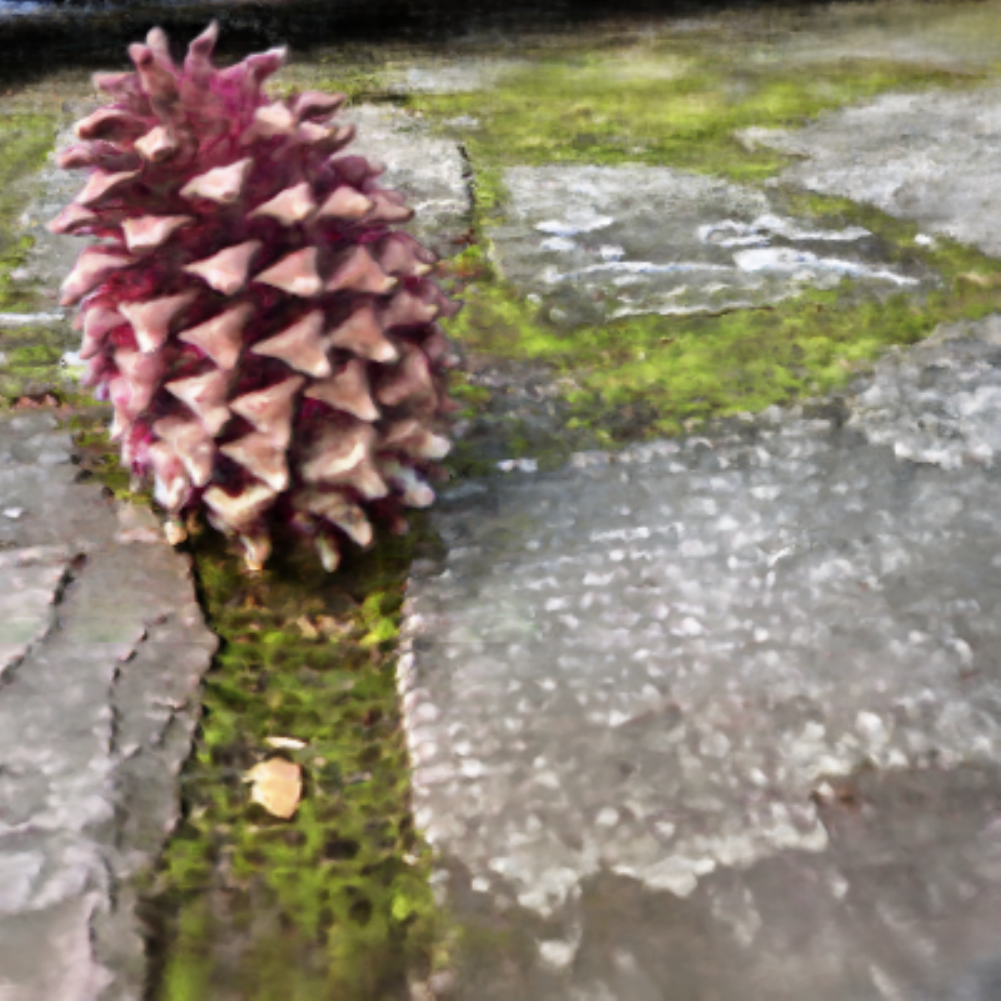} &
        \includegraphics[width=\ww,frame]{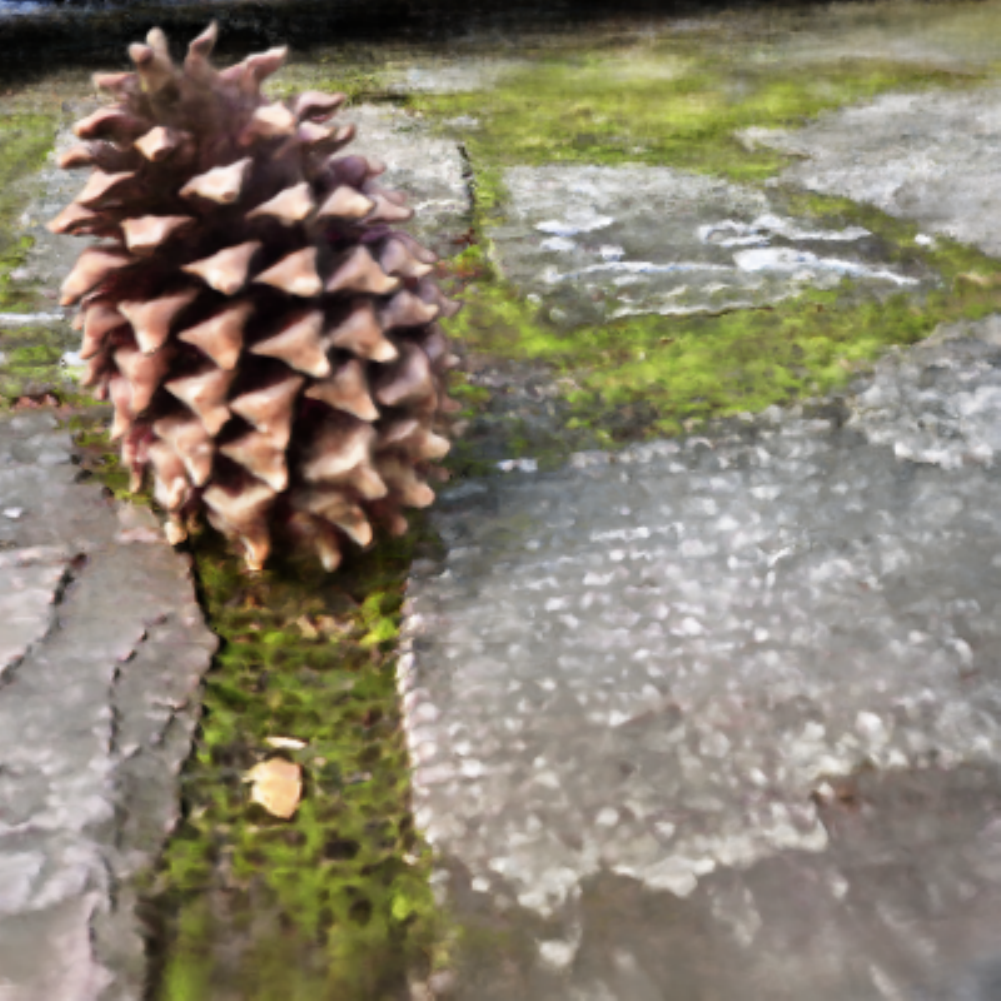} \\
		
		\scriptsize{$\alpha=0$} & 
		\scriptsize{$\alpha=0.5$} & 
		\scriptsize{$\alpha=2$} &
            \scriptsize{$\alpha=4$} &
            \scriptsize{$\alpha=10$}\\
    \end{tabular}
    
    \caption{\textbf{Distance Smoothing Operator.} We demonstrate our suggested smoothing operator in \cref{distance_blending} on a range of $\alpha$ values, When $\alpha$ is zero all the weight goes to the edited scene, and as we increase $\alpha$, more attention is given to closer points from the original scene.}\label{fig:alpha_range}
\end{figure*}

\newcommand{\imgenc}{E_{\textit{img}}}
\newcommand{\txtenc}{E_{\textit{txt}}}

\section{Method}
\label{sec:method}

Given an existing 3D scene $x_o$ represented by a NeRF model $F_{\theta}^{O}$, and a 3D region of interest (ROI), indicated by a box $B$ localized inside the scene, our goal is to modify the scene inside the ROI, according to a user-provided text prompt.
In other words, we aim to obtain a modified scene $x_e$, where $x_e \odot B$ is consistent with the user prompt from any point of view, while matching $x_o$ outside the box ($x_e \odot (1-B) = x_o \odot (1-B)$).

To preform the edits inside the ROI we initialize a 3D MLP model $F_{\theta}^{G}$ with the weights of the original scene model $F_{\theta}^{O}$ and steer the weights towards the given prompt using a pretrained language-image model, such as CLIP \cite{CLIP2022}.
We enable local edits in any region of the scene $x_o$ using a simple GUI for localizing a 3D box inside the scene by rendering the original NeRF model $F_{\theta}^{O}$ from any view and using the output depth map of the model to obtain 3D understanding of the scene. Using the given ROI box we can disentangle the scene inside the box and outside it by decomposing the radiance fields accordingly. To obtain consistent results from any view direction, we perform volumetric blending of the original and the edited radiance fields by sampling 3D points along each camera ray $\boldsymbol{r}$ in both $F_{\theta}^{O}$ and $F_{\theta}^{G}$, and blending the samples while accounting for their densities, colors and distance from the center of the scene.

To get more realistic and natural-looking results we present existing \cite{DreamFields2022} and novel augmentations and priors such as transmittance and depth regularization, background augmentations, pose sampling and directional dependent prompts. An overview of our approach is depicted in \Cref{fig:teaser}.

In \Cref{sec:blending} we describe our 3D object generation and blending process, we continue and present the model objectives and proposed priors in \Cref{sec:objectives}.

\subsection{Image-Text driven 3D synthesis and blending}
\label{sec:blending}

Given a 3D scene represented by a NeRF model $F_{\theta}^{O}$, an ROI box $B$, and a camera pose, we use a duplicate of $F_{\theta}^{O}$, $F_{\theta}^{G}$ as our starting point for generating the content of $B$. The rest of the scene is preserved by rendering only the rays which have sample points inside $B$. The training of $F_{\theta}^{G}$ is guided by a language-image model, \eg, \cite{CLIP2022,li2022blip,li2023blip2,LiT_2022_CVPR} to align the content generated inside $B$ with a user-provided text prompt.

To get a smoothly blended result, we query both models $F_{\theta}^{O}, F_{\theta}^{G}$ using the same set of rays. For sample points outside the ROI, we use the density and color inferred by $F_{\theta}^{O}$, while for points inside the ROI, we blend the results of the two radiance fields using one of two modes, depending on the type of the edit: adding a new object in empty space, or completely replacing an existing one, vs.~adding an object in a non-empty area.

$F_{\theta}^{G}$ is optimized using guidance from a language-image model, such as CLIP \cite{CLIP2022}, by aiming to minimize the cosine similarity score between the user-provided text prompt $y$ and rendered views of the generated content inside the ROI box, $I_{ROI}$:
\begin{equation} \label{sim_eqn}
L_{sim} = -\imgenc(I_{ROI})^T \txtenc(y),
\end{equation}
where $\imgenc$, $\txtenc$ are the image and text encoders of the image-language model.
During optimization, we render $I_{ROI}$ using only the 3D sample points contained inside $B$ by sampling only along rays $\boldsymbol{r}$ that pass through the box and setting the density to zero for all sample points outside $B$, according to \cref{quad_eqn}: 
\begin{equation} \label{density_eqn}
C(\boldsymbol{r})=
        \begin{cases}
        \sum_{x_i\in B}T_i(1-e^{-\sigma_i\delta_i})c_i,\exists x_i\in \boldsymbol{r}\; s.t. \; x_i\in B \\
        0\;\;\;\;\;\;\;\;\;\;\;\;\;\;\;\;\;\;\;\;\;\;\;\;\;\;\;\;\;\;\;\;\;\;,\text{otherwise}
		\end{cases}
\end{equation}

 After training, we blend the scenes inside and outside the ROI with the same set of rays by querying both $F_{\theta}^{O}$ and $F_{\theta}^{G}$ where the points inside the box are rendered by $F_{\theta}^{G}$ and the points outside the box are rendered by $F_{\theta}^{O}$. To get smooth blending between the two scenes we suggest distance smoothing operator per point inside the box considering its distance from the center of the ROI scene (center of mass, computed during training) and alpha compositing the density and color of the two scenes inside the ROI as follows: 
\begin{align}\label{distance_blending}
         f(\textbf{x}) & =1- \exp(\frac{-\alpha d(\textbf{x})}{\textit{diag}}) \\
         \sigma_{\textit{blend}}(\textbf{x}) & = f(\textbf{x}) \cdot \sigma_{O}(\textbf{x}) + (1-f(\textbf{x}))\cdot \sigma_{G}(\textbf{\textbf{x}}) \nonumber \\ 
    c_{\textit{blend}}(\textbf{x}) & = f(\textbf{x}) \cdot c_{O}(\textbf{x}) + (1-f(\textbf{x}))\cdot c_{G}(\textbf{\textbf{x}}) \nonumber
\end{align}
 where $\sigma_O$ and $\sigma_G$ are the densities returned by each model, $d(\textbf{x})$ is the Euclidean distance of a point $\textbf{x}$ inside the ROI from the center of the scene, $\textit{diag}$ is the box diagonal and $\alpha$ is a hyperparameter which controls the strength of the blending, as can be seen intuitively in Figure~\ref{fig:alpha_range}.
The resulted raw densities and RGB values inside and outside the ROI are then blended along each ray using \cref{quad_eqn} to get the current rendered view of the edited scene $x_e$.

\textbf{ Object Insertion/Replacement}. In this mode, a new synthetic object is added into an empty region of the scene, or entirely replaces another existing object inside the ROI. In this mode, we use the pipeline described above, when inside the ROI we consider only the radiance field of $F_{\theta}^{G}$ during training. After training, we blend the two scenes as described above.

\textbf{Object Blending}. In contrast to the above mode, here we aim to blend the new content with the existing scene inside the ROI. We query both the original $F_{\theta}^{O}$ and the edited $F_{\theta}^{G}$ fields inside the box and blend the resulting colors and densities at each ray sample. To blend the sample colors, we first compute the alpha values for each point $x_i$ on the ray separately from each model:
\begin{equation} \label{alpha_eqn}
\begin{split}
    \alpha_O(x_i) = 1-\exp(\phi(\sigma_O(x_i))\cdot \delta_i) \\
    \alpha_G(x_i) = 1-\exp(\phi(\sigma_G(x_i))\cdot \delta_i)
\end{split}
\end{equation}
where $\phi$ is the activation function enforcing that these density values are non-negative. To blend the colors $c_O$ and $c_G$ obtained from the two models, we use the above alpha values, followed by a sigmoid function:
\begin{equation} \label{rgb_eqn}
    c(x_i)= S(\frac{c_O(x_i)\cdot\alpha_O(x_i) + c_G(x_i)\cdot\alpha_G(x_i)}{\epsilon + \alpha_O(x_i) + \alpha_G(x_i)})
\end{equation}
where $\epsilon$ is a small constant, for numerical stability and $S$ is the sigmoid function.

\begin{figure}[hb]
    \centering
    \setlength{\tabcolsep}{0.5pt}
    \renewcommand{\arraystretch}{0.5}
    \setlength{\ww}{0.3\columnwidth}
  
    \begin{tabular}{ccc}
        \includegraphics[width=\ww,frame]{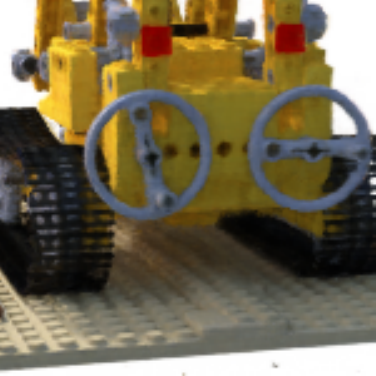} &
        \includegraphics[width=\ww,frame]{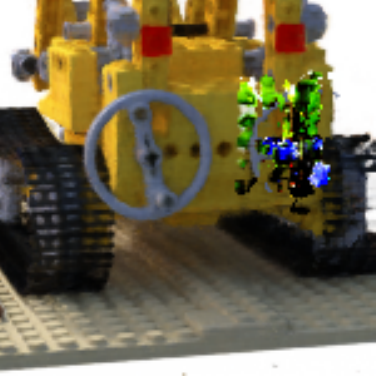} &
        \includegraphics[width=\ww,frame]{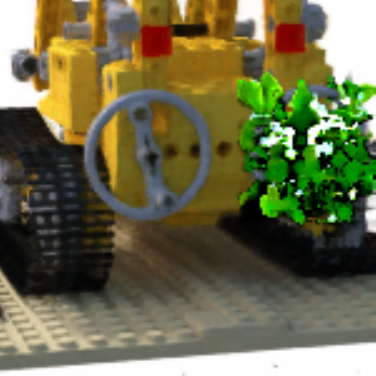} \\
        
        \includegraphics[width=\ww,frame]{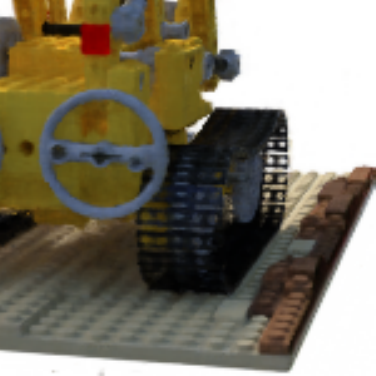} &
        \includegraphics[width=\ww,frame]{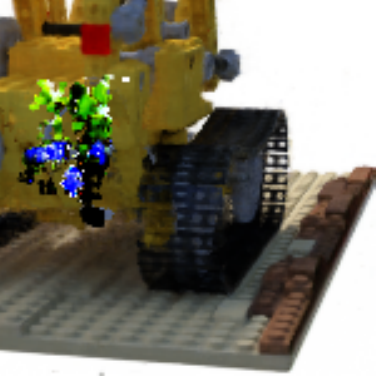} &
        \includegraphics[width=\ww,frame]{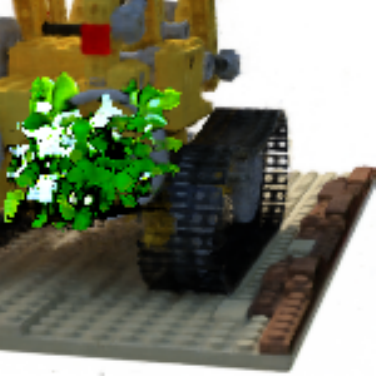} \\

            \scriptsize{original scene} & 
		\scriptsize{sum in activation} & 
		\scriptsize{sum out activation} \\
    \end{tabular}
        \caption{\textbf{Blending Modes.} Guided by ``plant with green leaves and white and blue flowers". When using \cref{sigma_in_eqn} (second column), we allow $F_{\theta}^{G}$ to change the density of the original scene, in this case removing parts of the wheel. When utilizing \cref{sigma_out_eqn} (third column), we can only add additionally density to the scene, so the plant warps around the wheel without changing it.} \label{fig:blender_lego_blending_modes}
\end{figure}

For the density of the blended sample, we consider two options, which have different impact on the results of the blending:
\begin{equation} \label{sigma_in_eqn}
    \sigma(x_i) = \phi(\sigma_O(x_i) + \sigma_G(x_i))
\end{equation}
\begin{equation} \label{sigma_out_eqn}
    \sigma(x_i) = \phi(\sigma_O(x_i)) + \phi(\sigma_G(x_i))
\end{equation}
i.e., summing the densities inside or outside the activation function. When using \cref{sigma_in_eqn} we are summing inside the activation function thus allowing the generator $F_{\theta}^{G}$ to change the original scene density and even remove densities (if $\sigma_G(x_i)< 0$), while in \cref{sigma_out_eqn} we allow $F_{\theta}^{G}$ to only add new densities to the scene. We can choose either of these two options depending on the edit we wish to apply. We then compute the joint transmittance and alpha values according to \cref{quad_eqn}. The resulting blended image $I_{ROI}$ is then used to guide $F_{\theta}^{G}$ during training by measuring its similarity to the input caption using \cref{sim_eqn}. The blending process after training is the same as in Object Insertion/Replacement mode. An illustration of our blending modes on the blender Lego scene is presented in \Cref{fig:blender_lego_blending_modes}. 

\subsection{Objectives and Priors}
\label{sec:objectives}
Previous works \cite{DreamFields2022, VolumeDisentanglement2022, CLIPNeRF2021} and our experiments indicate that a scene representation depending on similarity loss alone (\cref{sim_eqn}) is too unconstrained, resulting in a scene that is not visually compatible to a human, but still satisfies the loss. Thus, we utilize the priors and augmentations mentioned in DreamFields \cite{DreamFields2022} and suggest additional priors 
to get more realistic results.

\textbf{Pose Sampling.} CLIP-NeRF \cite{CLIPNeRF2021} shows the multi-view consistency evaluation of CLIP \cite{CLIP2022}. When using different camera poses and rendering different views of the same object, they still have high similarity, in contrast to different objects which have low similarity even in identical view. DreamFields \cite{DreamFields2022} shows that sampling different camera poses is a good regularizer and improves the realism of the object geometry.  Thus, each iteration we sample a random camera pose around the scene depending on the scene type ($360^{\circ}$ and forward-facing scenes) including its azimuth and elevation angles $(\theta, \phi)$. We found it beneficial to be relatively close to the object during training  to get a bigger object in the rendered view, which in turn yields larger gradients from \cref{sim_eqn}. We set the initial distance $d$ from the ROI according to the camera $AFOV=2\gamma $ and the maximum dimension of the box $e_{\textit{max}}$ and we randomly sample the radius $r$ around this value:

\begin{equation}
\label{radius_eqn}
    d = \frac{e_{\textit{max}}}{2\tan(\gamma/2)}
\end{equation}

\textbf{Background Augmentation.} DreamFields \cite{DreamFields2022} note that when using white or black background during optimization, the scene populates the background, and eventually we get a diffused scene. Thus, we use the same random backgrounds as in DreamFields: Gaussian noise, checkerboard patterns and random Fourier textures from \cite{mordvintsev2018differentiable} to get more sharp and coherent objects.

\textbf{Directional Dependent Prompts.} Due to the fact that there's no constraint on $F_{\theta}^{G}$ to describe the object differently in different views, we concatenate to the original caption a text prompt depending on the current view. For more details, please refer to the supplementary materials.  

\textbf{Transmittance loss.} Same as in DreamFields \cite{DreamFields2022}, in order to get more sparse and coherent results we encourage the generator to increase the average transmittance of the scene inside the box by adding a transmittance loss to the generator objective:
\begin{equation} \label{trans_loss_eqn}
    L_T = -\min(\tau,\; \textit{mean}(T(\boldsymbol{P})))
\end{equation}
Where $\textit{mean}(T(\boldsymbol{P}))$ is the average transmittance of a rendered view from pose $\boldsymbol{P}$ and $\tau$ is the max transmittance.

\textbf{Depth loss.} When blending in forward-facing scenes (such as LLFF dataset \cite{mildenhall2019llff}) and due to the limited viewing intervals, for some captions we get a flat billboard geometry effect and the resulting edit does not seem to have a volume. We encourage the generator to synthesize volumetric 3D shapes by adding a depth loss to the generator objective: 
\begin{equation} \label{depth_loss_eqn}
    L_D = -\min(\rho, \sigma^2(D(\boldsymbol{P})))
\end{equation} 
Where $\sigma^2(D(\boldsymbol{P})))$ is the variance of the disparity map of a rendered view from pose $\boldsymbol{P}$ and $\rho$ is the max variance we allow during training. We gradually introduce $L_T$ and $L_D$ during training using annealing strategy to prevent completely transparent or amorphous scenes. In summary, the final objective for the generator $F_{\theta}^{G}$ is:
\begin{equation} \label{total_loss_eqn}
    L_{\textit{total}} = L_{\textit{sim}} + \lambda_T L_T + \lambda_D L_D
\end{equation} 
Where $\lambda_T, \lambda_D$ are the weights for $L_T,L_D$ accordingly. For more information on implementation details and hyperparameters, please refer to the supplement.

\begin{figure}[h] 
    \centering
    \subfloat[``aspen tree'']
    {
    \includegraphics[width=0.20\textwidth,frame]{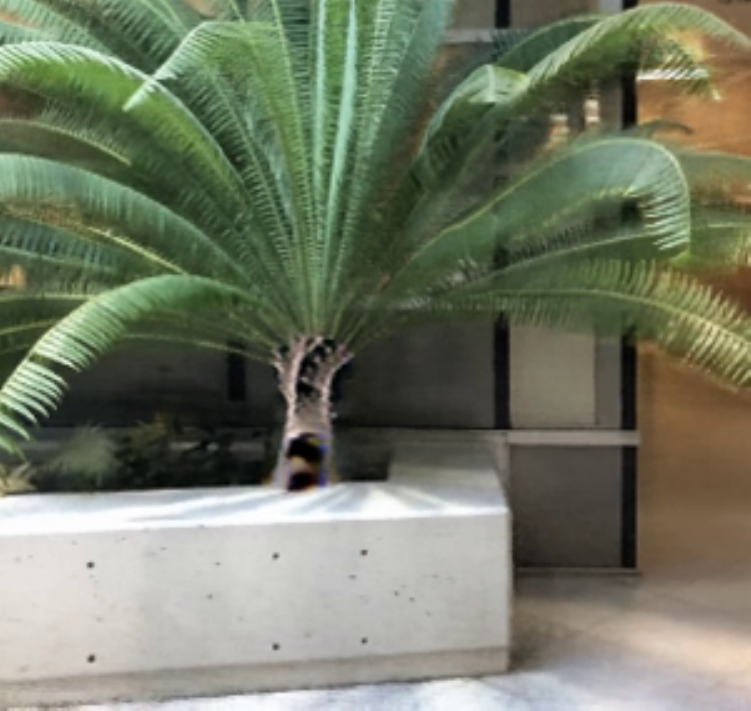}
    \hspace{0.001\textwidth}
    
    \includegraphics[width=0.1895\textwidth,height=0.1895\textwidth,frame]{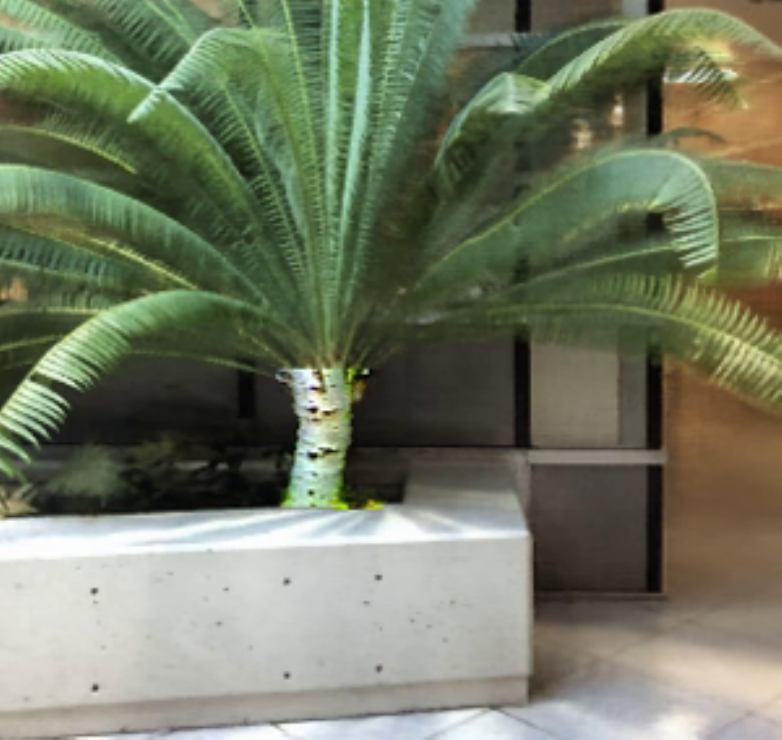}
    }  \label{fig:aspen_tree_comparision}
    \vspace{0.0001\paperheight}
    
    \subfloat[``strawberry'']
    {
    \includegraphics[width=0.20\textwidth,frame]{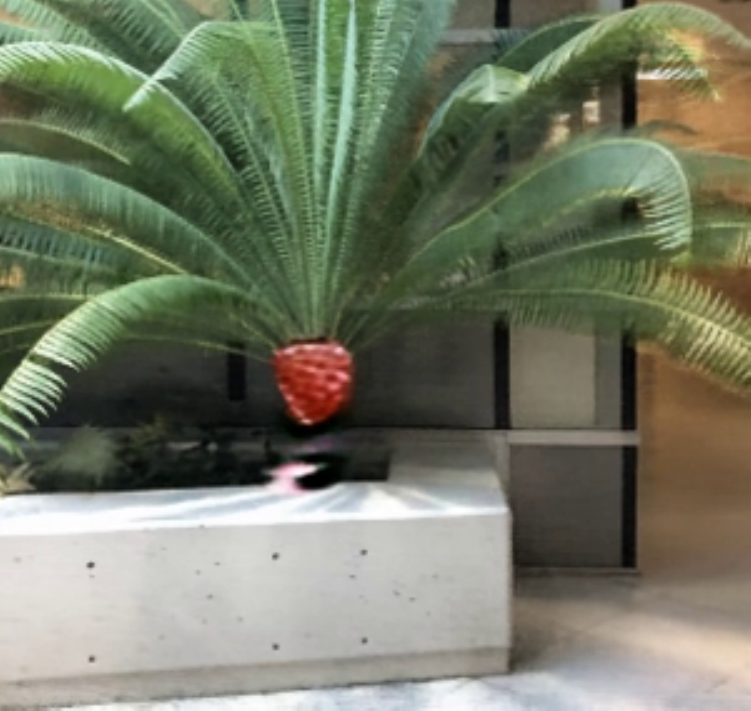}
    \hspace{0.001\textwidth}
    
    \includegraphics[width=0.1895\textwidth,height=0.1895\textwidth,frame]{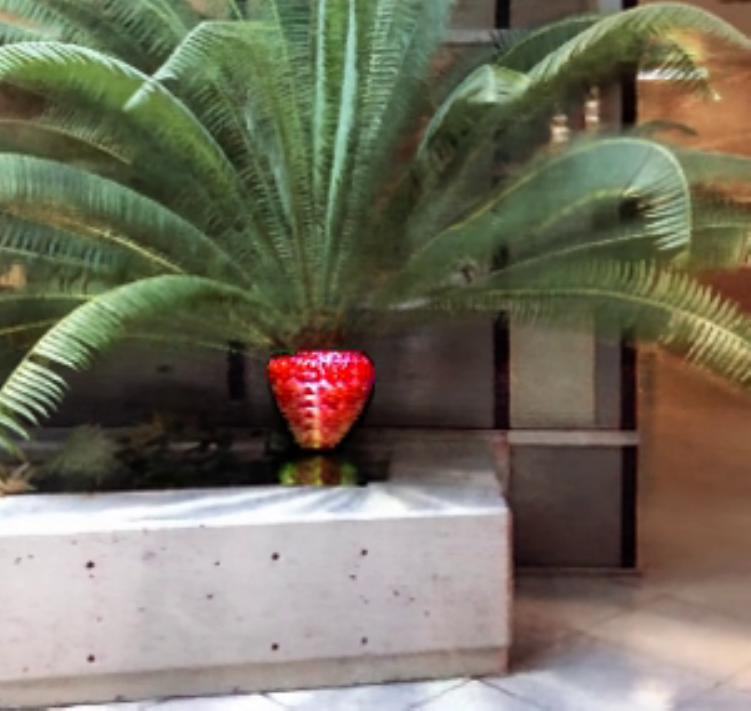}
    }\label{fig:strawberry_comparision}
    \vspace{0.0001\paperheight}

    

    \caption{\textbf{Comparison to \cite{VolumeDisentanglement2022} for object replacement.} We compare our editing capabilities to \cite{VolumeDisentanglement2022} in the fern scene from llff dataset \cite{mildenhall2019llff}. The left and right images in each row are \cite{VolumeDisentanglement2022} and ours, accordingly. Our proposed method exhibits more realistic results that agrees better with the text. For example the edit for the text ``aspen tree'' indeed looks like a trunk of an aspen tree in our edit.} \label{fig:baseline_comparision}
\end{figure}

\section{Experiments}
\label{sec:experiments}

In \Cref{sec:comparisons} we begin by comparing our method both qualitatively and quantitatively to the baseline Volumetric Disentanglement for
3D Scene Manipulation \cite{VolumeDisentanglement2022}. Next, in \Cref{sec:ablation} we demonstrate the effect of our suggested priors and augmentations on improving fidelity and visual quality. Finally, in \Cref{sec:applications} we demonstrate several applications enabled by our framework.

\begin{table}[t]
    \centering 
    \begin{tabular}{c p{1.5cm} p{1.8cm} c}
    \textbf{Method} & CLIP Direction Similarity$\uparrow$ & CLIP Direction Consistency$\uparrow$ & LPIPS$\downarrow$ \\ 
    \hline
    [Benaim 2022] & $0.128$ & $0.736$ & $0.3$  \\
    Ours & $0.143$ & $0.787$ & $0.024$ \\
    \end{tabular}
    \caption{\textbf{Quantitative Evaluation.} Quantitative comparison to \cite{VolumeDisentanglement2022} using the metrics described in \Cref{sec:comparisons}. Our method demonstrates edits that are better align to the input captions and consistent between views, while preserving the background of the scene.}
    \label{tab:baseline_table}
\end{table} 

\begin{figure}[h] 
    \centering
    \subfloat[Without Depth Loss]
    {
    
    \includegraphics[width=0.23\textwidth,frame]{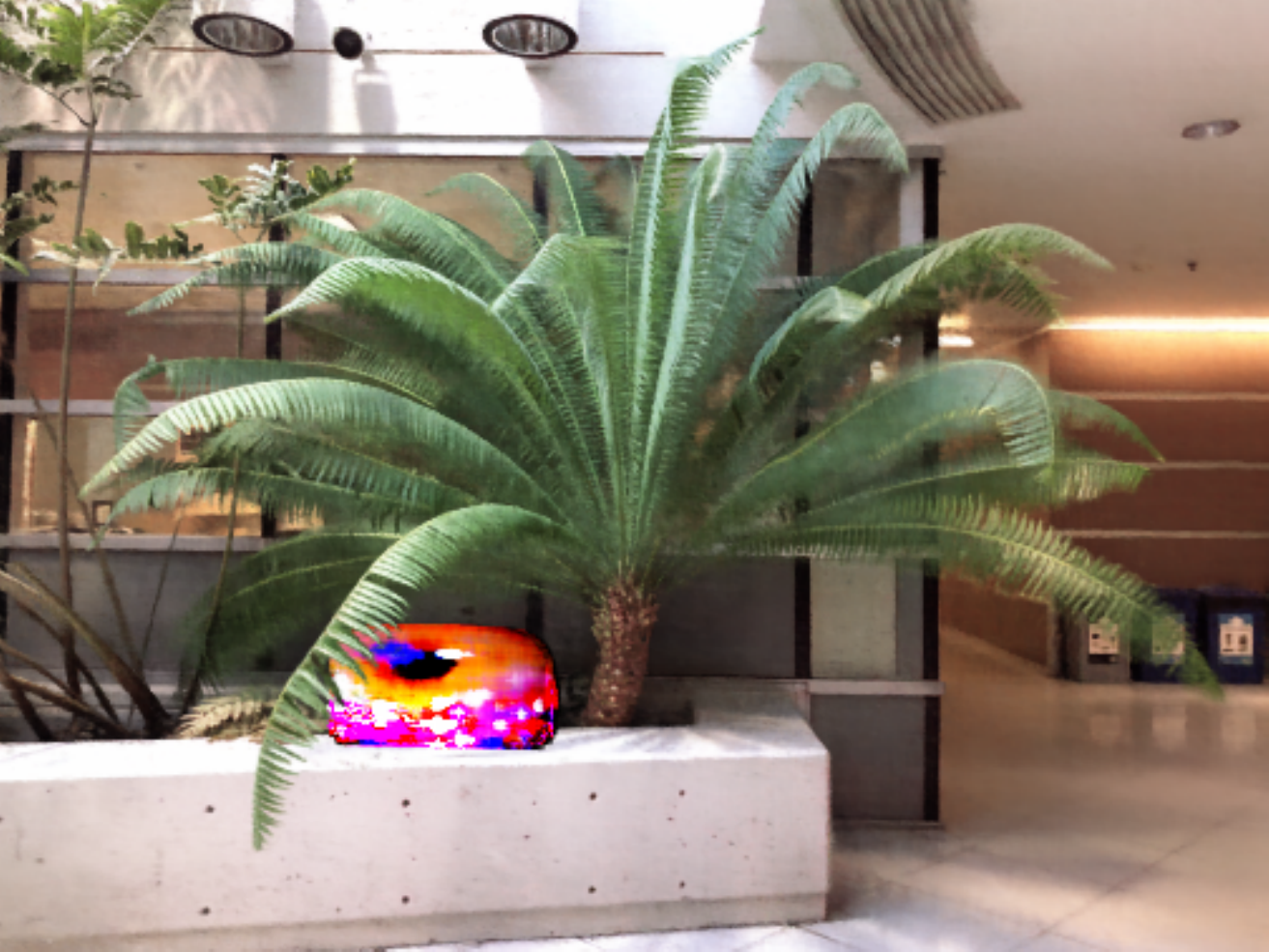}
    \hspace{0.001\textwidth}
    
    
    \includegraphics[width=0.2\textwidth,height=0.173\textwidth,frame]{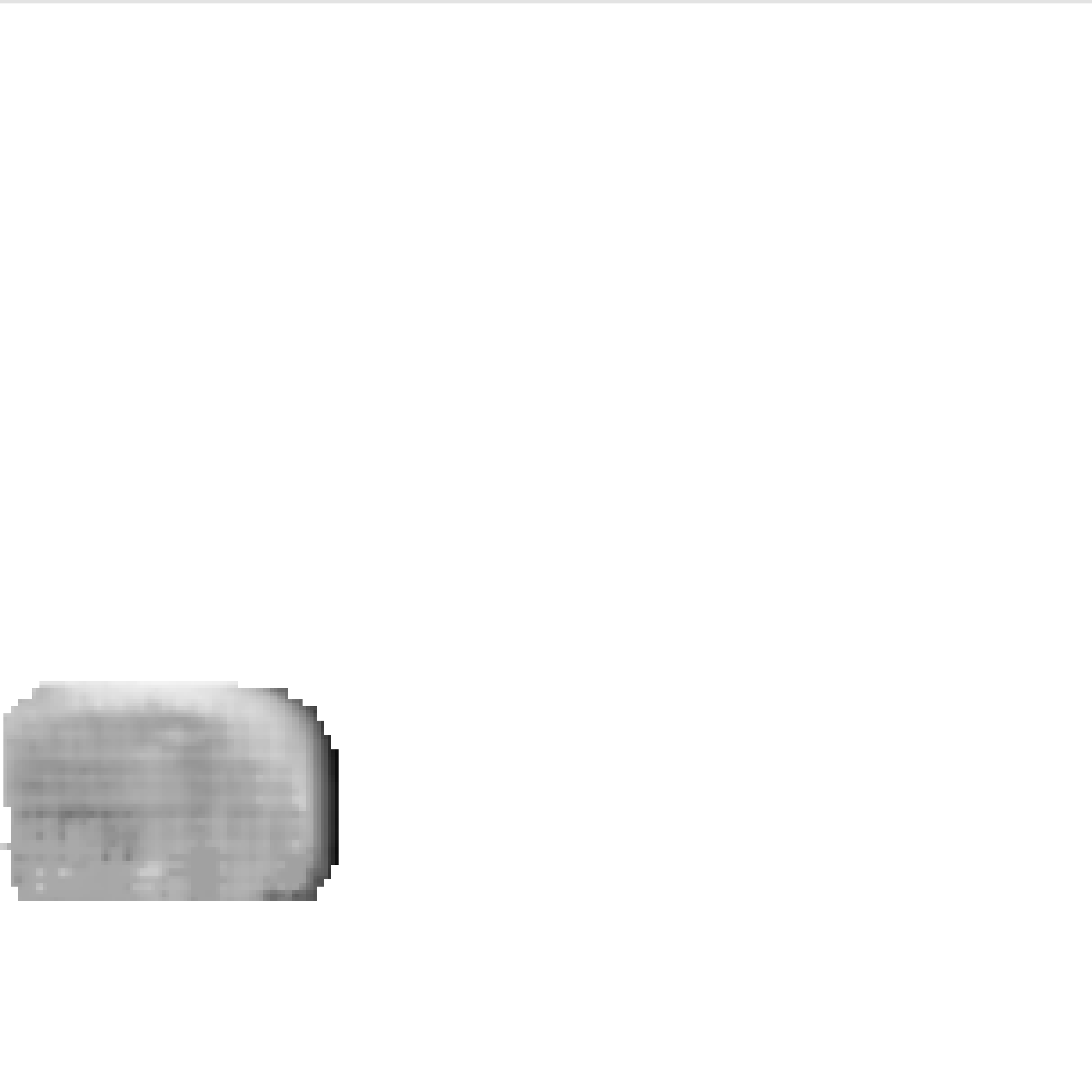}
    } \label{fig:no_depth_loss}
    
    \vspace{0.0001\paperheight}

    \subfloat[With Depth Loss]
    {
    
    \includegraphics[width=0.23\textwidth,frame]
    {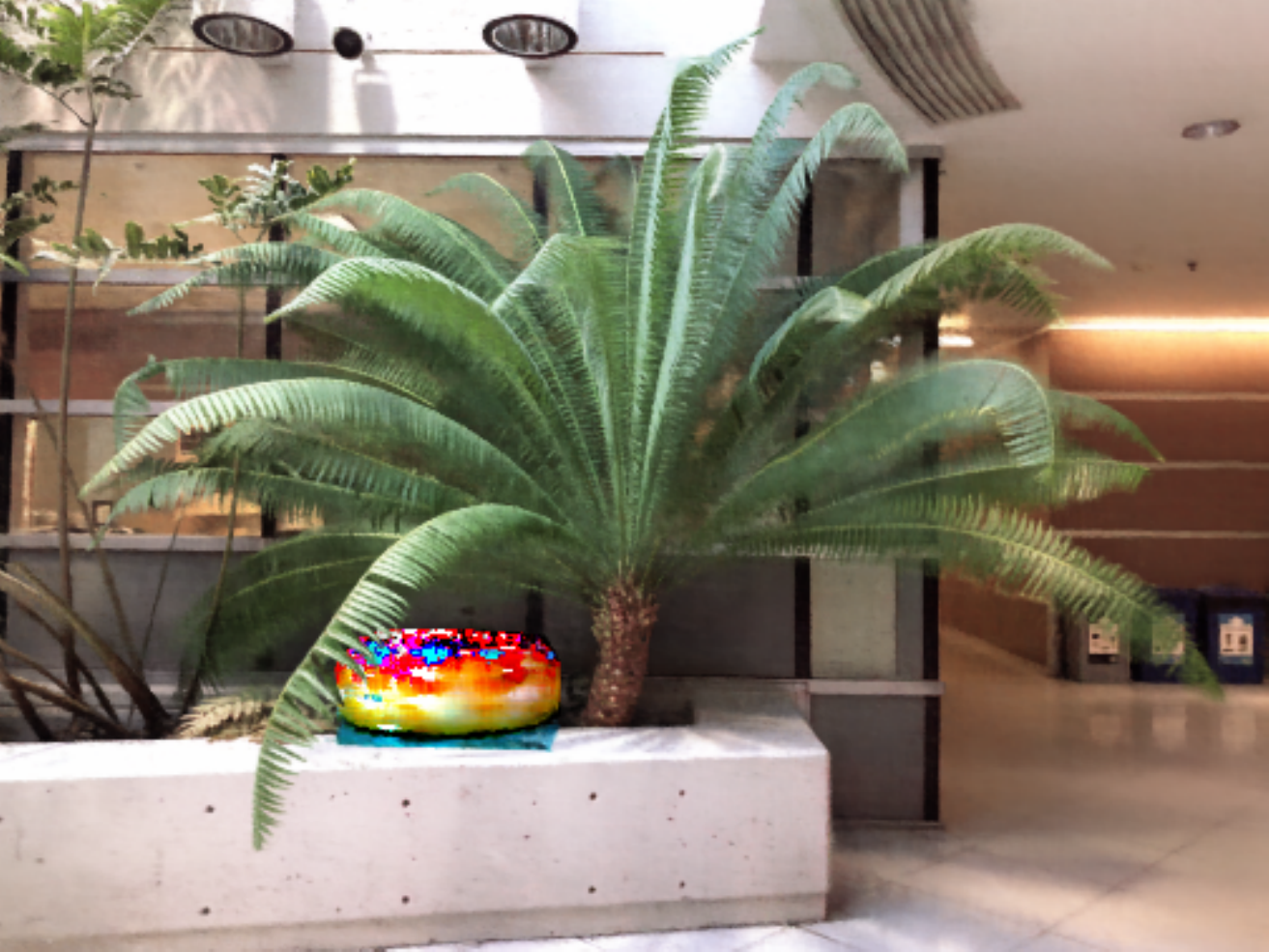}
    \hspace{0.001\textwidth}
    

    \includegraphics[width=0.2\textwidth, height=0.173\textwidth,frame]{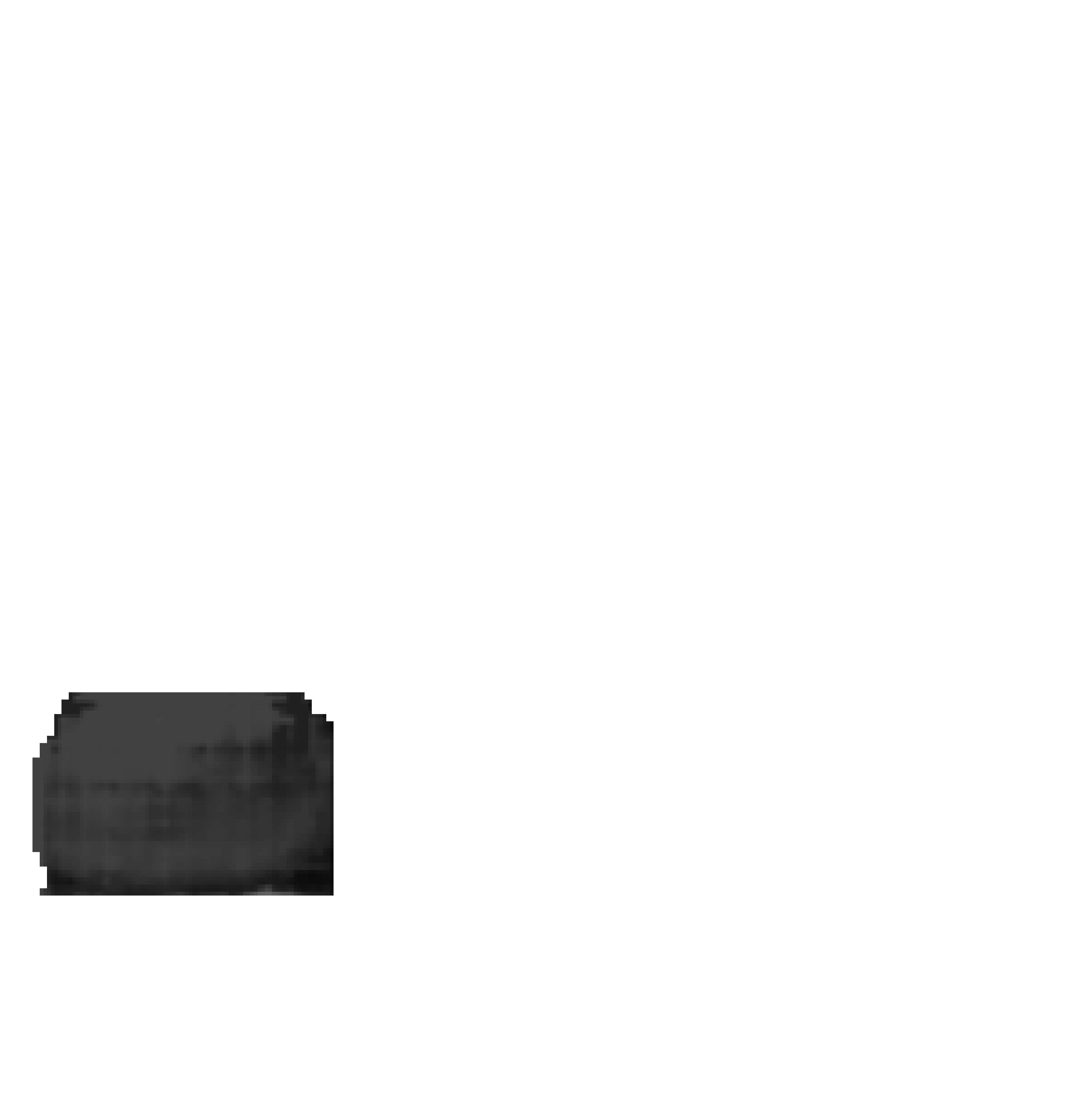}
    
    }  \label{fig:with_depth_loss}
    \vspace{0.0001\paperheight}

    \caption{\textbf{Depth Loss Impact.} Comparison of synthesizing a ``donut covered with glaze and sprinkles'' from COCO dataset \cite{lin2014microsoftcoco} on a limited view scene with and without our suggested depth prior. The first column display a view of the edited scenes and the second column displays the disparity map of the synthesized objects. In (a) the results are more flat, which can be clearly seen in the disparity map.} \label{fig:depth_loss_impact}
\end{figure}

\subsection{Comparisons}
\label{sec:comparisons}


Our qualitative comparisons to Volumetric Disentanglement \cite{VolumeDisentanglement2022} are shown in \Cref{fig:baseline_comparision}. Since the implementation of \cite{VolumeDisentanglement2022} is not currently available, we preform the comparisons using the examples from their project 
page\footnote{\href{https://sagiebenaim.github.io/volumetric-disentanglement/}{\textit{https://sagiebenaim.github.io/volumetric-disentanglement/}}}. As can be seen from the results in \Cref{fig:baseline_comparision}, our results exhibit richer and more natural colors and are aligned better with the text. 
To test these observations quantitatively, in \Cref{tab:baseline_table} we compare our proposed method to \cite{VolumeDisentanglement2022} using three metrics: \\
\textbf{(1) CLIP Direction Similarity}, a metric originally introduced in StyleGAN-NADA \cite{gal2021stylegan}, measures how well the change between the original and edited views is aligned with the change in the texts describing them (in the CLIP embedding space). \\
\textbf{(2) CLIP Direction Consistency}, introduced by Haque \cite{haque2023instruct}, measures the cosine similarity of the CLIP embeddings of a pair of adjacent frames. For each edit, we take 6 consecutive frames, compute the metric for each consecutive pair, and average the results among all pairs.\\
Finally, we use \textbf{(3) LPIPS} \cite{lpips} to measure the difference between the original and edited scenes, with the ROI masked, for comparing the background preservation. As can be seen from \Cref{tab:baseline_table}, our model outperforms the baseline in all metrics, which implies that our generated objects match better to the input text captions, they are more consistent from any view and, on the other hand, our method manages to keep the rest of the scene untouched.

\begin{figure*}[h]
    \centering
    \setlength{\tabcolsep}{0.5pt}
    \renewcommand{\arraystretch}{0.5}
    \setlength{\ww}{0.5\columnwidth}
  
    \begin{tabular}{cccc}        
        \includegraphics[width=\ww,frame]{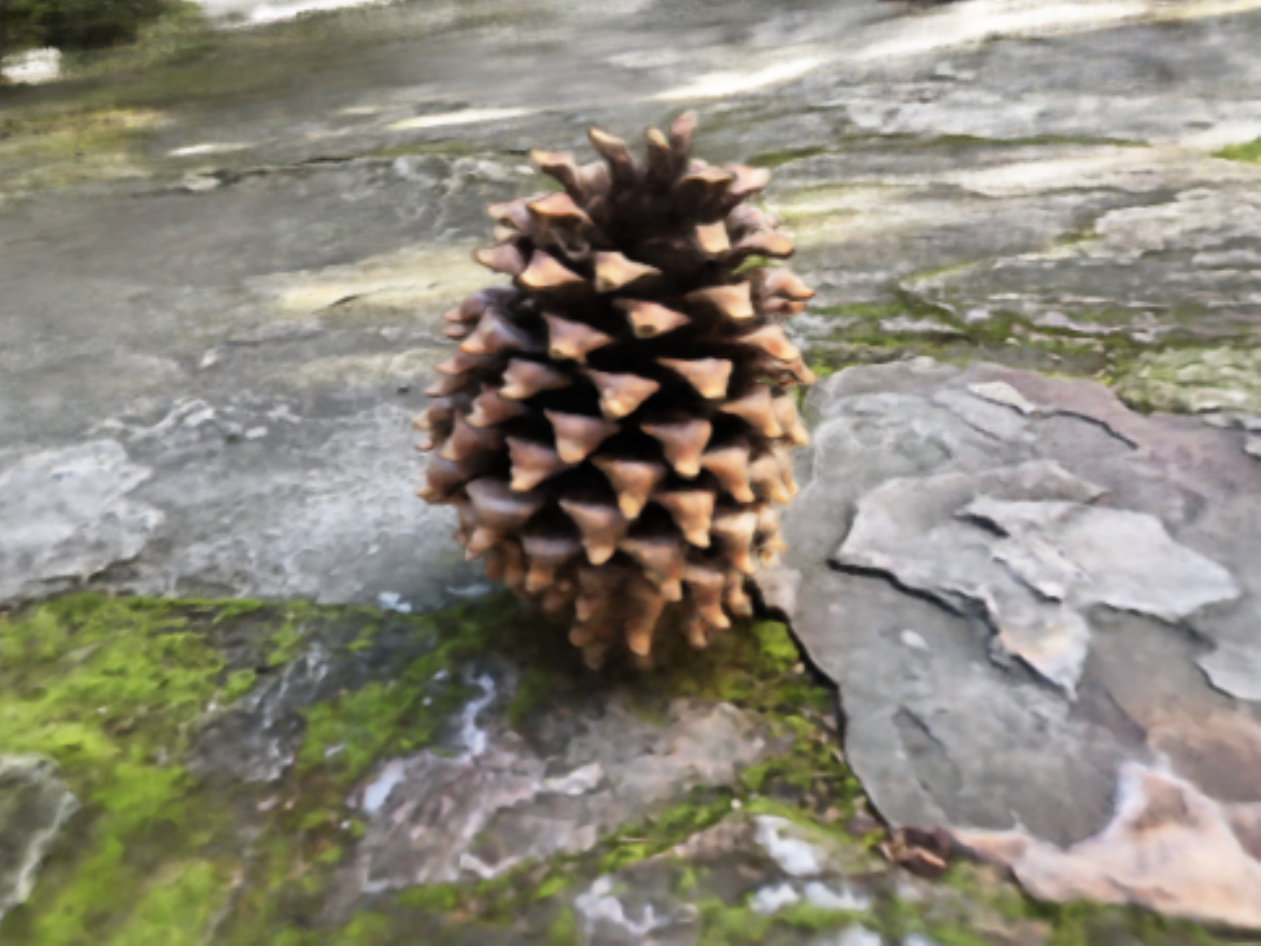} &
        \includegraphics[width=\ww,frame]{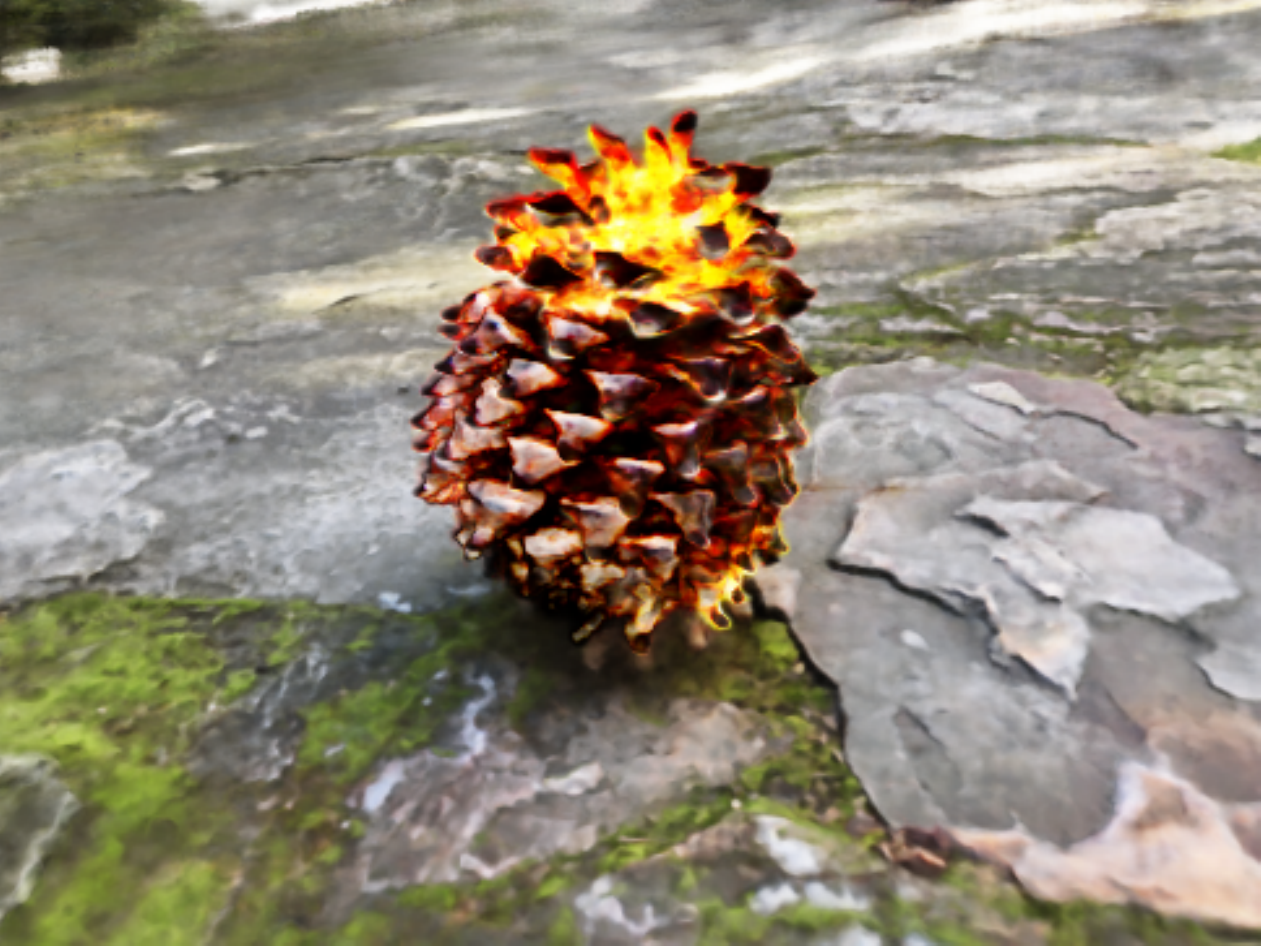}&
        \includegraphics[width=\ww,frame]{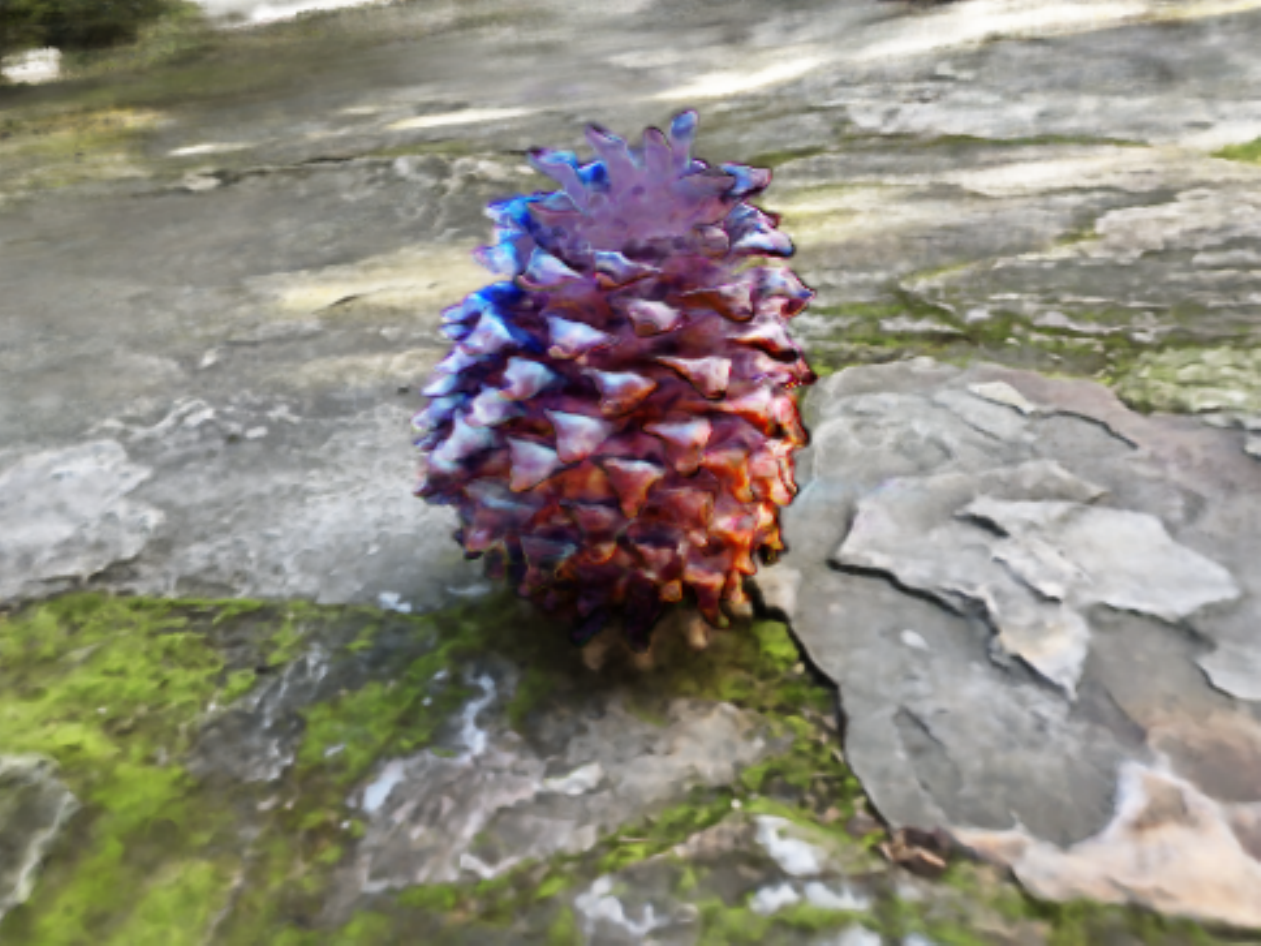} &
        \includegraphics[width=\ww,frame]{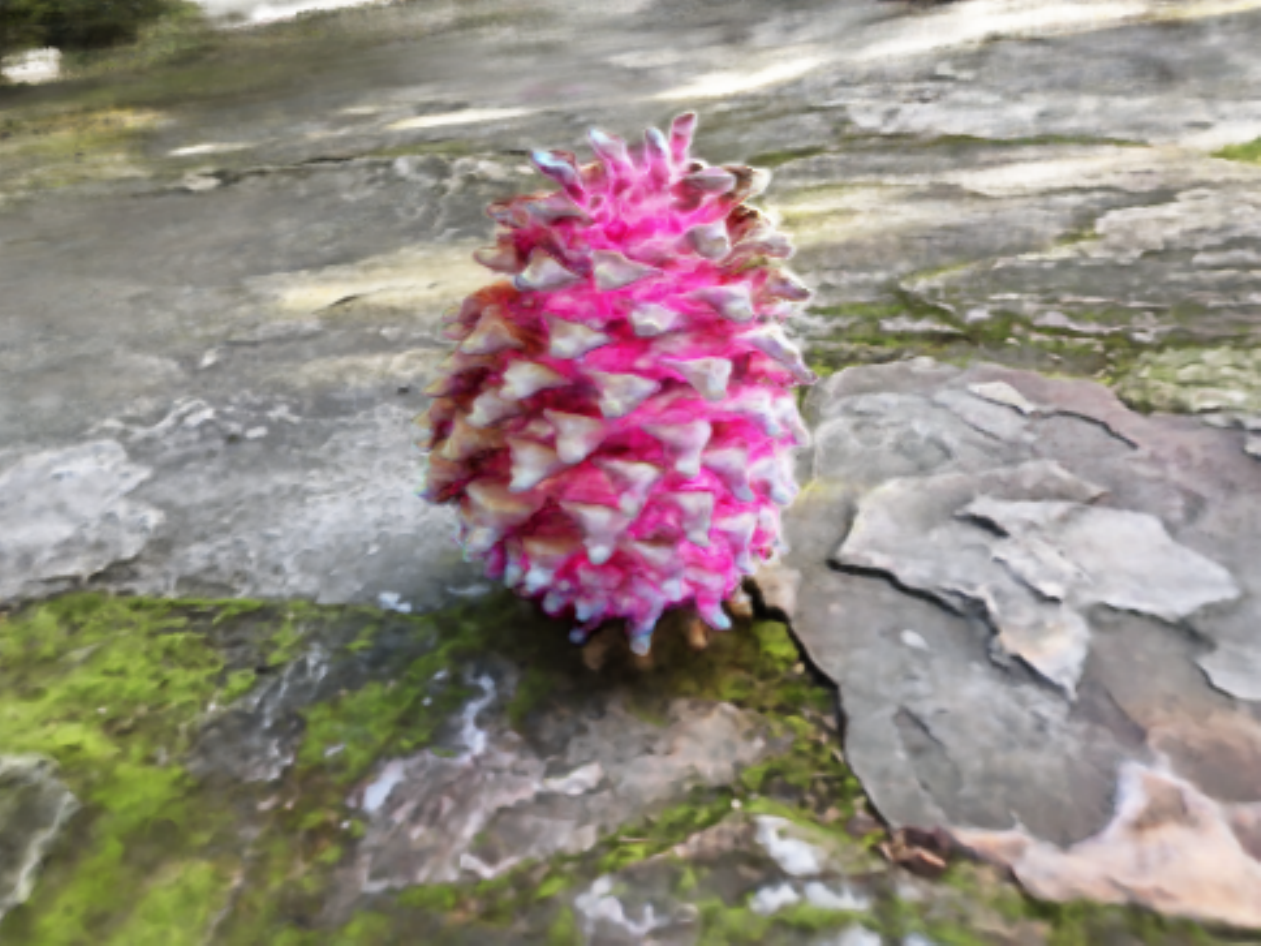} \\
		
		Original Scene & 
		``burning pinecone'' & 
		``iced pinecone'' &
            ``pinecone made of pink wool'' \\
            \\
    \end{tabular}

    \begin{tabular}{cccc}        
        \includegraphics[width=\ww,frame]{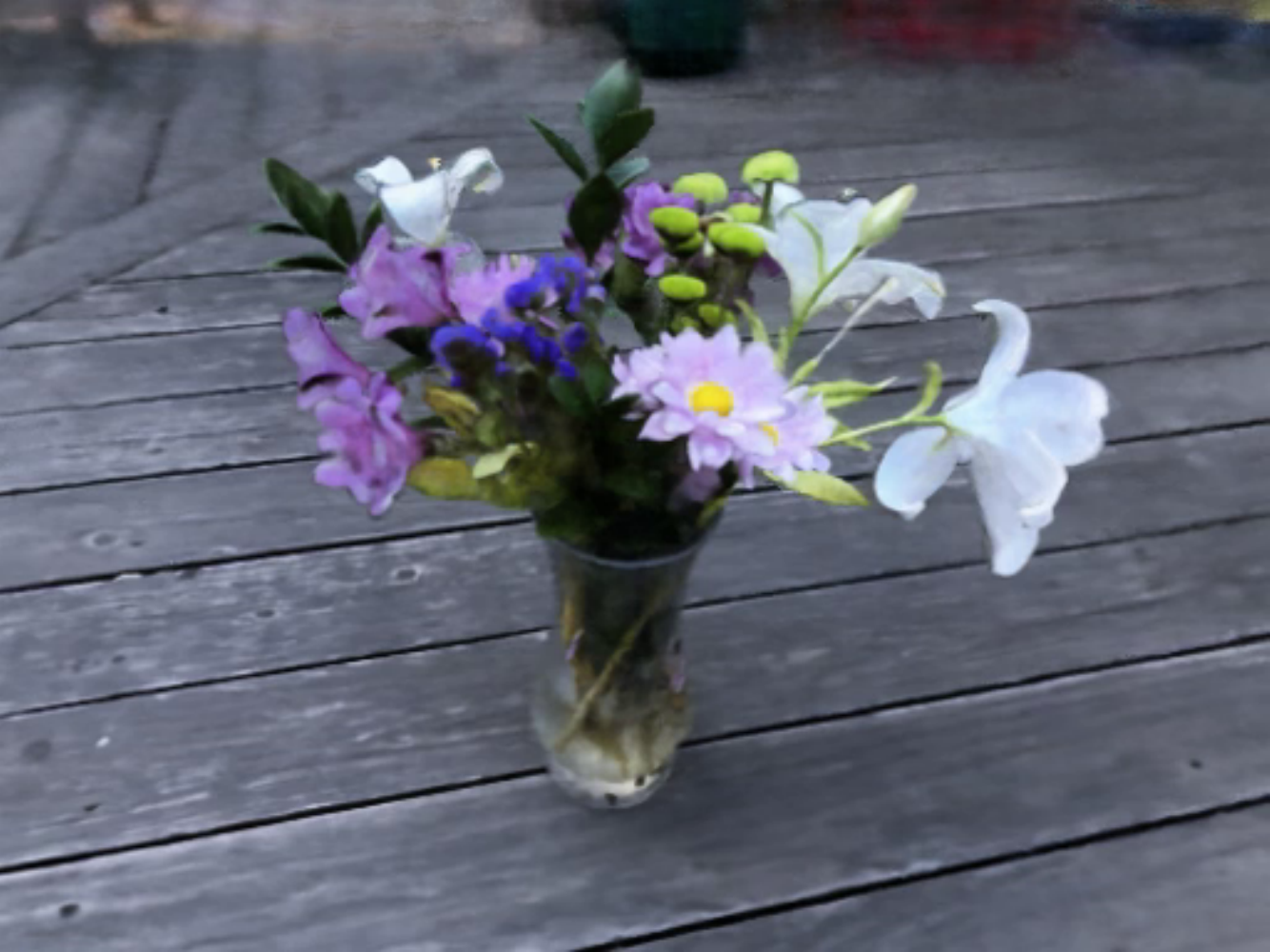} &
        \includegraphics[width=\ww,frame]{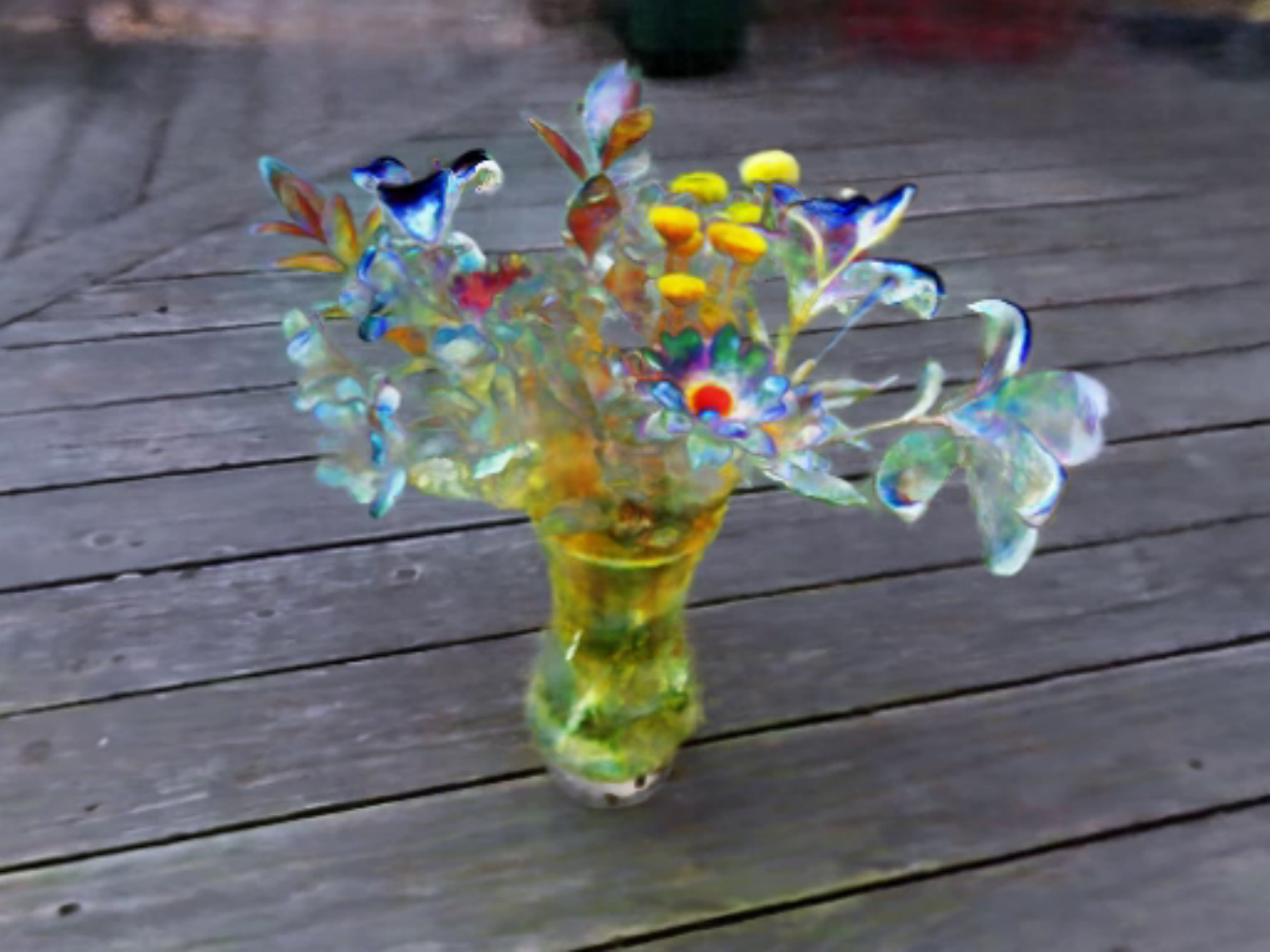}&
        \includegraphics[width=\ww,frame]{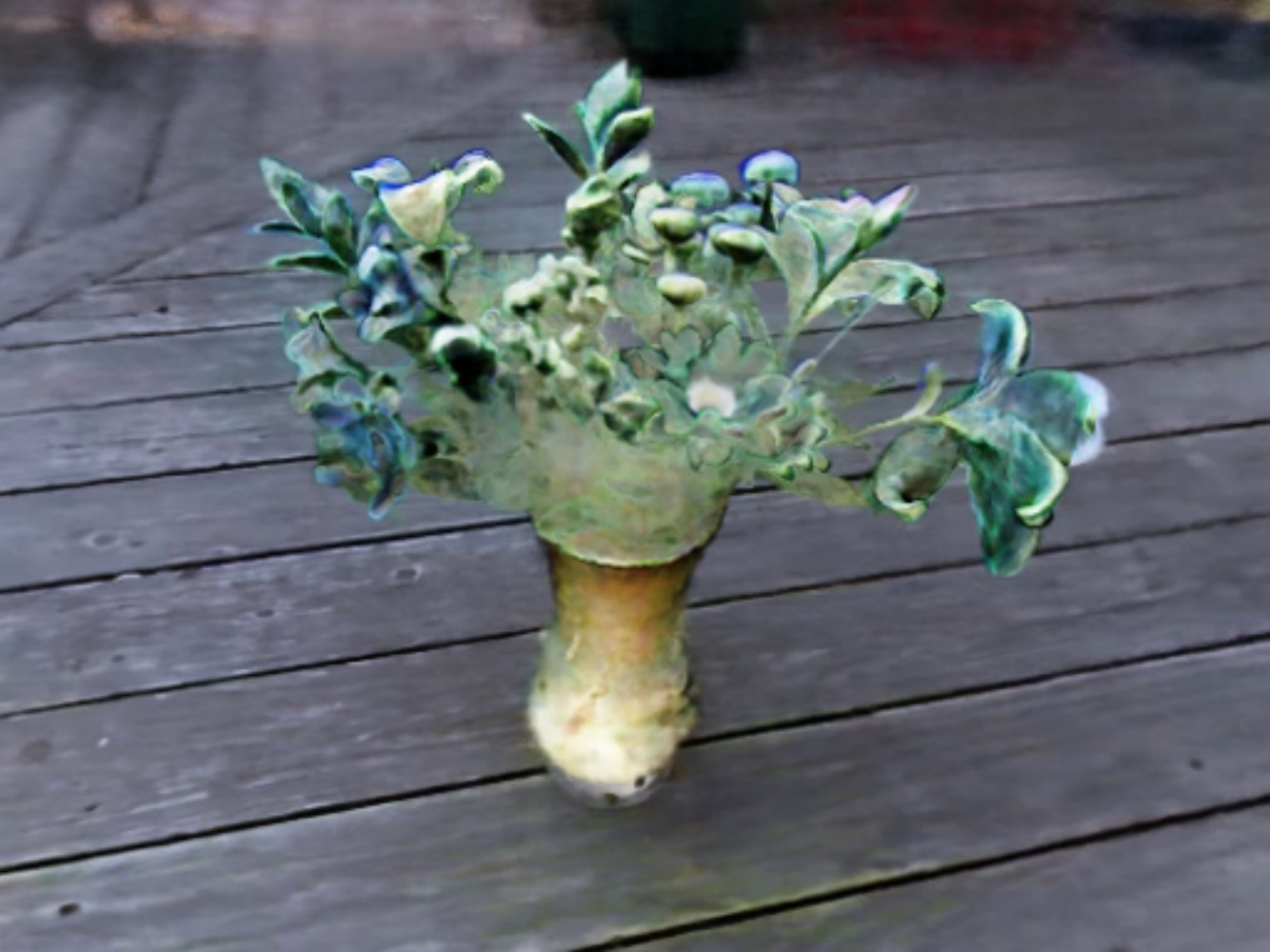}&
        \includegraphics[width=\ww,frame]{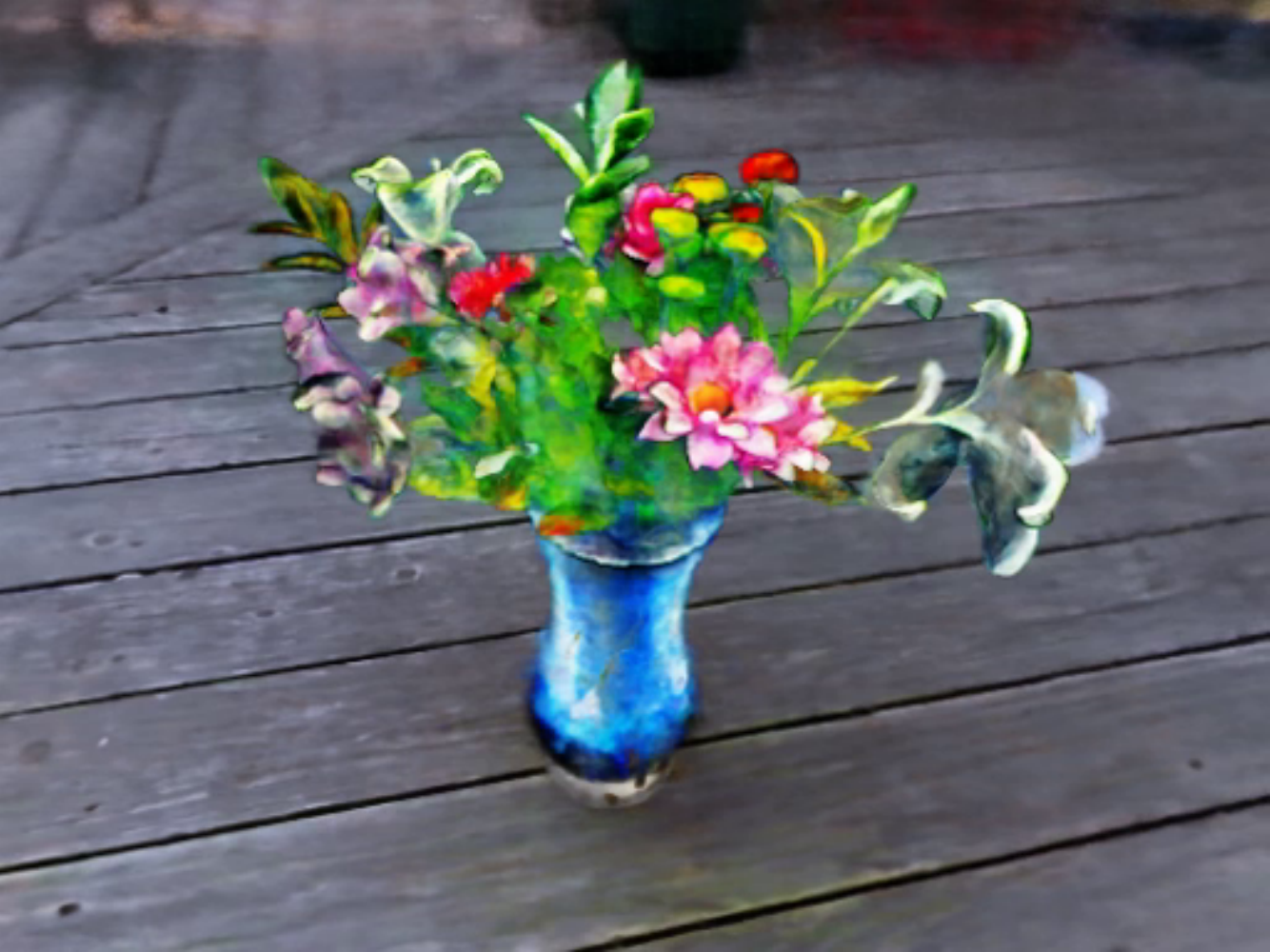}\\
		
		Original Scene & 
		``vase made of glass'' & 
		``vase made of stone'' &
            ``water paint of a vase'' \\
    \end{tabular}
    
    \caption{\textbf{Texture Editing.} We can change only the texture of an object by freezing the layers responsible for the density and training only the layers that impact the color of the scene. To get a smooth blending, we utilize \cref{distance_blending} to blend the scene inside and outside the ROI.} \label{fig:pinecone_texture}
\end{figure*}

\subsection{Ablation Study}
\label{sec:ablation}
To show the importance of our proposed augmentations and priors, we use the R-Precision score \cite{RPrecisionpark2021benchmark} using both CLIP and BLIP \cite{CLIP2022,li2022blip,li2023blip2} as the metric language-image model to measure how well the generated images align with the true caption. Similar to DreamFields \cite{DreamFields2022}, we use a randomly selected subset of 20 samples (due to time and resources limitations) from the object-centric dataset which contains 153 images and captions from COCO dataset \cite{lin2014microsoftcoco} as our ground truth. The objects are synthesized using the given captions and blended into an empty region in the llff fern scene. Due to the fact we are training on the same CLIP model, we test our results with a different language-image model, BLIP2 \cite{li2023blip2}. The results of both metrics are presented in \Cref{tab:ablation}. The directional dependent prompts seem to only slightly improve the results, probably due to the forward-facing nature of the scene. When rendering from limited camera positions and viewing angles and without our proposed depth priors, the results deteriorate. To test this conclusion visually, in \Cref{fig:depth_loss_impact} we compare the task of inserting a new object into an empty region of the fern llff scene \cite{mildenhall2019llff} with and without the depth loss. As can be seen from the figure, when using our proposed depth prior, the generated object has more volume and looks more natural and consistent. For additional details, please refer to the supplement.

\begin{table}[t]
    \centering 
    \begin{tabular}{p{3.cm} p{2cm} p{2cm}}
    \textbf{Method} & CLIP    & BLIP   \\ 
    &  R-Precision $\uparrow$ &  R-Precision $\uparrow$ \\
    \hline
    COCO GT & $0.933$ & $0.98$  \\
    Ours(full pipeline) & $0.86$ & $0.8$ \\
    Ours(no dir prompts) & $0.85$ & $0.8$ \\
    Ours(no depth prior) & $0.81$ & $0.78$ \\
    \end{tabular}
        \caption{\textbf{Ablation study.} We test our proposed priors and augmentations on a subset of captions and images from COCO dataset \cite{lin2014microsoftcoco}. The CLIP and BLIP R-Precision scores utilize CLIP B-32 and BLIP2 architecture accordingly. The first row shows the scores of the GT COCO image, the second row shows our method scores using all the priors and augmentations as described in \Cref{sec:method} and the last two rows present the scores when taking out the directional dependent prompts and the depth loss.}\label{tab:ablation}
\end{table}

\subsection{Applications}
\label{sec:applications}


In this section, we demonstrate the applicability of our framework for several 3D editing scenarios.

\textbf{New Object Insertion.} Using the method described in \Cref{sec:method}, and by placing the ROI box in an empty space of the scene, we can synthesize a new object given a text prompt and blend it into the original scene. Visual example of this application can be seen in \Cref{fig:depth_loss_impact} and in the supplement. 

\textbf{Object Replacement.} To replace an existing object in the scene with new synthesized content, we place the ROI 3D box in the required area (enclosing the object to be replaced), and perform the training process described in \Cref{sec:method}. In \Cref{fig:blender_ship} we demonstrate the replacement of the sea in the blender ship scene, while in \Cref{fig:baseline_comparision} we replace the fern's trunk.

\begin{figure}[hb]
    \centering
    \setlength{\tabcolsep}{0.5pt}
    \renewcommand{\arraystretch}{0.5}
    \setlength{\ww}{0.45\columnwidth}
  
    \begin{tabular}{cc}        
        \includegraphics[width=\ww,frame]{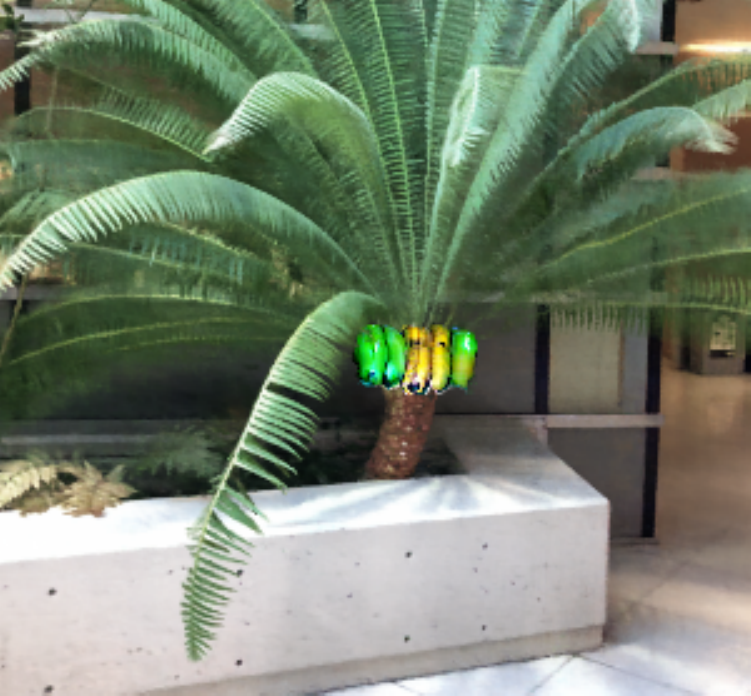} &
        \includegraphics[width=\ww,frame]{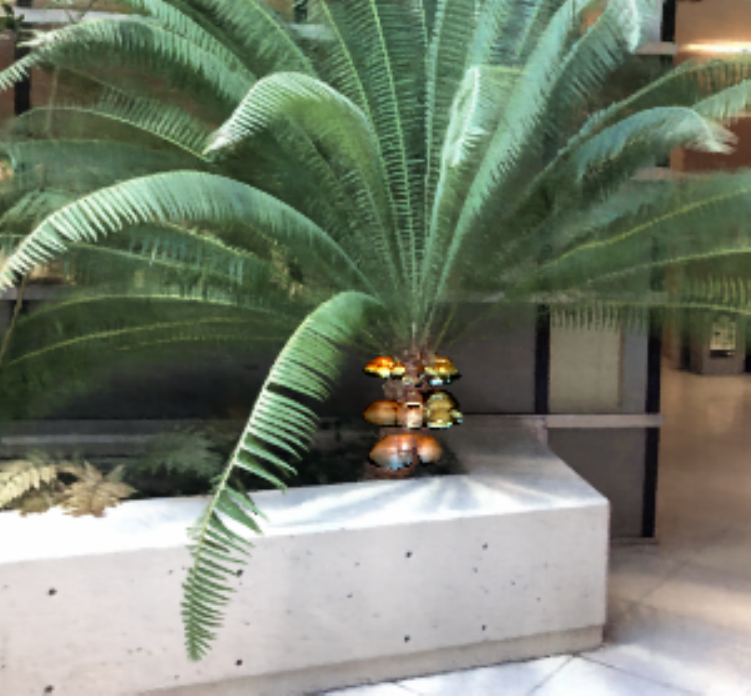} \\
        
        \includegraphics[width=\ww,frame]{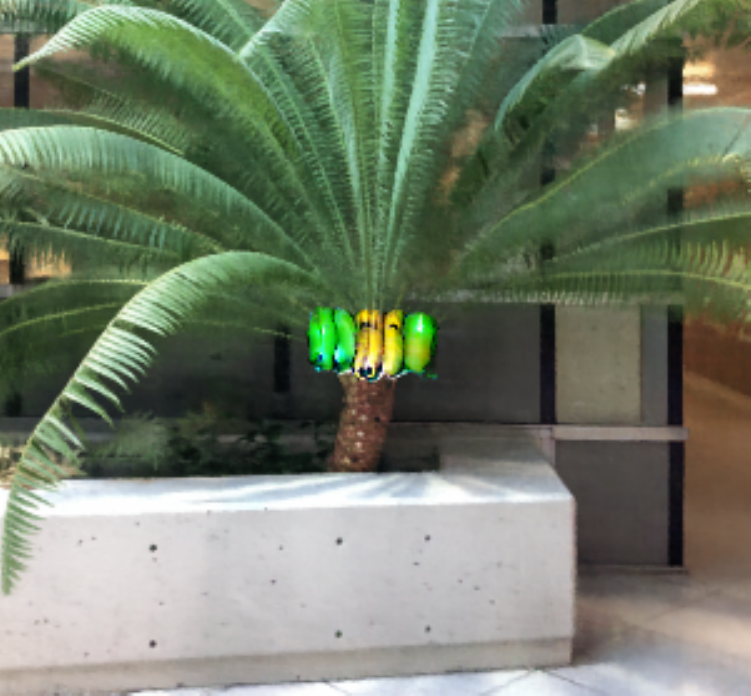} &
        \includegraphics[width=\ww,frame]{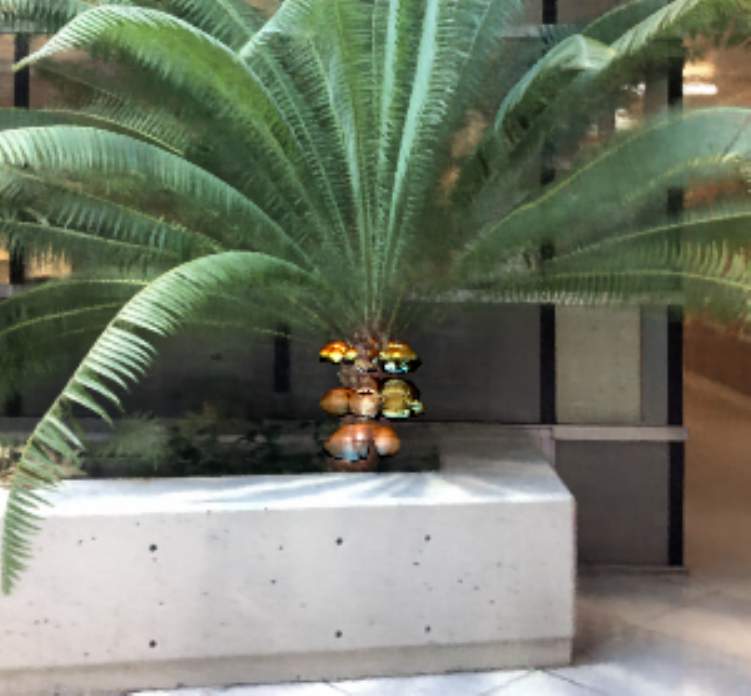} \\
		
            \scriptsize{"a green and yellow bananas".} &
            \scriptsize{"a clusters mushrooms".} \\
    \end{tabular}
    \caption{\textbf{Blending Densities Inside Activation.} We demonstrate our suggested blending procedure for blending the original and synthesized objects inside the ROI in llff fern scene \cite{mildenhall2019llff} using \cref{sigma_in_eqn} for summing the densities.} \label{fig:fern_blending}
\end{figure}

\textbf{Blending of Objects.} To preform blending between the original and the generated object inside the ROI, we utilize the object blending process described in \Cref{sec:method}. In \Cref{fig:blender_lego_blending_modes} and \Cref{fig:fern_blending} we demonstrate this blending on blender lego and llff fern scenes. 

\textbf{Texture Editing.} We enable texture editing by training only the color-related layers of $F_{\theta}^{G}$ and freezing all the other layers in a similar way as in \cite{CLIPNeRF2021}. For seamless blending results, we utilize \cref{distance_blending}. In \Cref{fig:pinecone_texture} we demonstrate this edit method on 360 scenes. For additional results and videos please refer to supplement.

\section{Limitations and Conclusions}
\label{sec:limitations}

We introduced a novel solution to blend new objects into an existing NeRF scene with natural looking and consistent results by utilizing a language-image model to steer the generation process towards the edit and by introducing novel priors, augmentations and volumetric blending techniques for improving the final edited scene. We tested our method on a variety of scenes and text prompts and showed the applicability of our framework on several editing applications. We believe that our framework can be utilized in a variety of applications due to the ease and intuitive interaction enabled by our interface.

One of the limitations of our framework is that currently it can't edit multiple objects in a given scene, such as changing two wheels of a 3D car without impacting the rest of the scene. Additionally, the use of a box as our ROI scene shape can be sometimes limiting; for example, when trying to edit a circular scene like the blender ship scene in Figure \ref{fig:blender_ship}, a cylinder could be preferable.
Due to the fact we are rendering one view in each training step, we may get artifacts like multiple heads on the generated object.
The quality of our generated objects can be improved by utilizing the recent progress in diffusion models, we leave it as a future work to combine our suggested blending framework with these models.


\paragraph{Acknowledgements:}
This work was supported in part by the Israel Science Foundation (grants No.~2492/20, and 3611/21).


{\small
\bibliographystyle{ieee_fullname}
\bibliography{egbib}
}

\clearpage
\appendix

\section{Implementation Details}
\label{sec:implemenation_details}
In this section we provide additional implementation details.

\subsection{ROI Specification Interface}
\label{sec:localization}
To specify the ROI and use it to decompose the scene, we introduce a graphic interface that enables positioning an axis-aligned 3D box inside the scene. Given the 3D position of the center, as well as the axis dimensions, of the box, rendering of the scene is performed from the provided camera position using the original NeRF model  $F_{\theta}^{O}$. The edges of the 3D box are then projected onto the image plane using the camera matrix. To provide intuitive feedback regarding the location of the box in the scene, we utilize the depth map of the scene to remove parts of the box edges that are occluded by the scene. In this manner, the user is able to specify the ROI in a precise and intuitive way by moving the box  and modifying its dimensions while being able to inspect the location from any point of view.

\subsection{Pose Sampling}
\label{sec:pose_sampling}
In each training step, we sample a camera pose from a range of distances and angles, depending on the scene type.
In the blender and 360 scenes, we sample azimuth and elevation angles in the ranges: $\theta\in[-180^{\circ},180^{\circ}]$, $\phi\in[-90^{\circ},15^{\circ}]$. For the radius, we first calculate the initial distance according to eq. (10) and then randomly sample the radius around this value. In llff dataset \cite{mildenhall2019llff} we sample the camera pose from a spiral curve as used in the original NeRF implementation
\footnote{https://github.com/bmild/nerf}
. The curve is randomly sampled from a range of distances and radii in each axis.
After sampling a camera pose, we recenter its rays around the ROI by moving its center location according to the center of mass inside the ROI
(tracked by exponential moving average during training), but allow with a probability $p\in [0,1]$ (hyperparameter, set to 0.1 in our experiments) to recenter the rays to a different point inside the ROI, with the aim of obtaining more versatile objects and densities. 
Additionally, we set the near and far planes $(n,f)$ according to the box location and size in order to be more concentrated around the ROI and get more sample points per ray in this area:
\begin{equation} \label{nf_plains_eqn}
n = d - \frac{D}{2}, \quad f = d + D,
\end{equation}
where $d$ is the distance of the camera from the center of mass inside the box and $D$ is the box diagonal length.

\subsection{Hyperparameters}
In our experiments we set the max transmittance of $L_T$, the max variance of $L_D$ and the weights of the losses to:  $\tau=0.88$, $\rho=0.2$, $\lambda_{T}=0.25$, $\lambda_{D}=4$. 
We use the same network architecture as in \cite{NeRF2020} and the same hyperparameters and learning rates. To guide our model, we use the CLIP B/32 architecture.

\subsection{Training}
We train our model with a random seed value of 123 for all of our experiments. In experiments, we render the views at 168x168 resolution and up-sample to 224x224 resolution before feeding them to CLIP \cite{CLIP2022}. In the Comparisons and a Ablation study sections, we train the generator for 40,000 iterations and for the other figures in the main paper, the views resolution and the number of iterations depends on the complexity of the synthesized object and hardware limitations. We train with  $4 \times 24$ GB A5000 GPUs. Training takes between a few hours to one day. We find that the primary driver for runtime/hardware requirements are the view resolution and the size of ROI (which require rendering more points along each ray).

\subsection{Directional Dependent prompts}
As described in the main paper, each iteration we concatenate a text prompt to the input caption depending on the camera location in the scene. We use the direction prompts below depending on the location:
\begin{itemize}
\item ", top-down view"
\item ", front view"
\item ", side view"
\item ", back view"
\end{itemize}

In forward-facing scenes like llff dataset \cite{mildenhall2019llff} we use the first three captions.

\section{Additional Experiments Details}
\label{sec:additional_exp_detials}
In this section we provide additional information regrading the experiments from the main paper.
\subsection{Metrics}
In our quantitative evaluation we report four metrics: CLIP Direction Similarity, CLIP Direction Consistency, LPIPS and R-Precision.\\

\textbf{CLIP Direction Similarity} introduced in \cite{gal2021stylegan} as a direction loss which measures the similarity between the change in the text descriptions and the change in the images. We use a variation of this metric so that high similarity will have high metric score:

\begin{equation}
\begin{split}
    \Delta T = E_T(T_{e}) - E_T(T_{o}) \\
    \Delta I = E_I(I_{e}) - E_I(I_{o}) \\
    L_{direction} = \frac{\Delta T\cdot \Delta I}{|\Delta T\ ||\Delta I\ |}
\end{split}
\end{equation}

When $E_T$, $E_I$ are the text and image encoders of CLIP, $T_{e}$, $T_{o}$ are the text captions describing the edited and original scene inside the ROI and $I_{e}$, $I_{o}$ are the according edited and original scenes views. In our experiments on the fern llff scene \cite{mildenhall2019llff}, we set $T_{o}$ to: "a photo of a fern trunk". \\

\textbf{CLIP Direction Score} introduced in \cite{haque2023instruct} measures the consistency between adjacent frames by calculating the CLIP embeddings of two corresponding pairs of consecutive views, one from the original scene and one from the edited scene. Similar to CLIP Direction Similarity metric, we then compute the similarity between the change in the original and edited scene views to get the final consistency score:

\begin{equation}
\begin{split}
    \Delta I_{o} = E_I(I^{o}_{i+1}) - E_I(I^{o}_{i}) \\
    \Delta I_{e} = E_I(I^{e}_{i+1}) - E_I(I^{e}_{i}) \\
    L_{direction} = \frac{\Delta I_{o}\cdot \Delta I_{e}}{|\Delta I_{o}\ | |\Delta I_{e}\ |}
\end{split}
\end{equation}

When $I^{o}_{i}$, $I^{o}_{i+1}$ and $I^{e}_{i}$, $I^{e}_{i+1}$  are the original and edited consecutive views pairs. 
In our experiments we compute this score on six consecutive views and average the results.\\

\textbf{LPIPS} or Learned Perceptual Image Patch Similarity, is used to judge the perceptual similarity between two images, \cite{lpips} shows that this metric match human perception. The metric computes the similarity between the activation's of the two images for some network architecture. In our experiments we use LPIPS with pre-trained alexnet architecture \cite{krizhevsky2017imagenet} to measure the background similarity between the original and the edited scenes by masking the ROI region.\\

\textbf{R-Precision} \cite{RPrecisionpark2021benchmark} measures how well a rendered view of the synthesis object align with the text caption used to generate it. It computes the precision of the rendered views over a group of text captions using a retrieval model. Similar to DreamFields \cite{DreamFields2022} we collect an object-centric captions dataset from COCO dataset \cite{lin2014microsoftcoco} and sample 20 captions that will be used for training our model. We than compute the precision of the rendered views per synthesis object over the 153 captions. As the language image model backbone of the score, we use both CLIP \cite{CLIP2022} and BLIP2 \cite{li2023blip2}, since we use CLIP to train our model.

\section{Concurrent Work}
\label{sec:concurrent_work}

Concurrently with our work, Instruct-NeRF2NeRF \cite{haque2023instruct} present a diffusion-based method for editing a NeRF scene guided by text instructions. It utilizes InstructPix2Pix \cite{brooks2022instructpix2pix}, which enables editing images based on text instructions. The edit is preformed by iteratively updating the image dataset of the original scene while training NeRF using these edited images. They demonstrate an impressive high quality local edit results on real scenes but sometimes can't preserve the rest of the scene and get a blurrier scene compared to the original, and sometimes even introduce texture and color changes to the entire scene.

SKED \cite{mikaeili2023sked} research the possibility to edit a NeRF scene using guidance from 2D sketches from different views additional to an input text prompt describing the edit. They utilize the SDS loss presented in \cite{poole2023dreamfusion} to steer the edit towards the input caption and present preservation and silhouette priors to preserve the original scene and to preform the edit only on the sketched regions. In experiments they apply their method mainly on synthetic objects and demonstrate its applicability on objects insertion and replacement tasks such as hats, flowers and glasses.

In SINE \cite{bao2023sine}, they suggest a method for editing NeRF scene by only editing a single view, and than apply the edit to the entire scene. To do this they encode the changes in geometry and texture over the original NeRF scene, by learning a prior-guided editing field. Using this field they render the modified object geometry and color and present color compositing layer supervised by the single edited view to apply the edit on novel views. They apply their method on real and synthetic scenes by changing the geometry and texture of objects in the scene.
\section{Additional Examples}
\label{sec:additional_examples}
\begin{figure*}[h]
    \centering
    \setlength{\tabcolsep}{0.5pt}
    \renewcommand{\arraystretch}{0.5}
    \setlength{\ww}{0.35\columnwidth}
  
    \begin{tabular}{cccccc}        
        \includegraphics[width=\ww,frame]{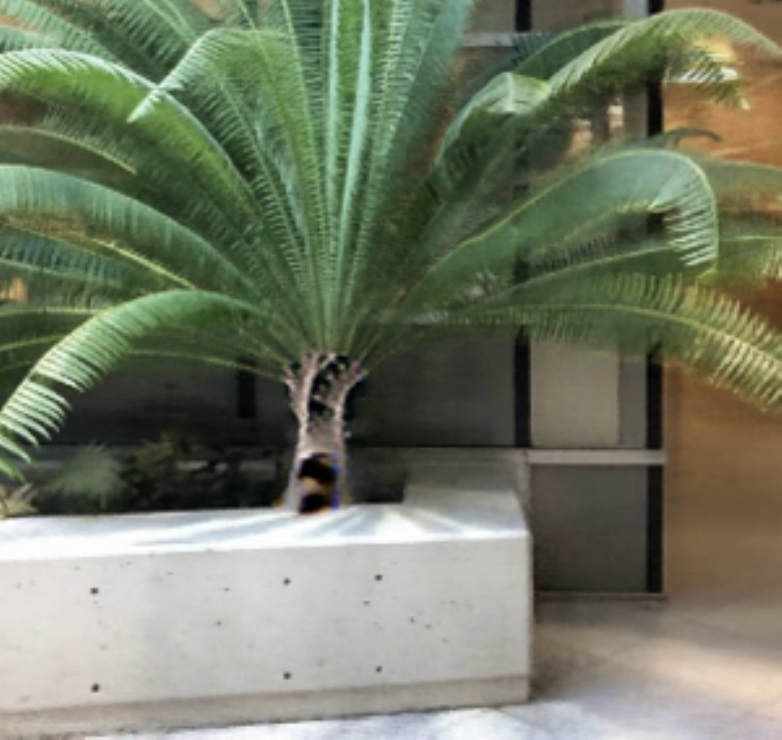} &
        \includegraphics[width=\ww,frame]{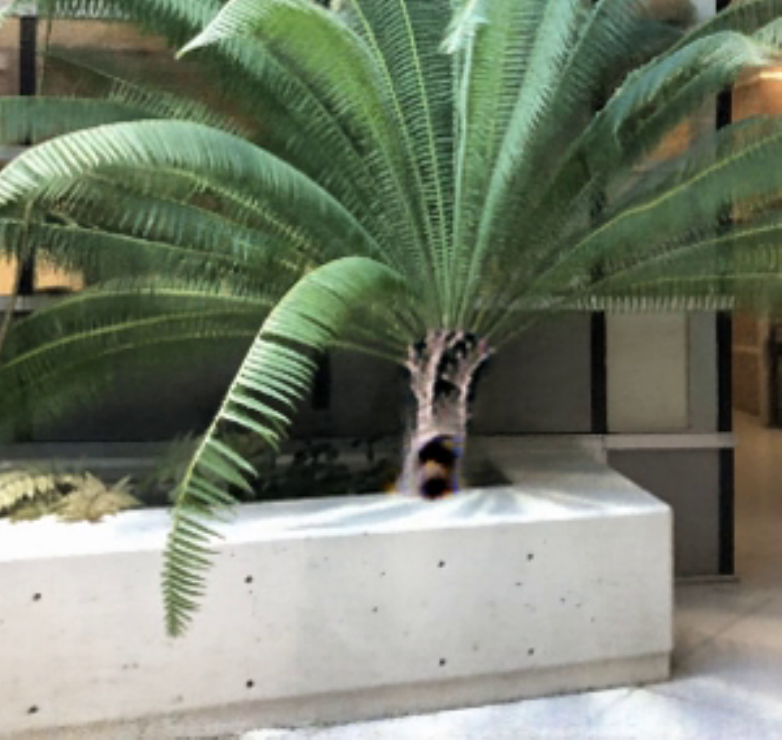} &
        \includegraphics[width=\ww,frame]{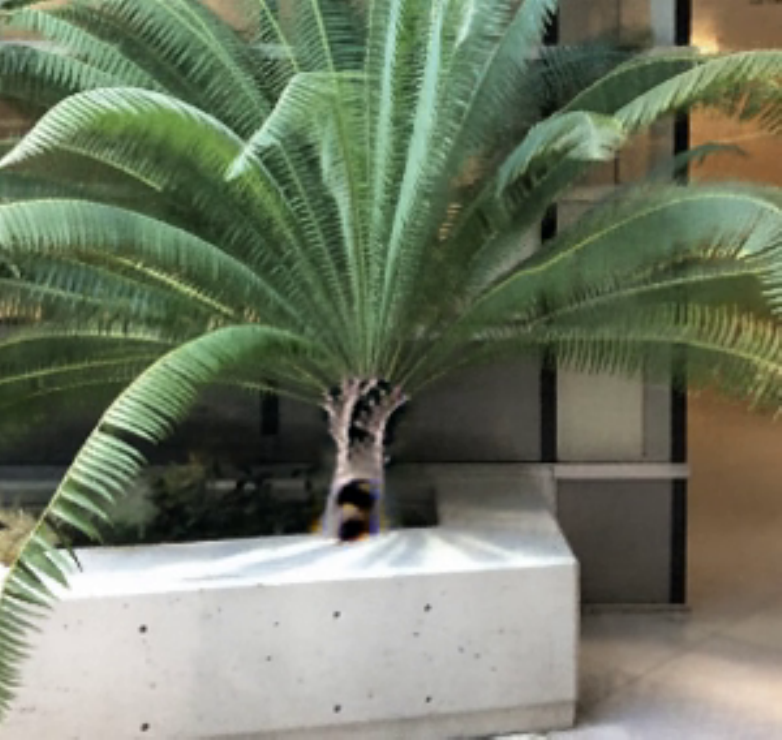}&
        \includegraphics[width=\ww,frame]{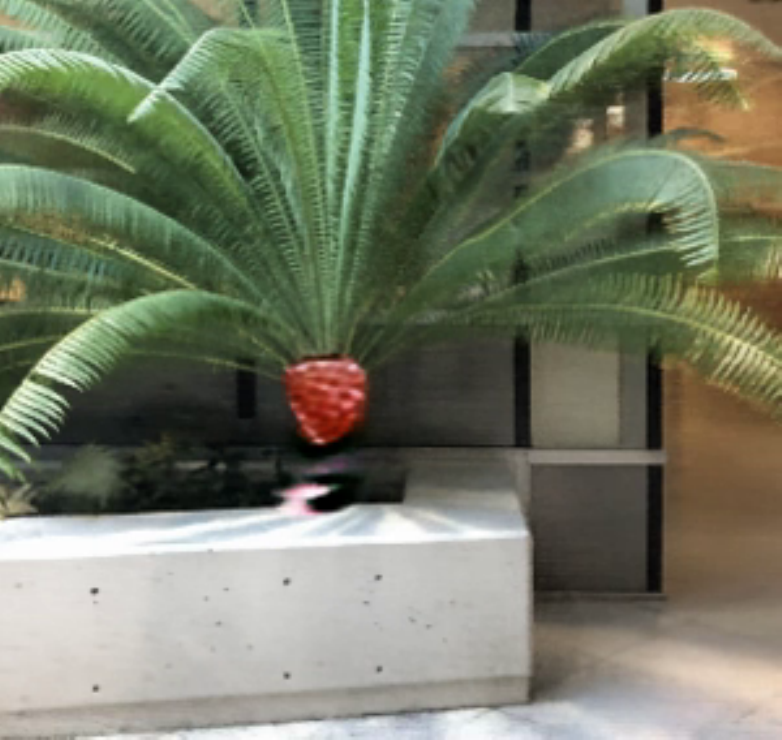} &
        \includegraphics[width=\ww,frame]{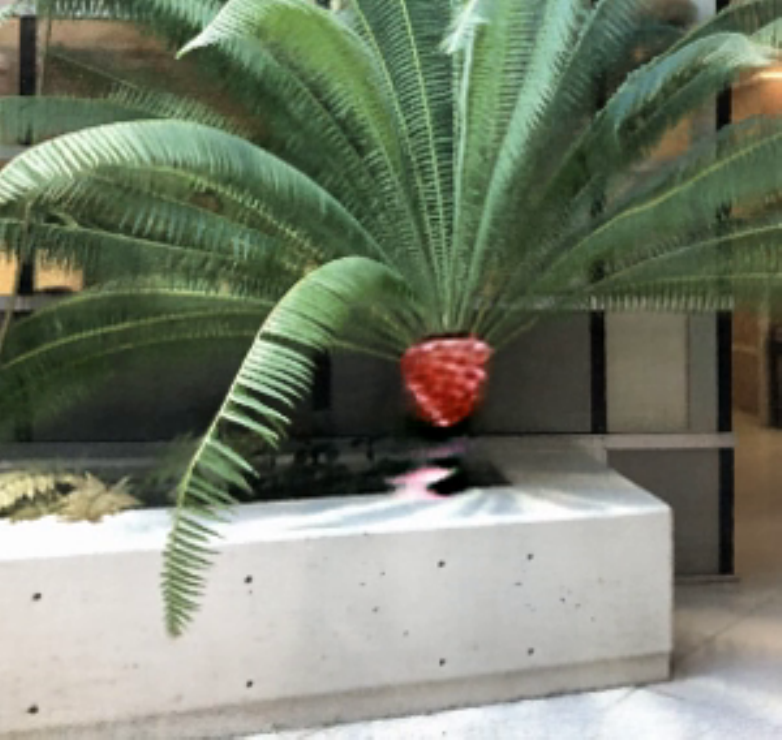} &
        \includegraphics[width=\ww,frame]{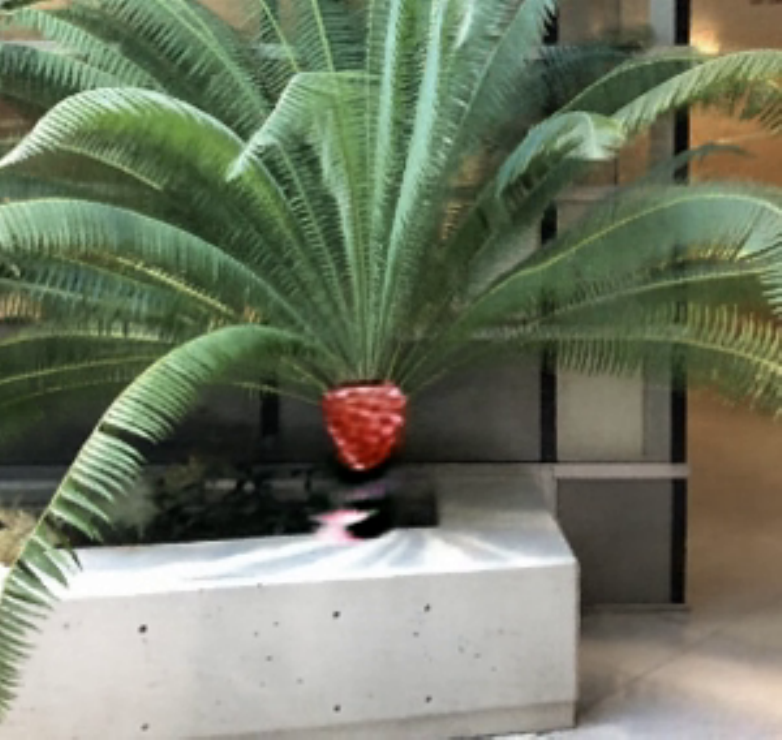}\\

        \includegraphics[width=\ww,frame]{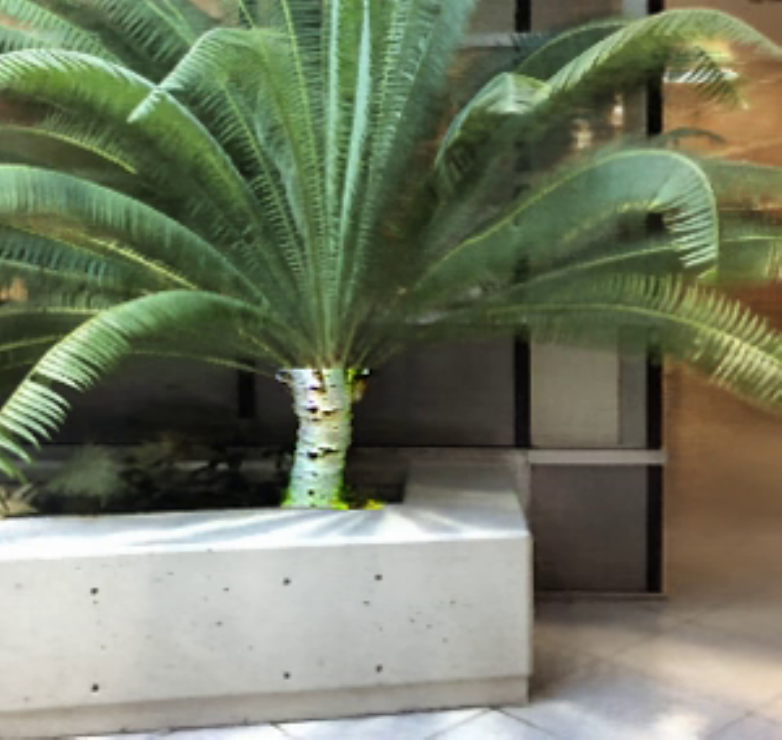}&
        \includegraphics[width=\ww,frame]{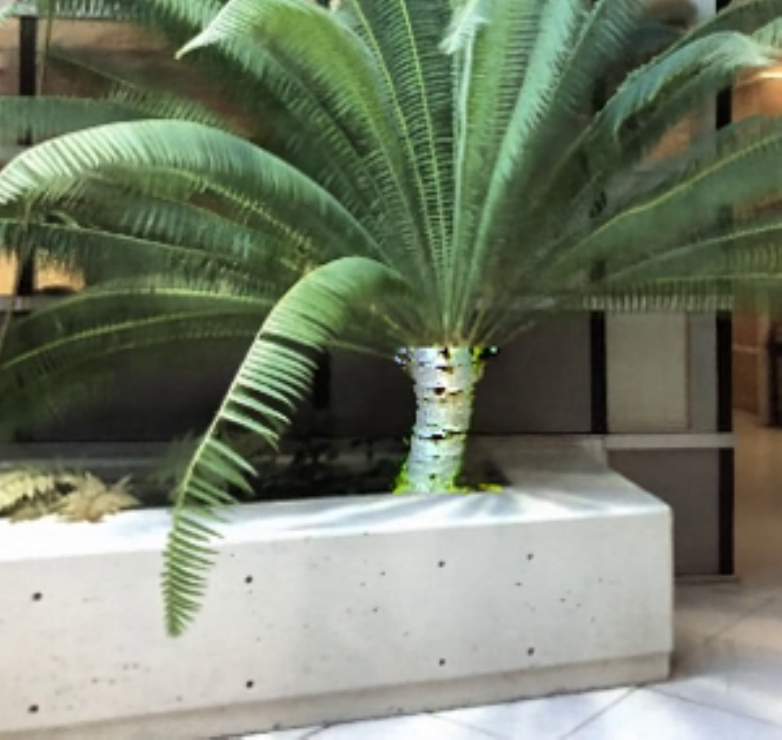}&
        \includegraphics[width=\ww,frame]{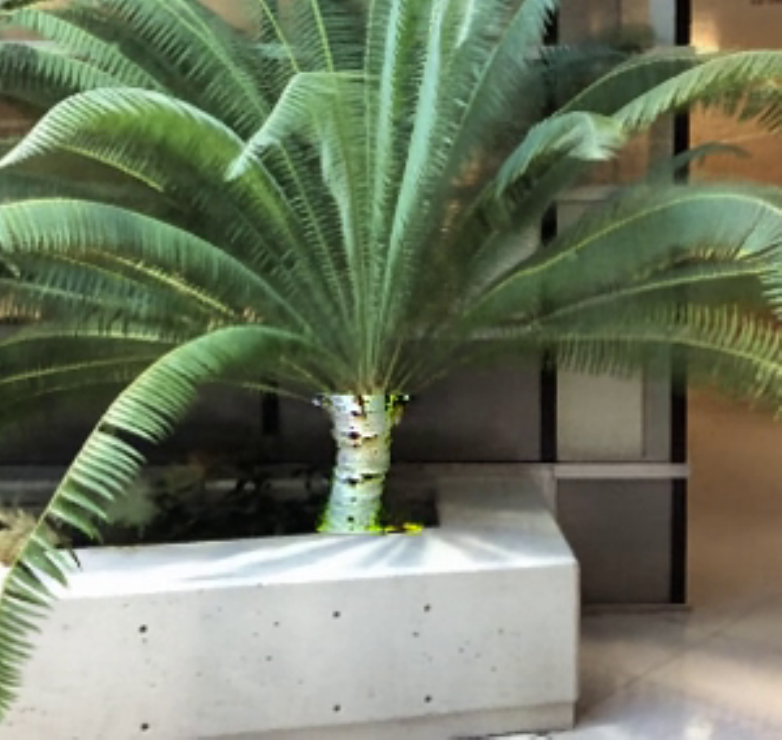}&
        \includegraphics[width=\ww,frame]{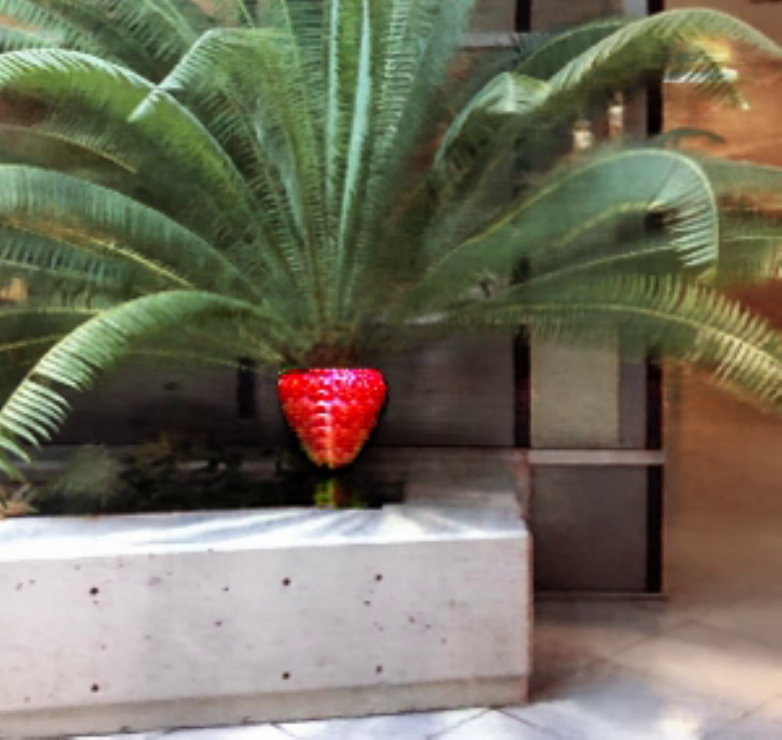}&
        \includegraphics[width=\ww,frame]{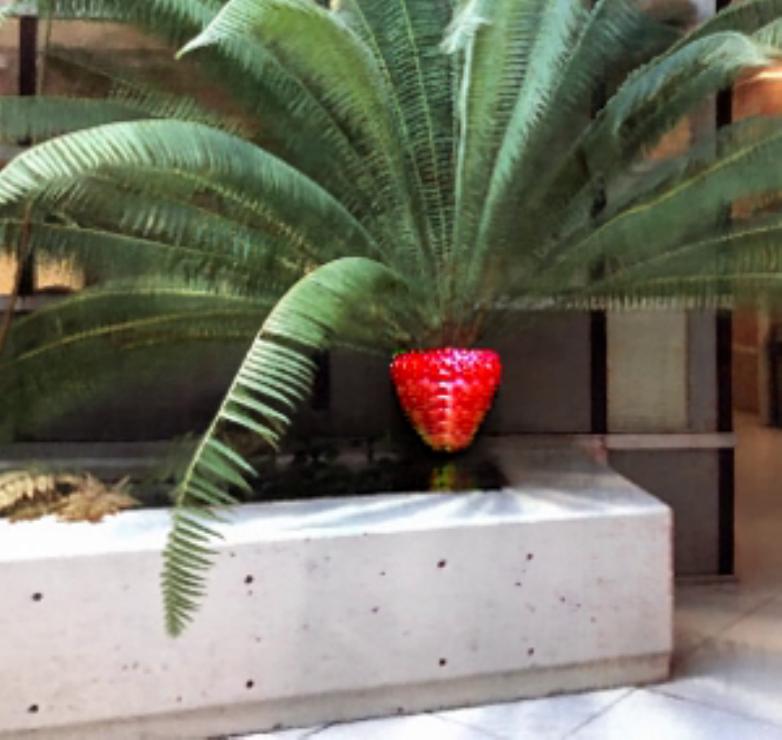}&
        \includegraphics[width=\ww,frame]{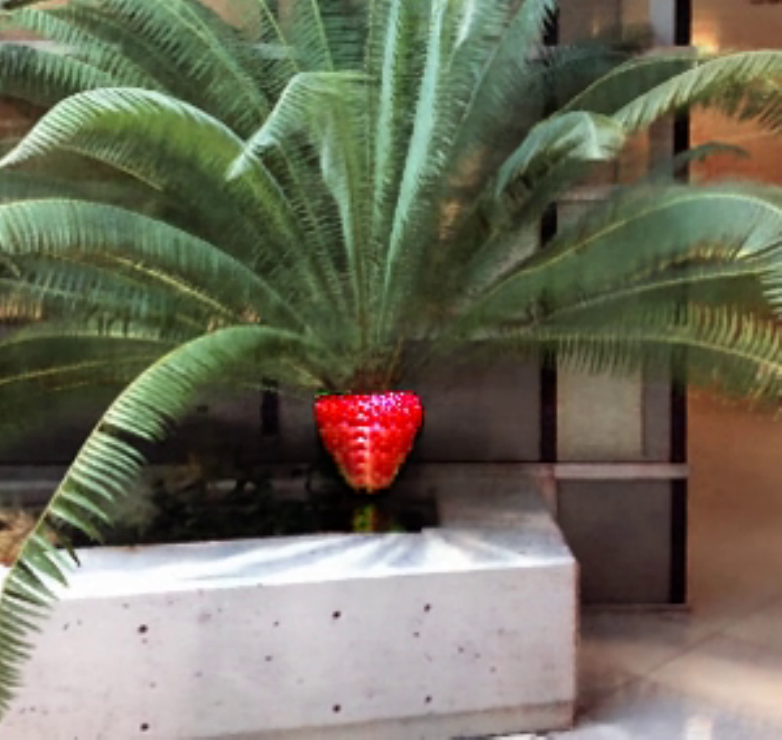}\\
		
        & \scriptsize{(a) "aspen tree"} &
        & & \scriptsize{(b) "strawberry"}
        
    \end{tabular}
    \caption{\textbf{Additional views for object replacement comparison.} Additional views for the object replacement comparison with Volumetric Disentanglement \cite{VolumeDisentanglement2022}. The first and second rows display \cite{VolumeDisentanglement2022} and our results accordingly.} \label{fig:volume_comp_views}
\end{figure*}

\begin{figure*}[h]
    \centering
    \setlength{\tabcolsep}{0.5pt}
    \renewcommand{\arraystretch}{0.5}
    \setlength{\ww}{0.42\columnwidth}
  
    \begin{tabular}{ccccc}        
        \includegraphics[width=\ww,frame]{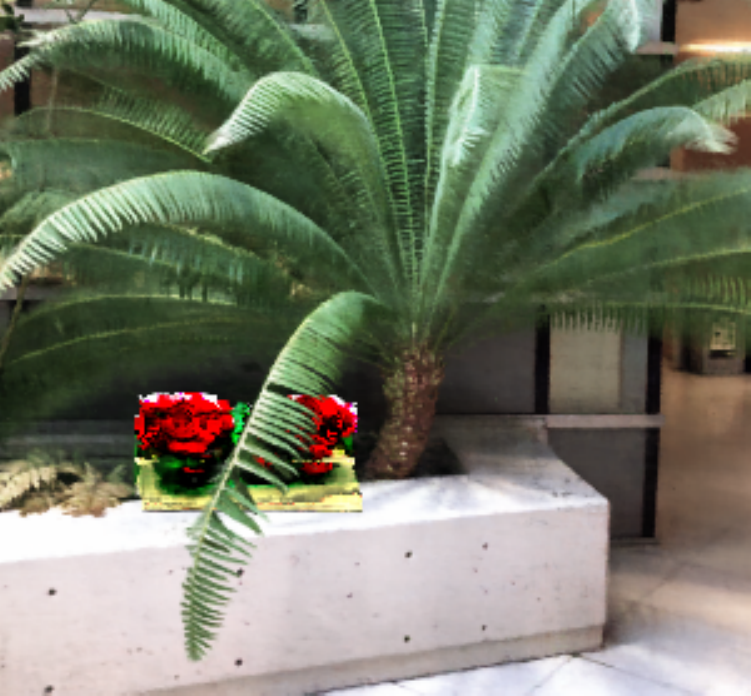} &
        \includegraphics[width=\ww,frame]{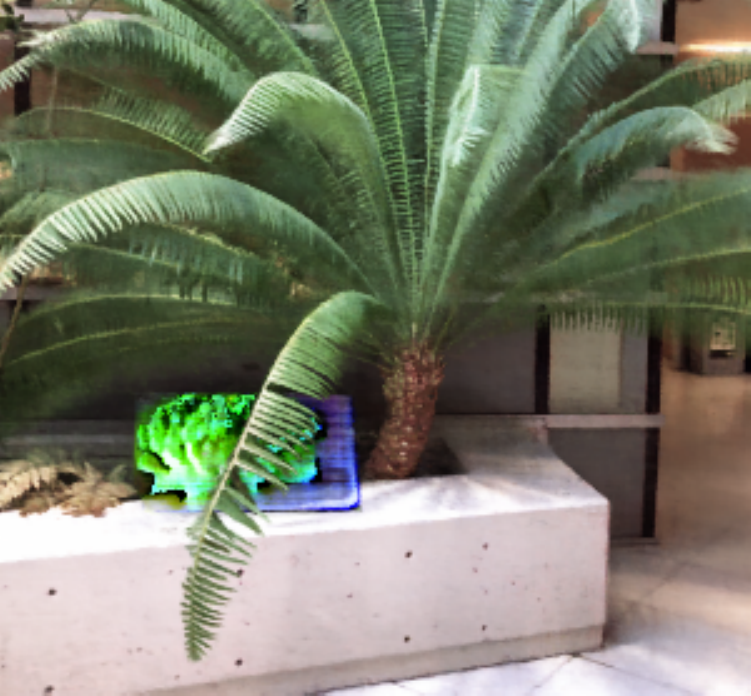} &
        \includegraphics[width=\ww,frame]{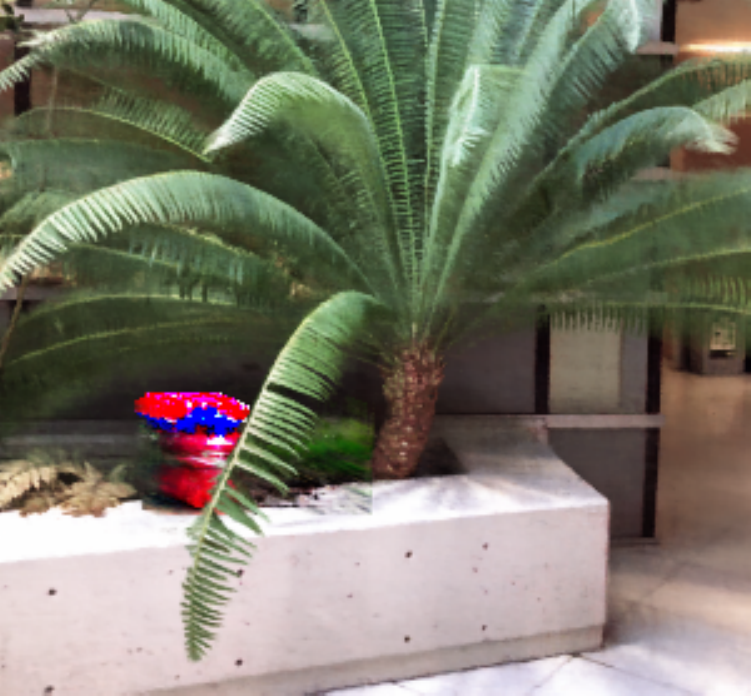}&
        \includegraphics[width=\ww,frame]{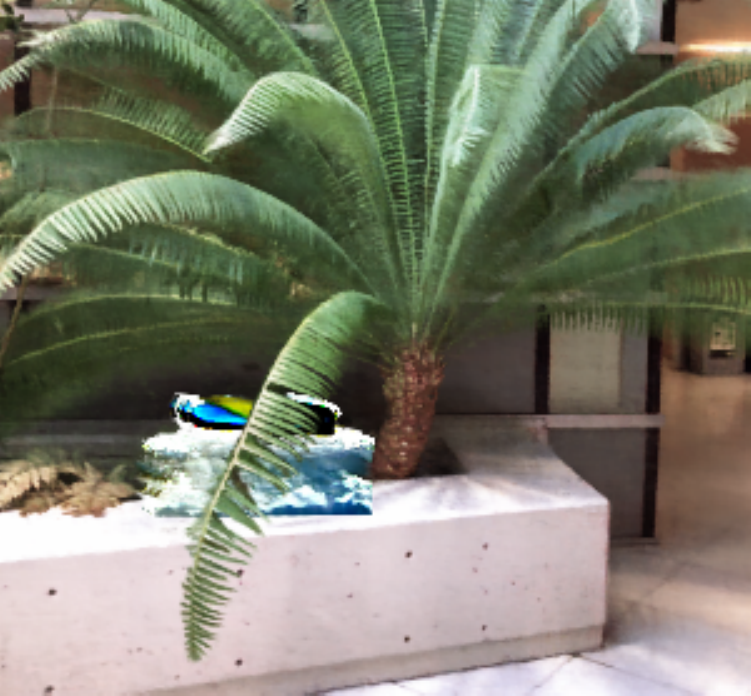}&
        \includegraphics[width=\ww,frame]{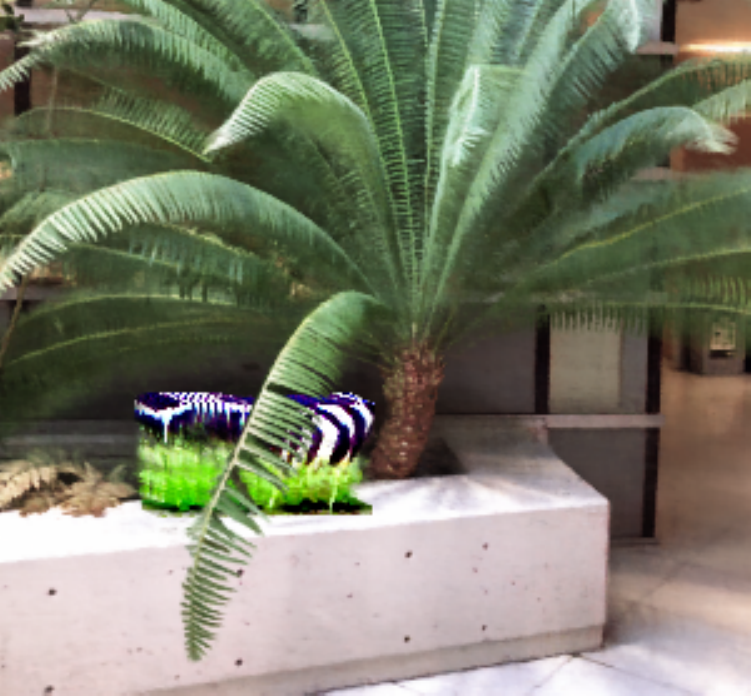}\\
        
        \includegraphics[width=\ww,frame]{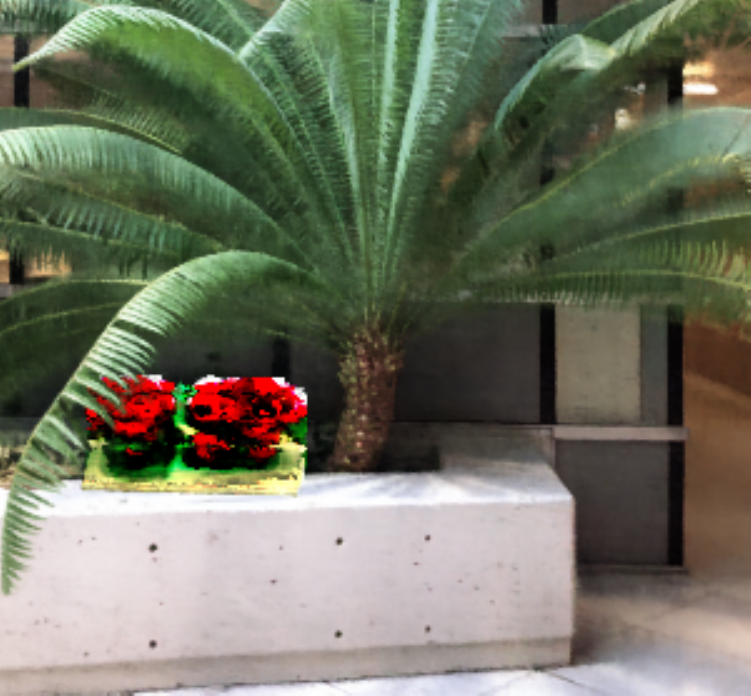} &
        \includegraphics[width=\ww,frame]{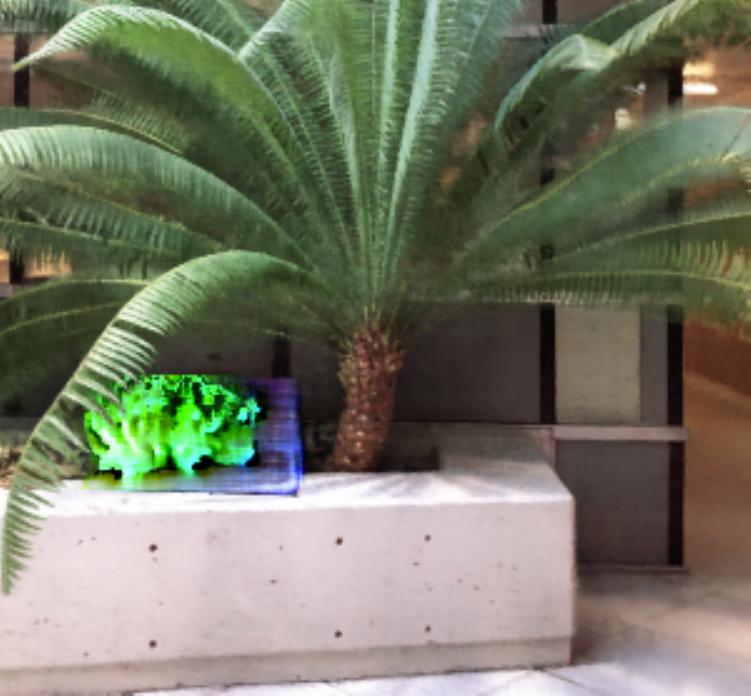} &
        \includegraphics[width=\ww,frame]{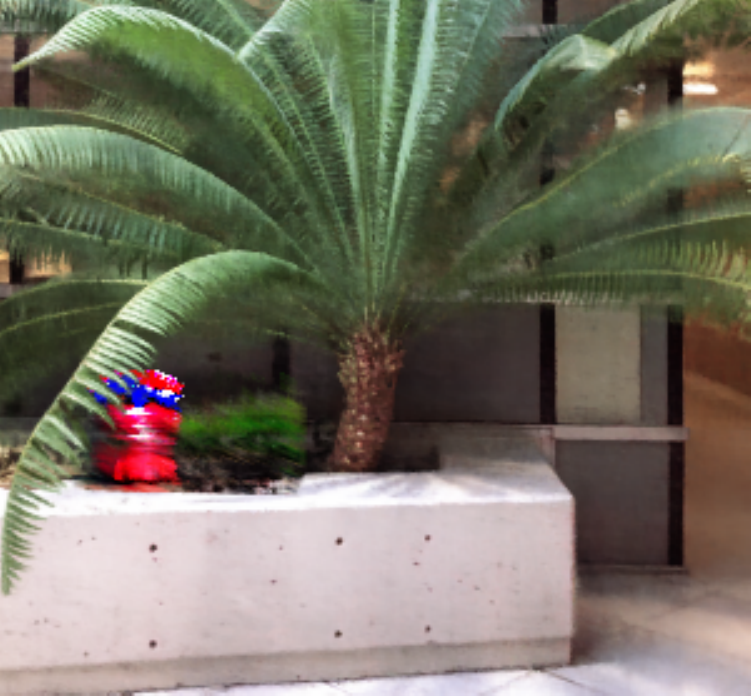}&
        \includegraphics[width=\ww,frame]{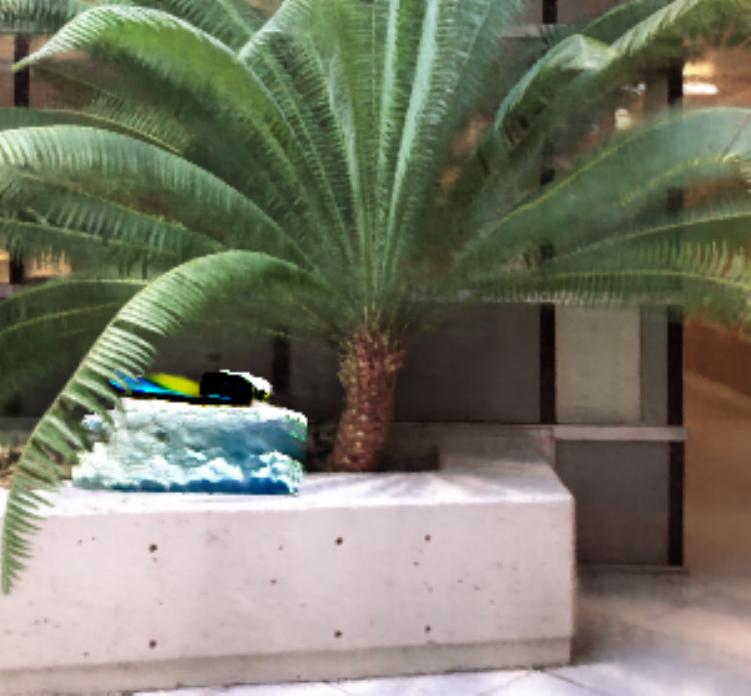}&
        \includegraphics[width=\ww,frame]{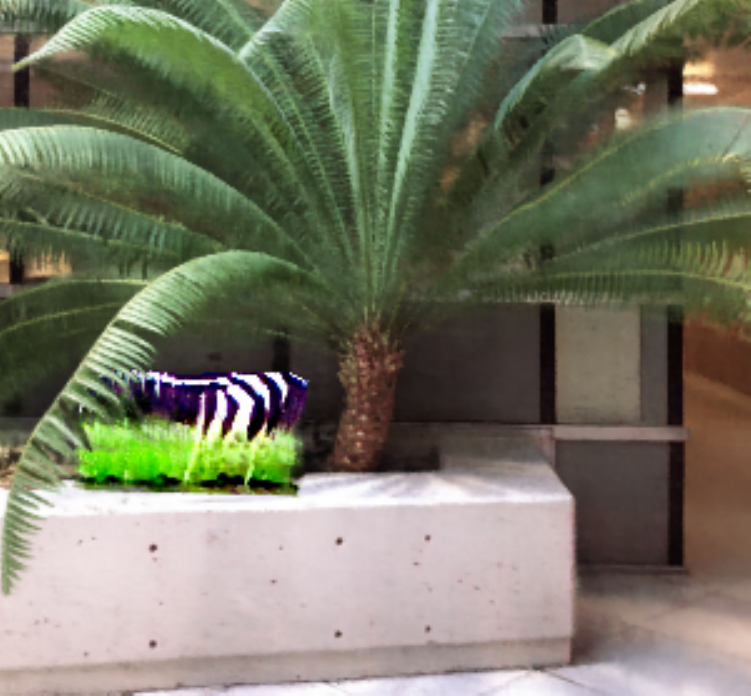}\\
		
        \scriptsize{"bouguet of wilted}  & 
        \scriptsize{"broccoli laying on} & 
        \scriptsize{"red and blue fire hydrant."} &
        \scriptsize{"snowboard standing in} &
        \scriptsize{"zebra eating grass} \\

        \scriptsize{red roses on a table."} & 
        \scriptsize{on a plastic board."} & 
        &
        \scriptsize{a snow bank."} &
        \scriptsize{on the ground."} \\
        
    \end{tabular}
    \caption{\textbf{Object Insertion.} Insertion of new objects from COCO dataset \cite{lin2014microsoftcoco} into an empty region in fern llff scene. Each column shows two views of the same edited scene \cite{mildenhall2019llff}.} \label{fig:coco_object_insertion}
\end{figure*}

\begin{figure*}[h] 
    \centering
    \setlength{\tabcolsep}{0.5pt}
    \renewcommand{\arraystretch}{0.5}
    \setlength{\ww}{0.6\columnwidth}
    \subfloat[original scene.]
    {
    \includegraphics[width=\ww,frame]{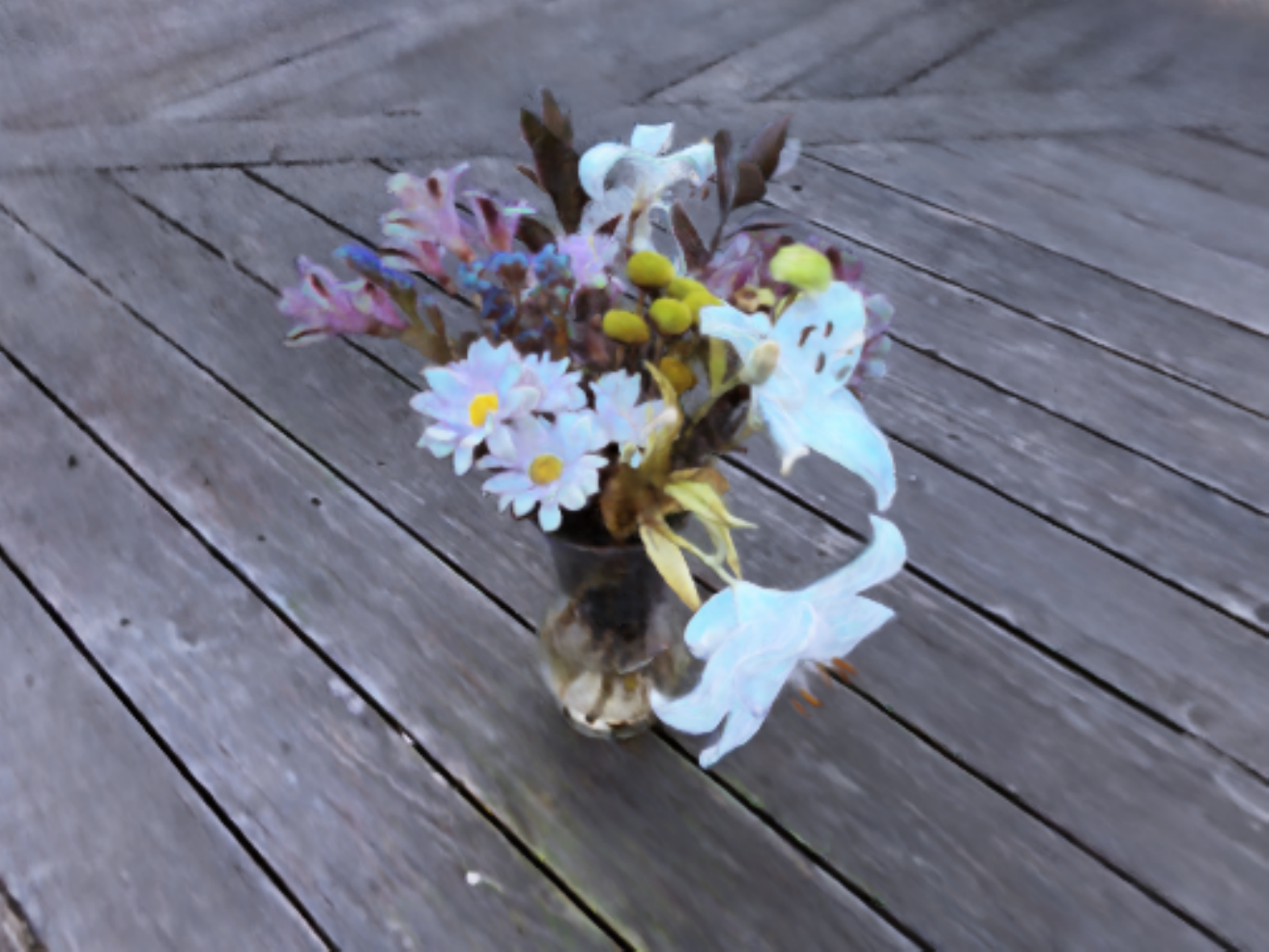}
    \hspace{0.0001\textwidth}

    \includegraphics[width=\ww,frame]{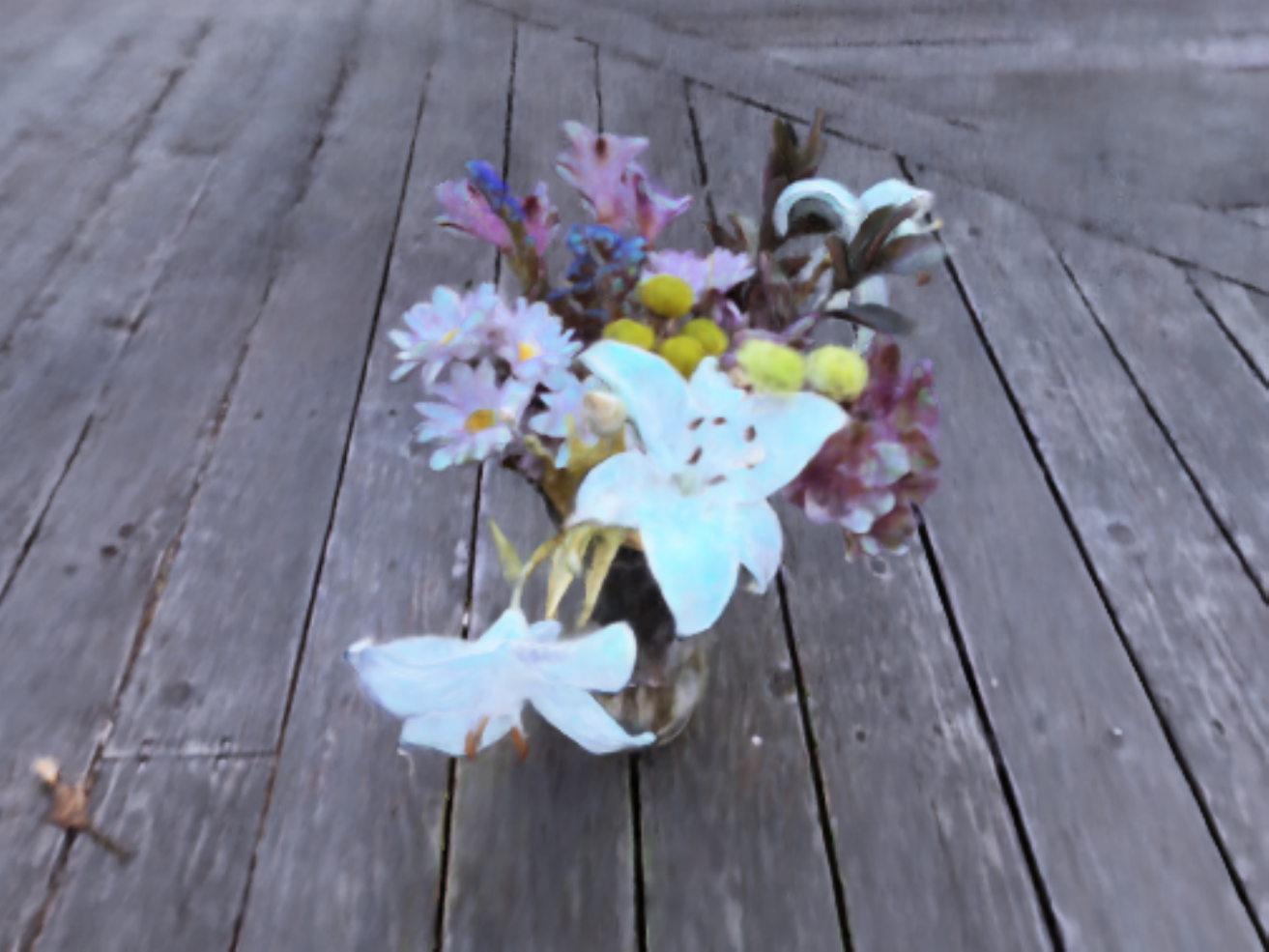}
    \hspace{0.0001\textwidth}
    
    \includegraphics[width=\ww,frame]{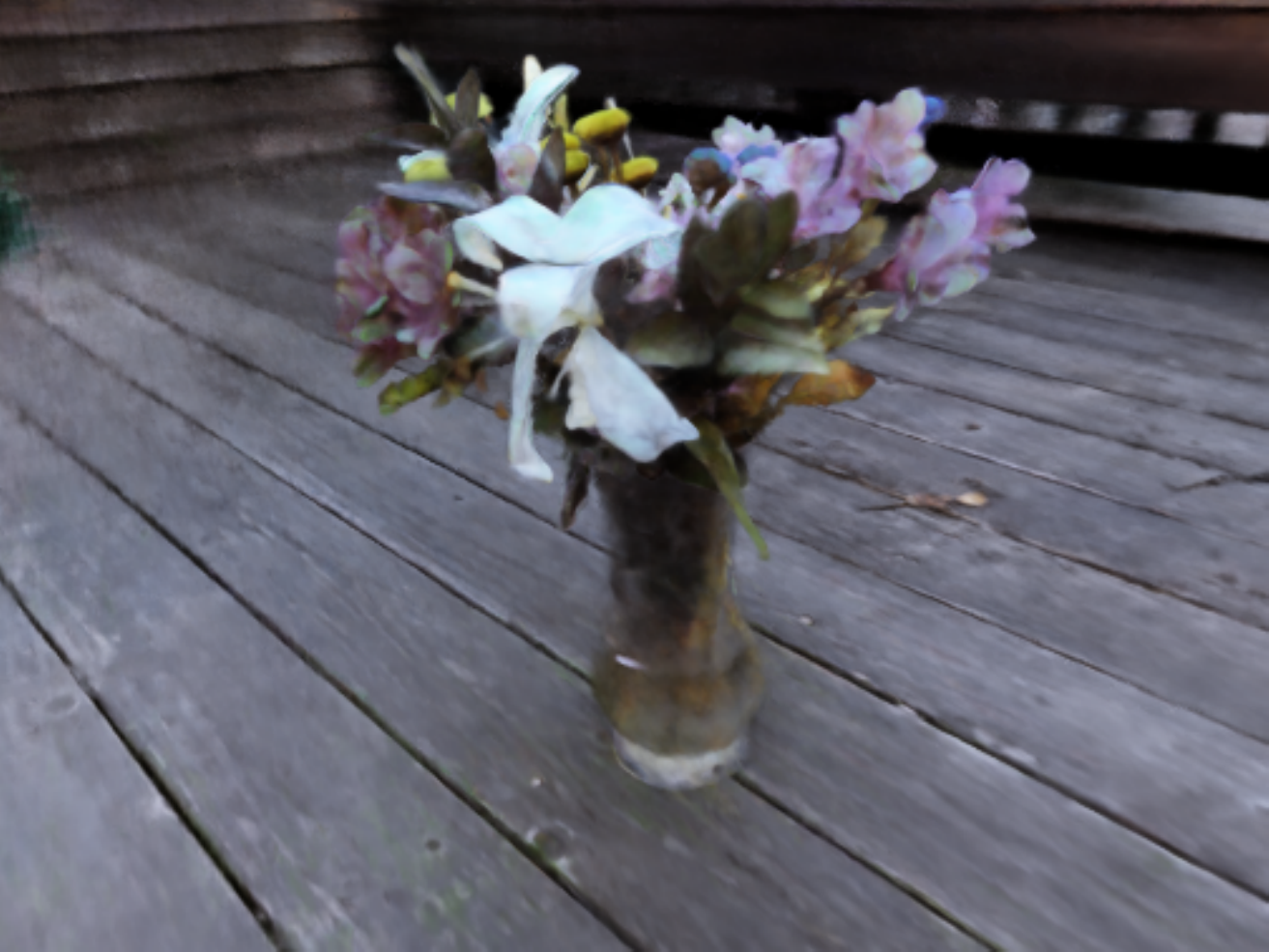}

    }  \label{fig:vase_original}
    \vspace{0.0001\paperheight}
    
    \subfloat[edited scene.]
    {
    \includegraphics[width=\ww,frame]{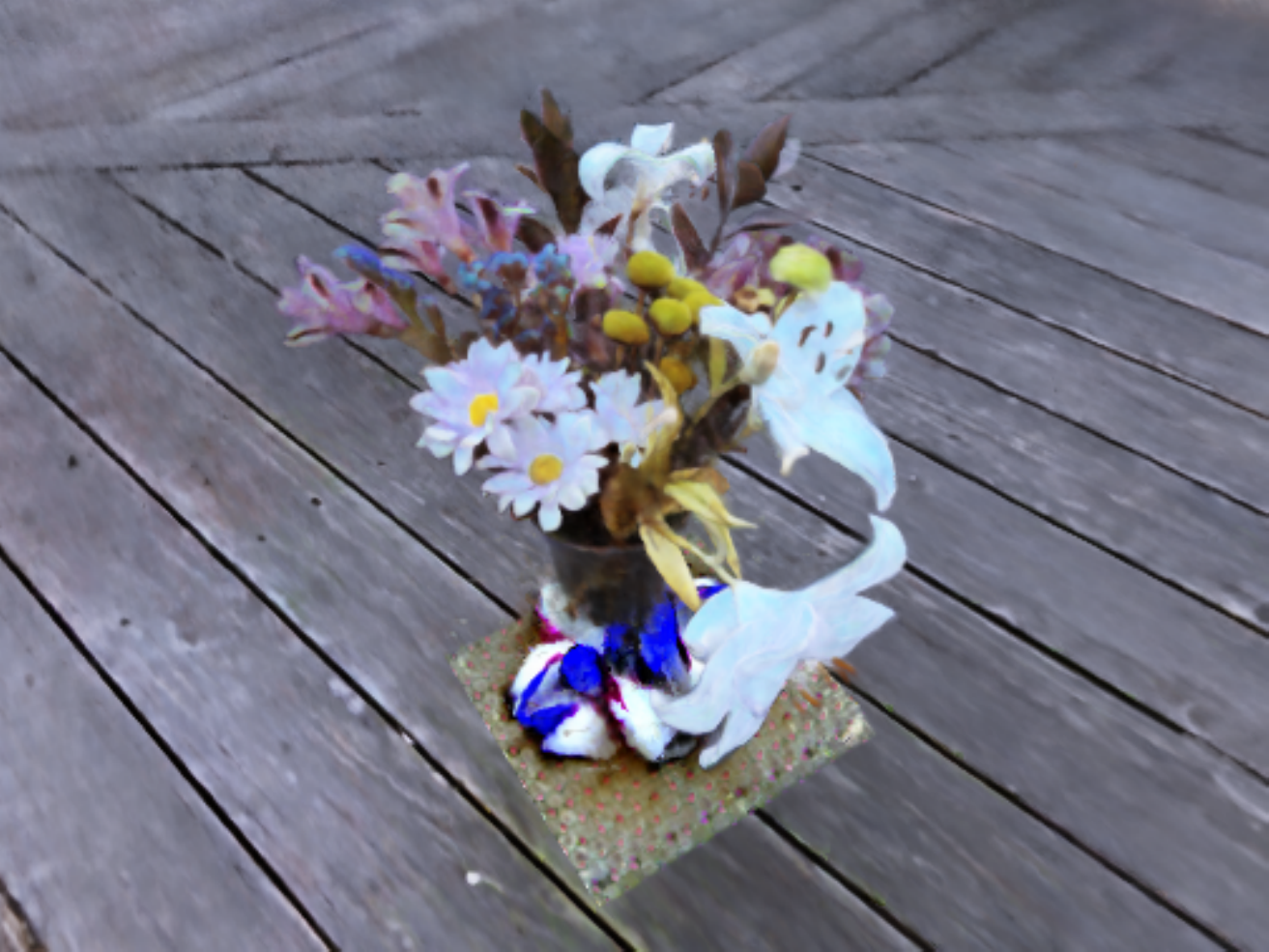}
    \hspace{0.0001\textwidth}

    \includegraphics[width=\ww,frame]{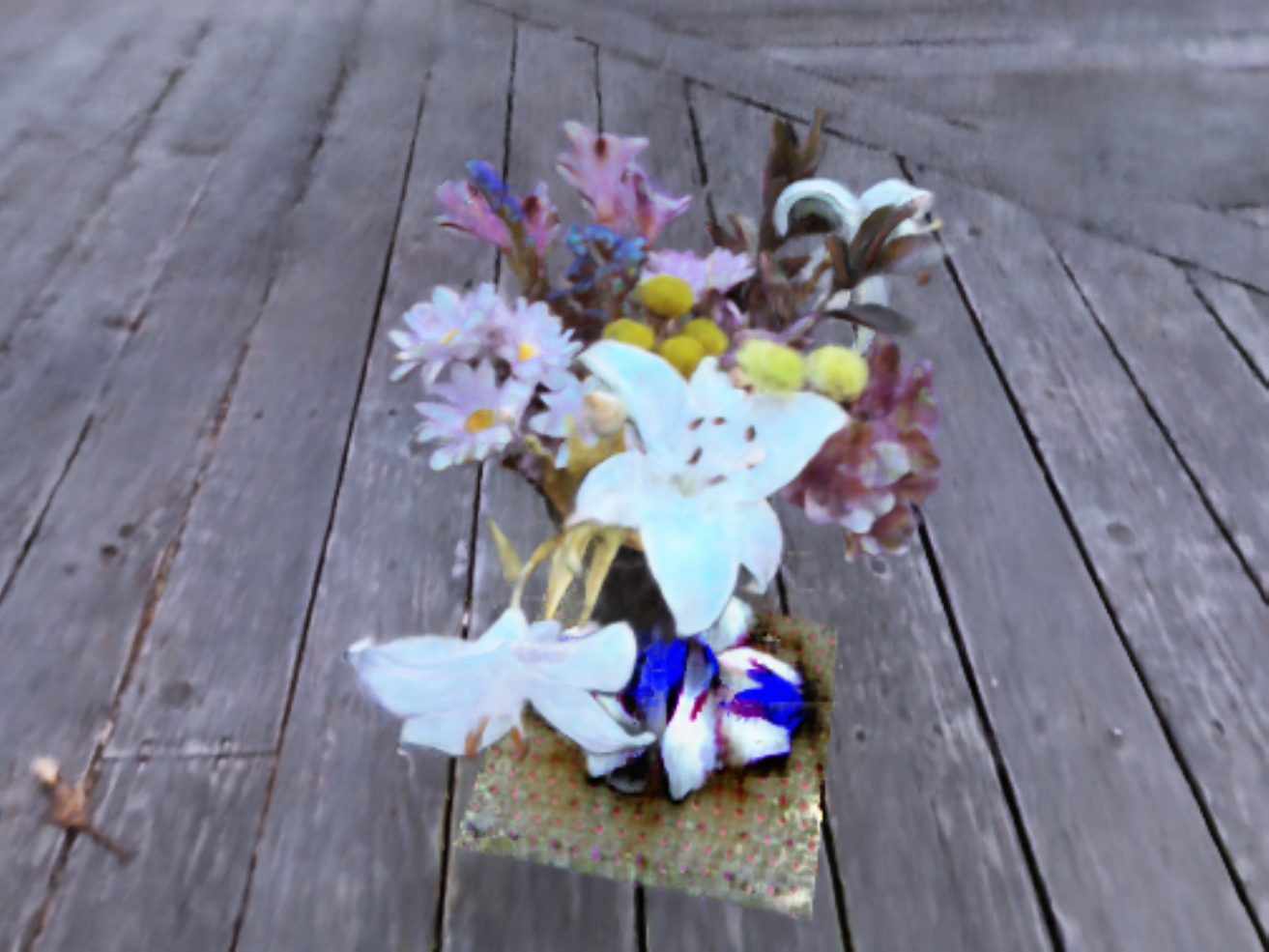}
    \hspace{0.0001\textwidth}
    
    \includegraphics[width=\ww,frame]{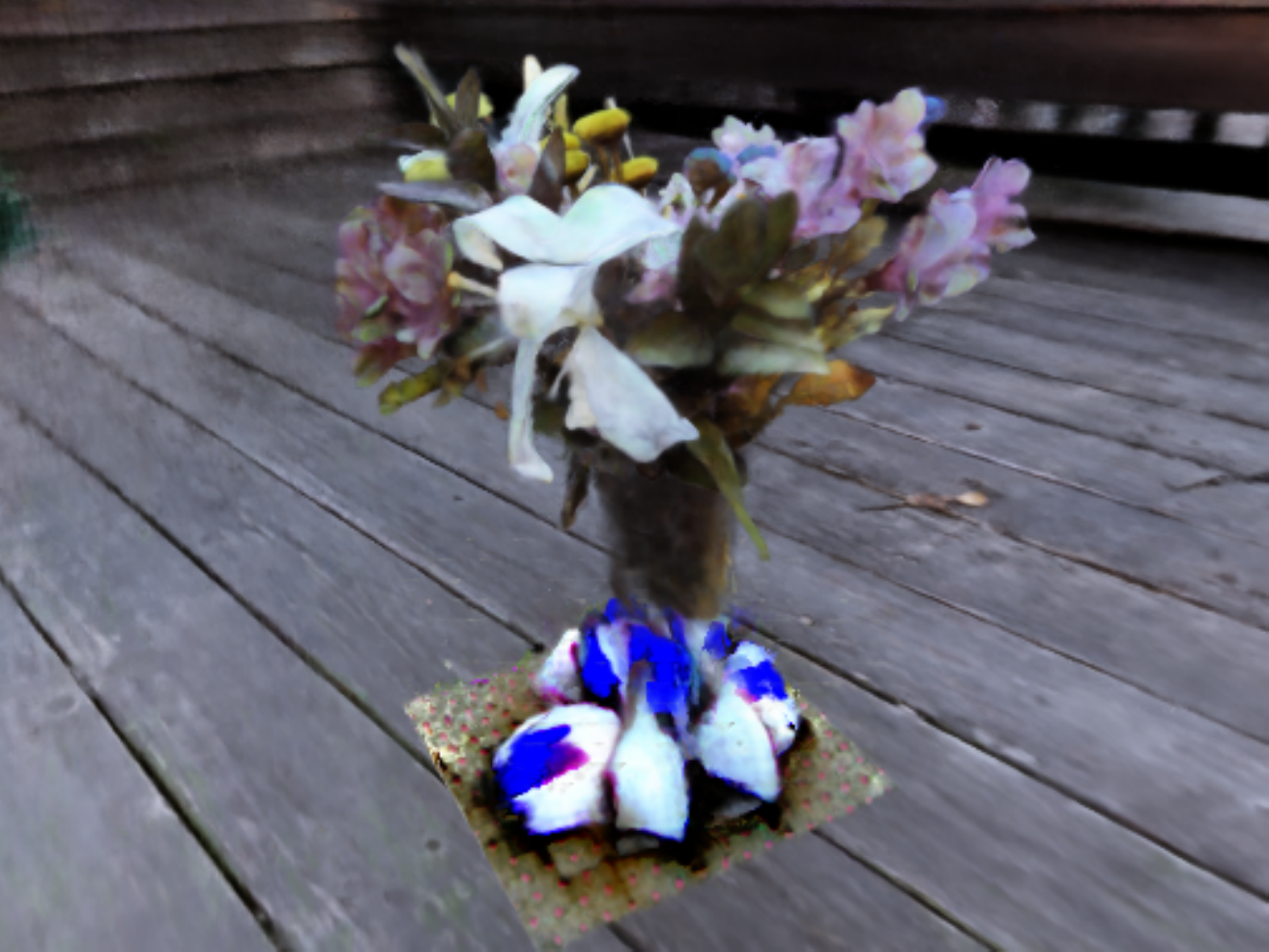}
    }  \label{fig:vase_edited}
    \vspace{0.0001\paperheight}

    \caption{\textbf{Object Insertion in vasedeck 360 scene.} We used the text: "a photo of a purple, white and blue flowers petals on the ground" and eq. (5) with $\alpha=3.5$ to generate the edit.} \label{fig:vasedeck_flower_petals}
\end{figure*}

\begin{figure*}[h] 
    \centering
    \setlength{\tabcolsep}{0.5pt}
    \renewcommand{\arraystretch}{0.5}
    \setlength{\ww}{0.5\columnwidth}
    \subfloat[original scene.]
    {
    \includegraphics[width=\ww,frame]{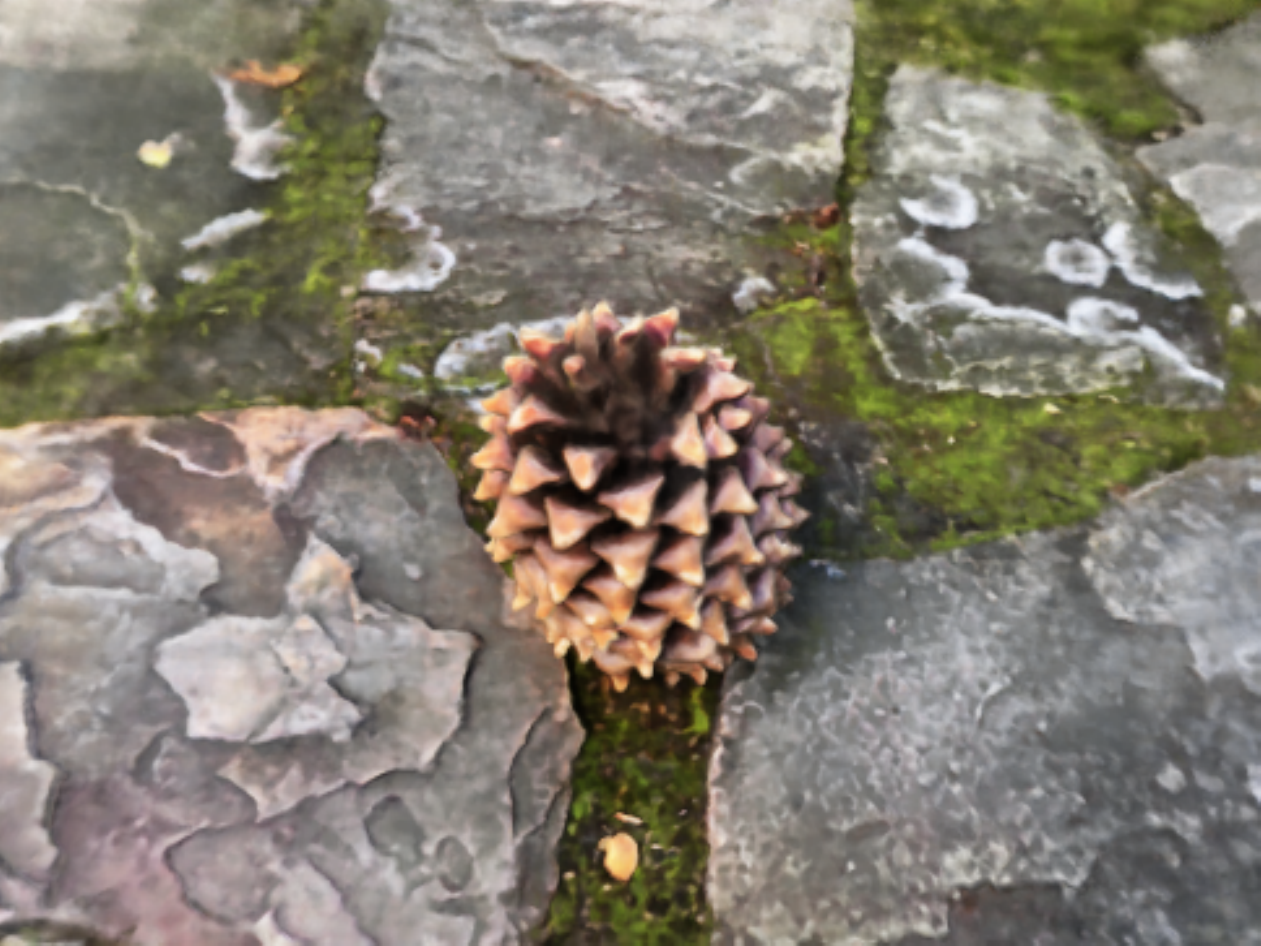}
    \hspace{0.001\textwidth}

    \includegraphics[width=\ww,frame]{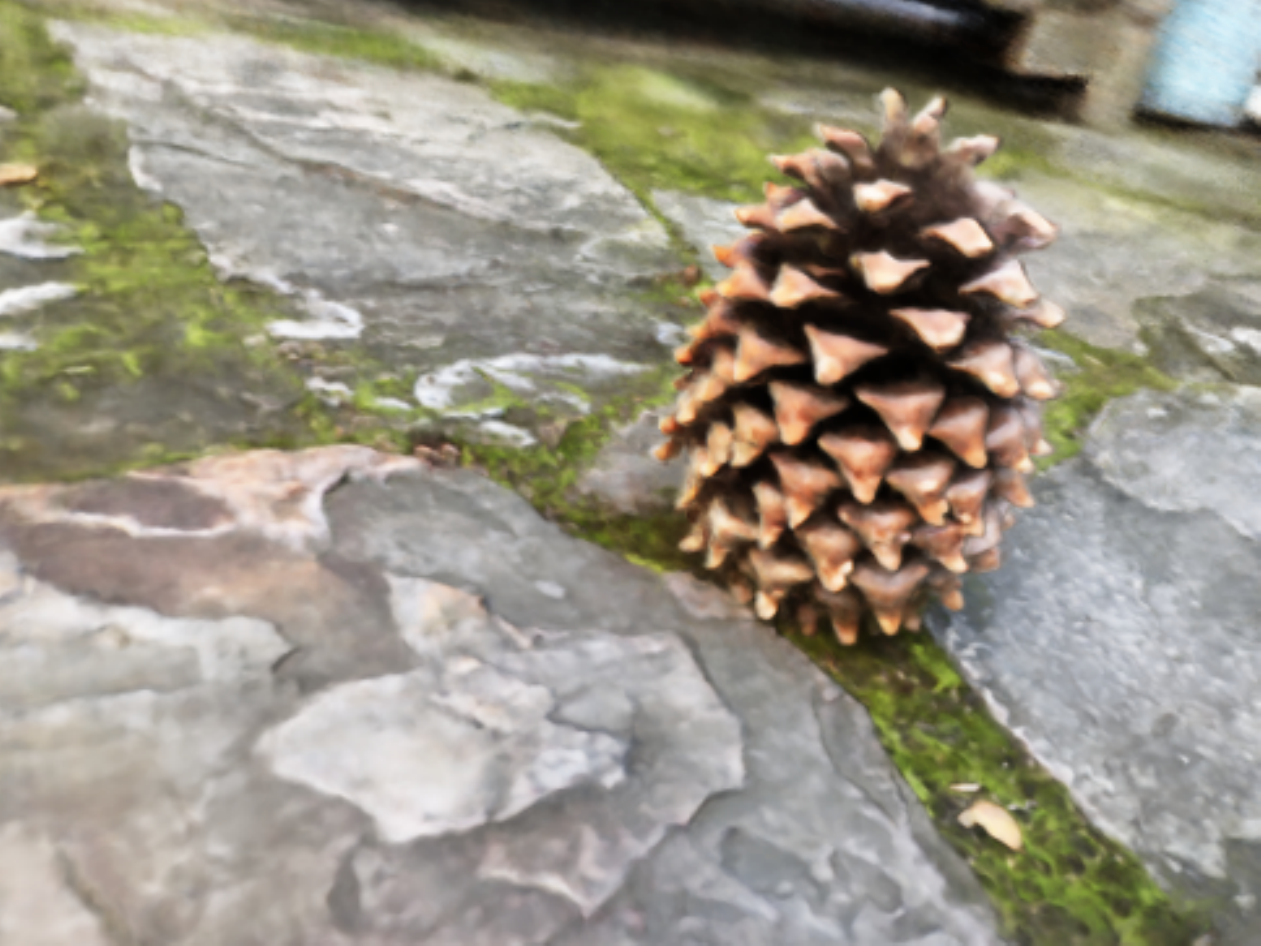}
    \hspace{0.001\textwidth}
    
    \includegraphics[width=\ww,frame]{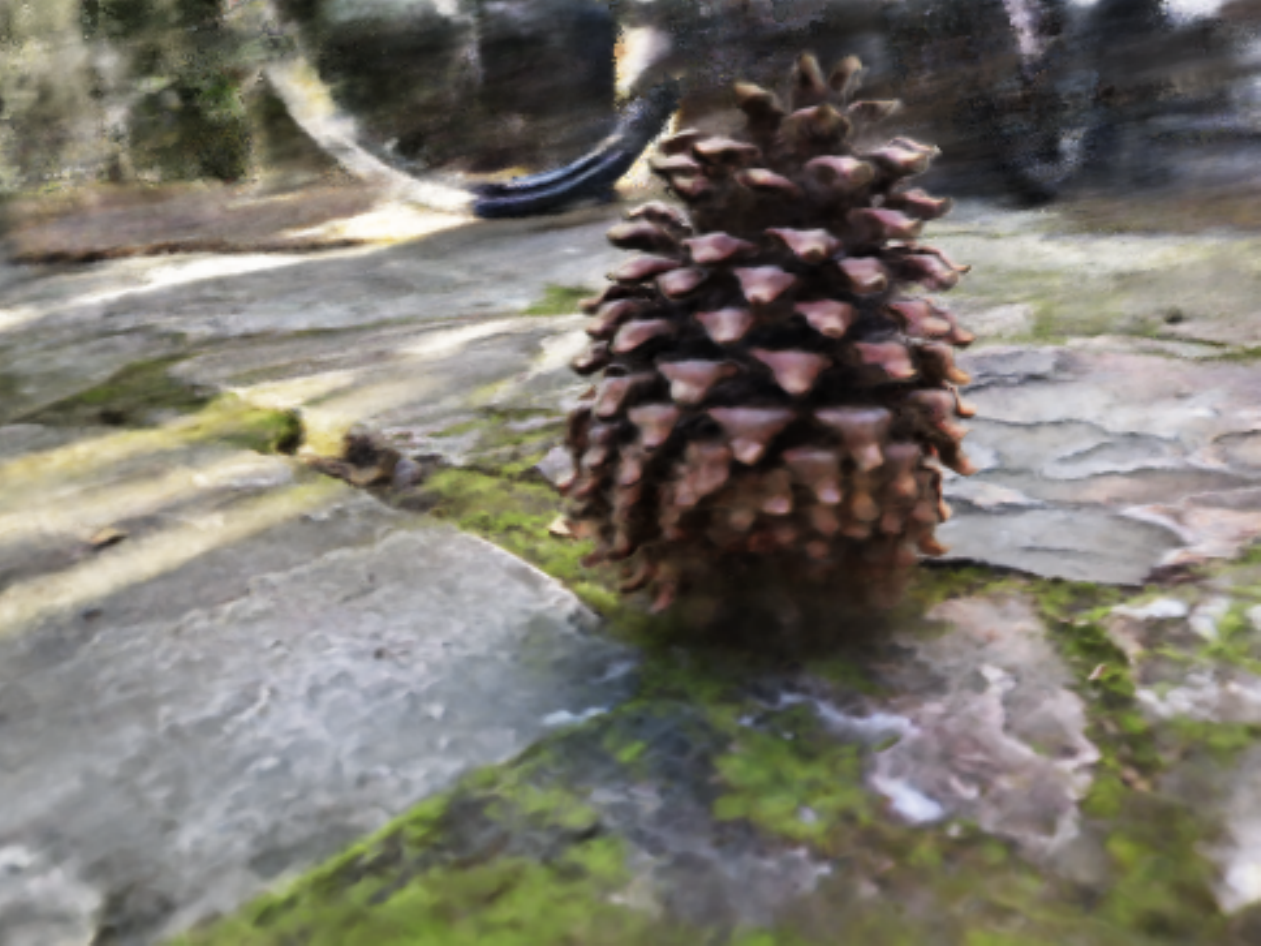}
    \hspace{0.001\textwidth}

    \includegraphics[width=\ww,frame]{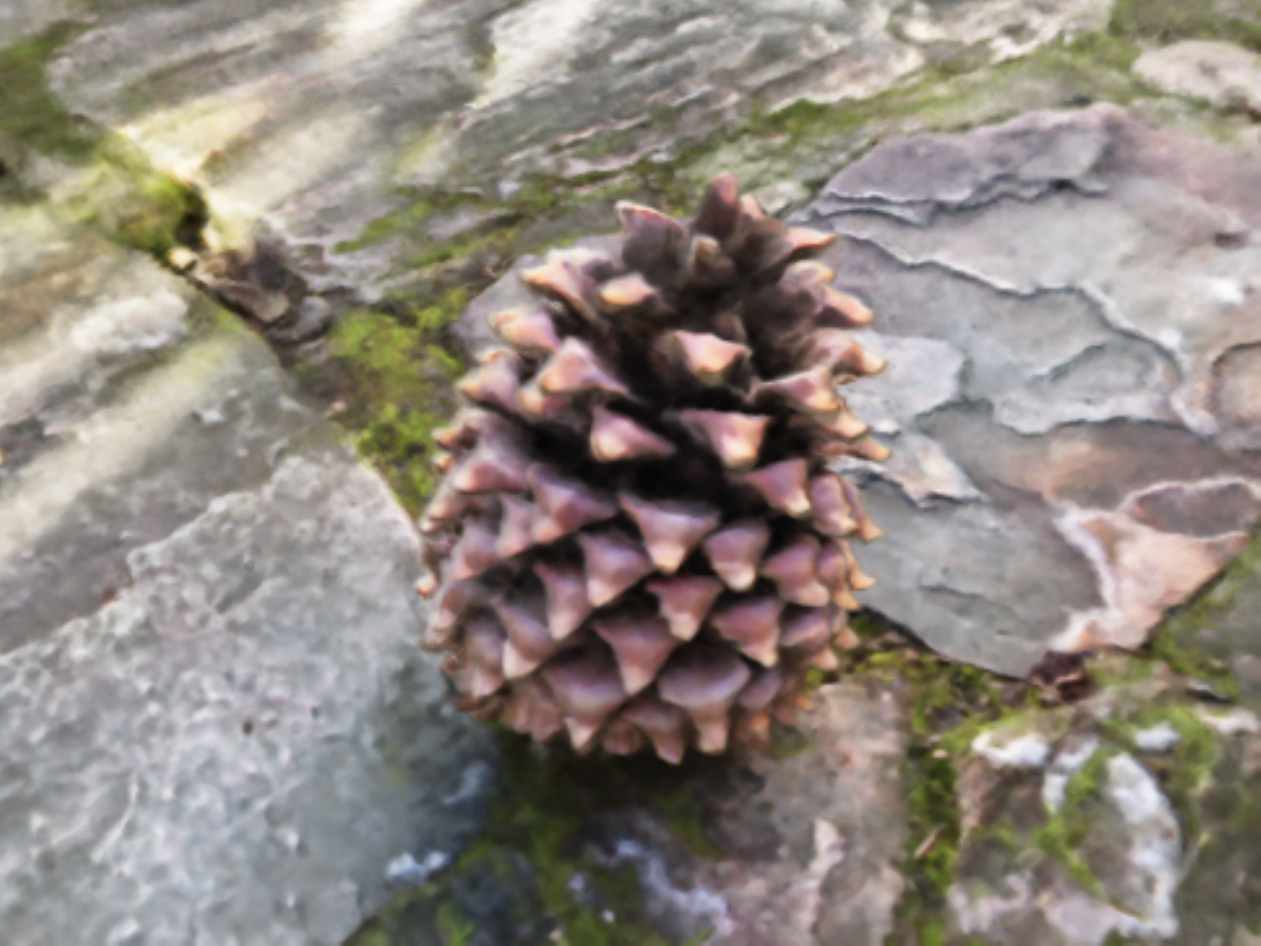}
    } 
    \vspace{0.0001\paperheight}
    
    \subfloat[``a pineapple.``]
    {
    \includegraphics[width=\ww,frame]{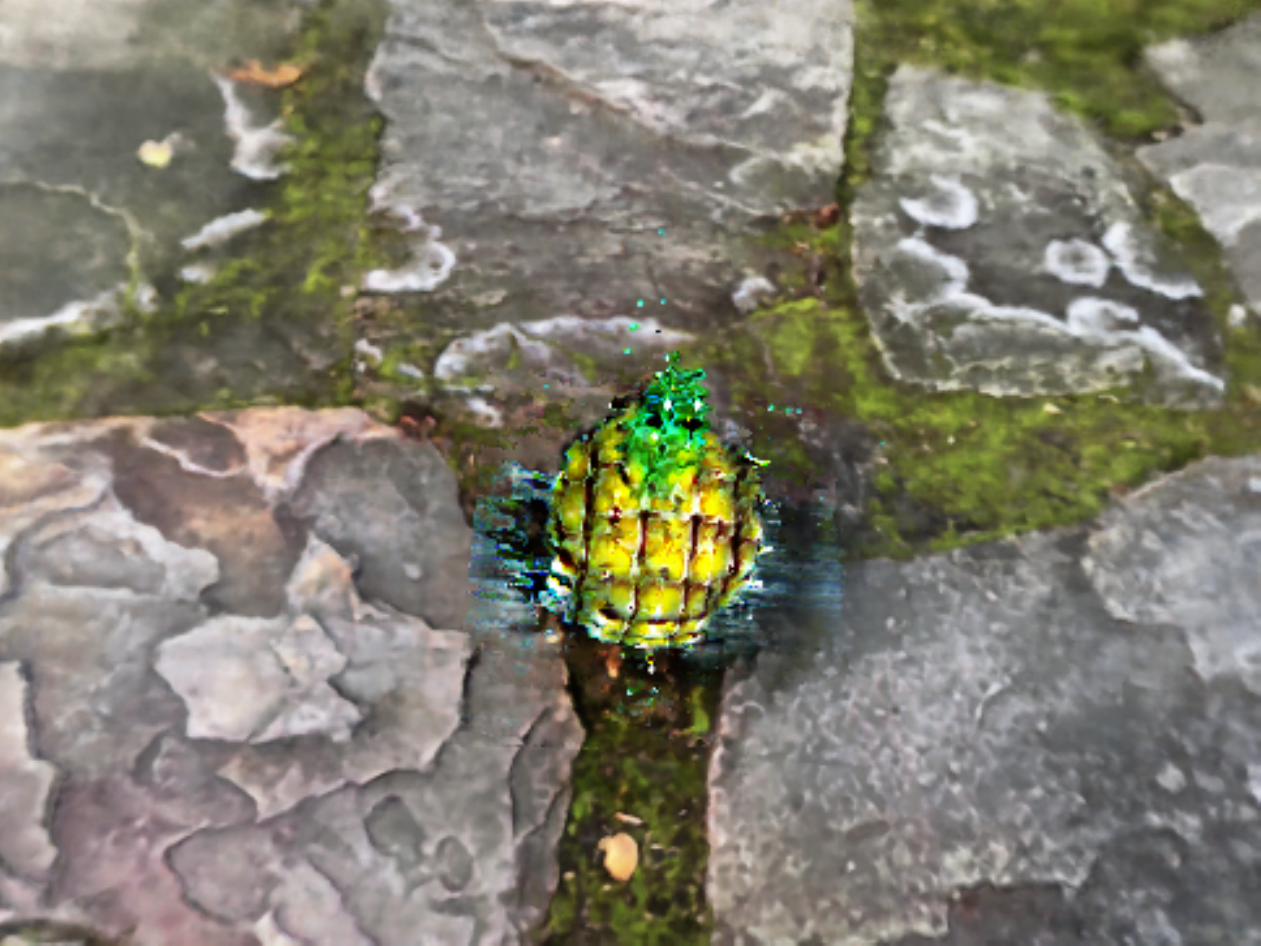}
    \hspace{0.001\textwidth}

    \includegraphics[width=\ww,frame]{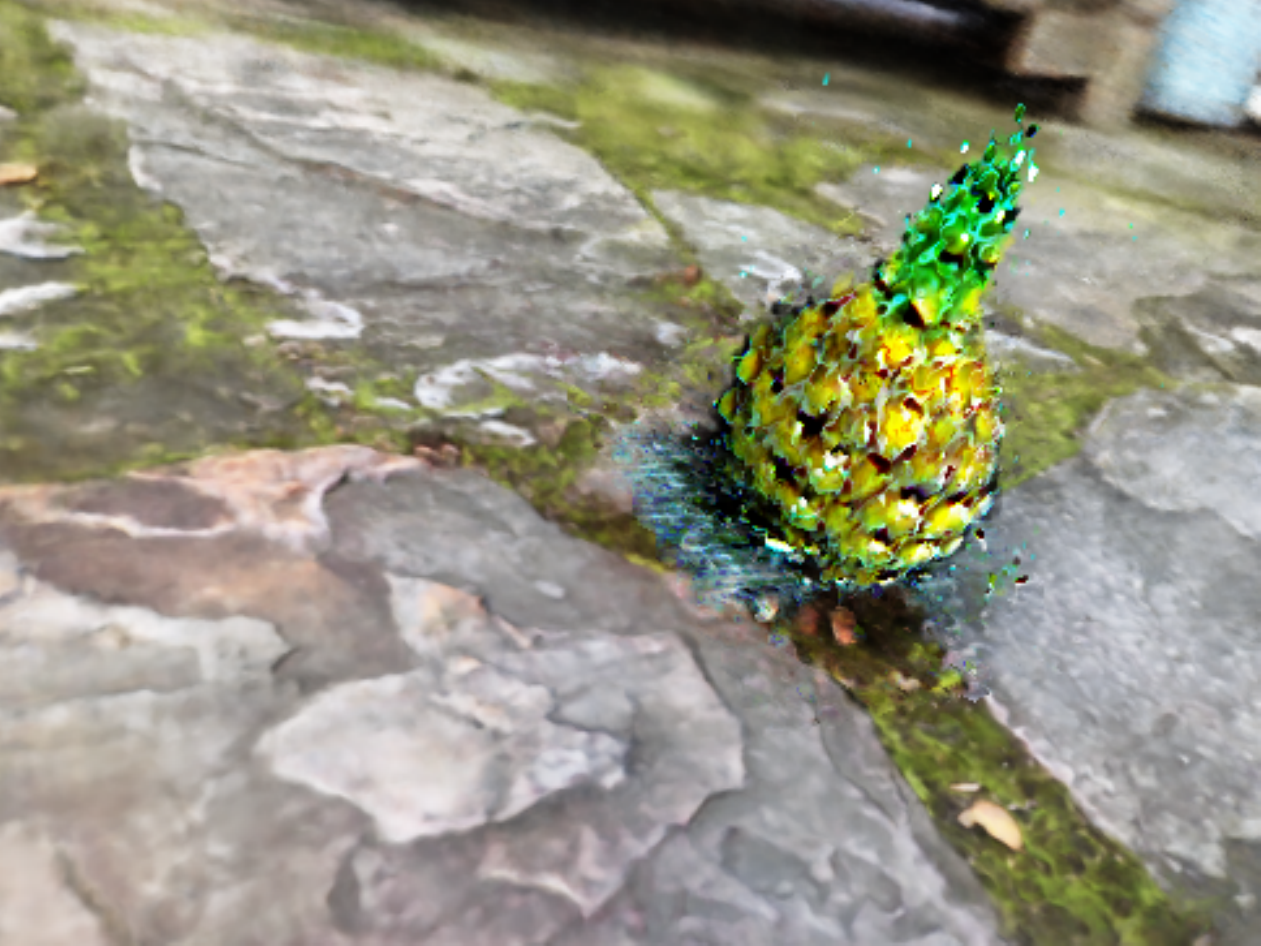}
    \hspace{0.001\textwidth}
    
    \includegraphics[width=\ww,frame]{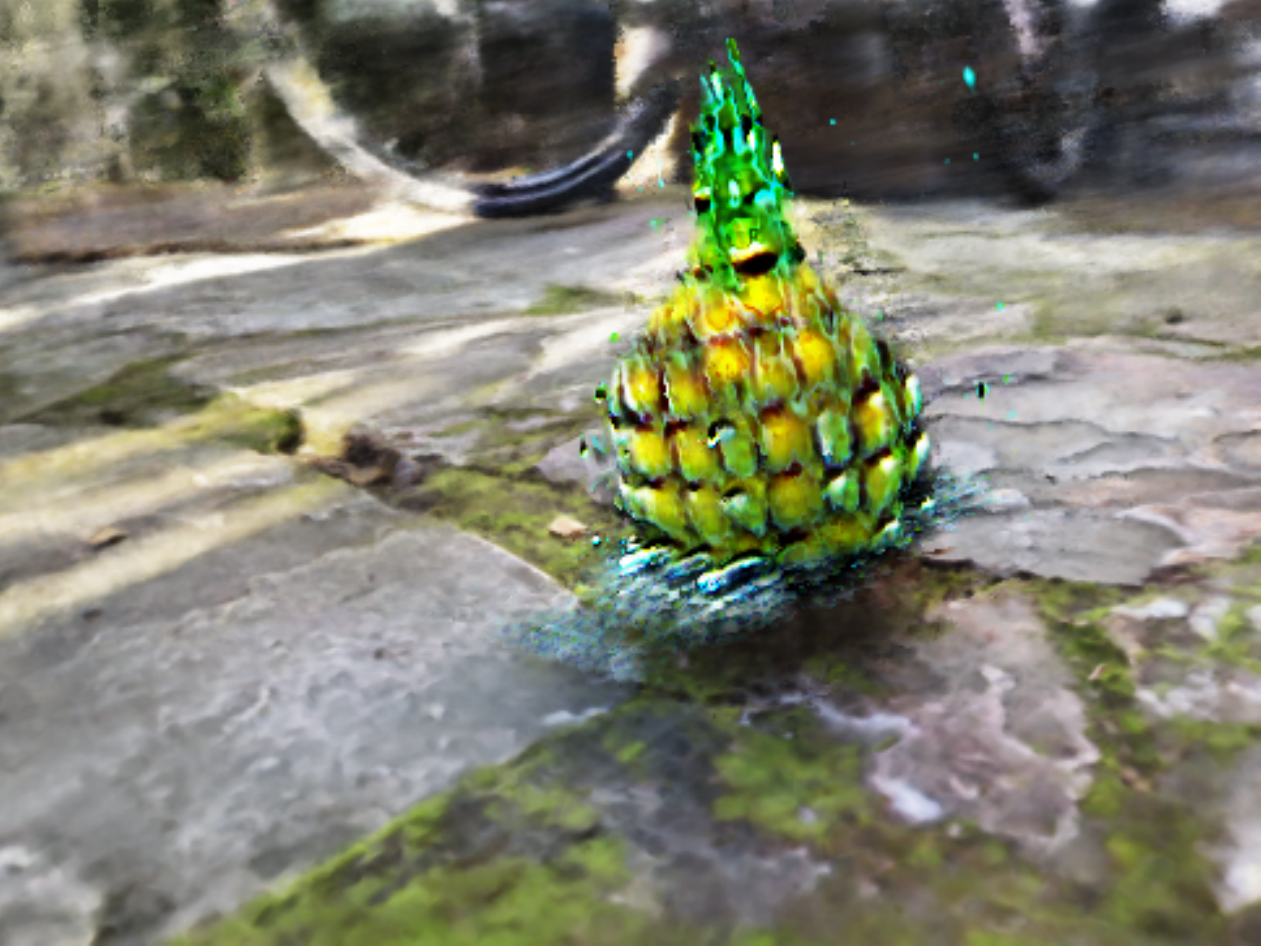}
    \hspace{0.001\textwidth}

    \includegraphics[width=\ww,frame]{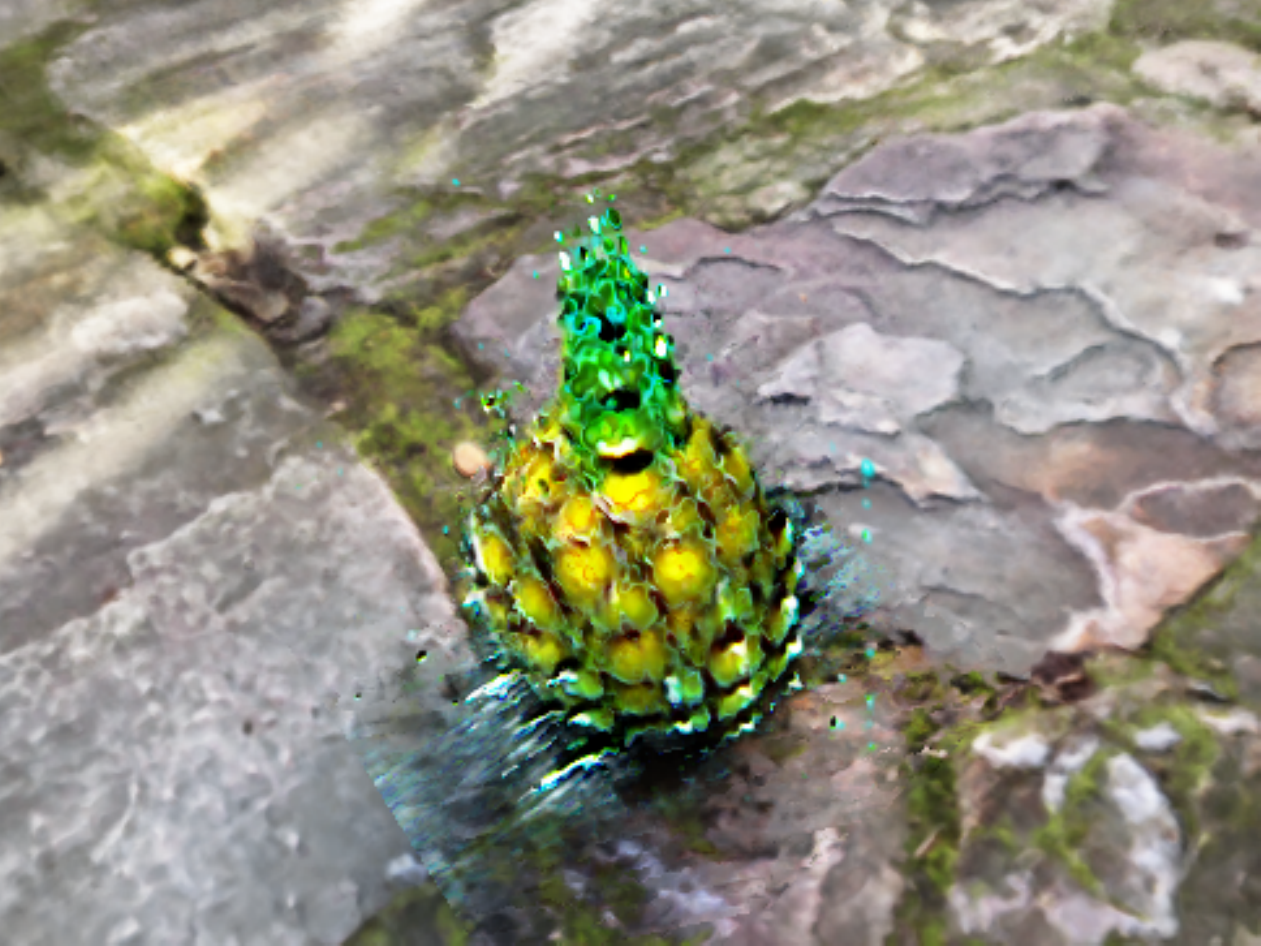}
    }  \label{fig:pineaple_edited}
    \vspace{0.0001\paperheight}

    \caption{ \textbf{Object replacement in 360 pinecone scene.} We replace the original pinecone object with pineapple using our proposed object replacement method.}  \label{fig:pineapple}
\end{figure*}

\begin{figure*}[h]
    \centering
    \setlength{\tabcolsep}{0.5pt}
    \renewcommand{\arraystretch}{0.5}
    \setlength{\ww}{0.40\columnwidth}
    \begin{tabular}{ccccc}        
        \includegraphics[width=\ww,frame]{figures/pinecone_texture/original_pinecone_009.pdf} &
        \includegraphics[width=\ww,frame]{figures/pinecone_texture/burning_pinecone_009.pdf} &
        \includegraphics[width=\ww,frame]{figures/pinecone_texture/ice_pinecone_009.pdf} &
        \includegraphics[width=\ww,frame]{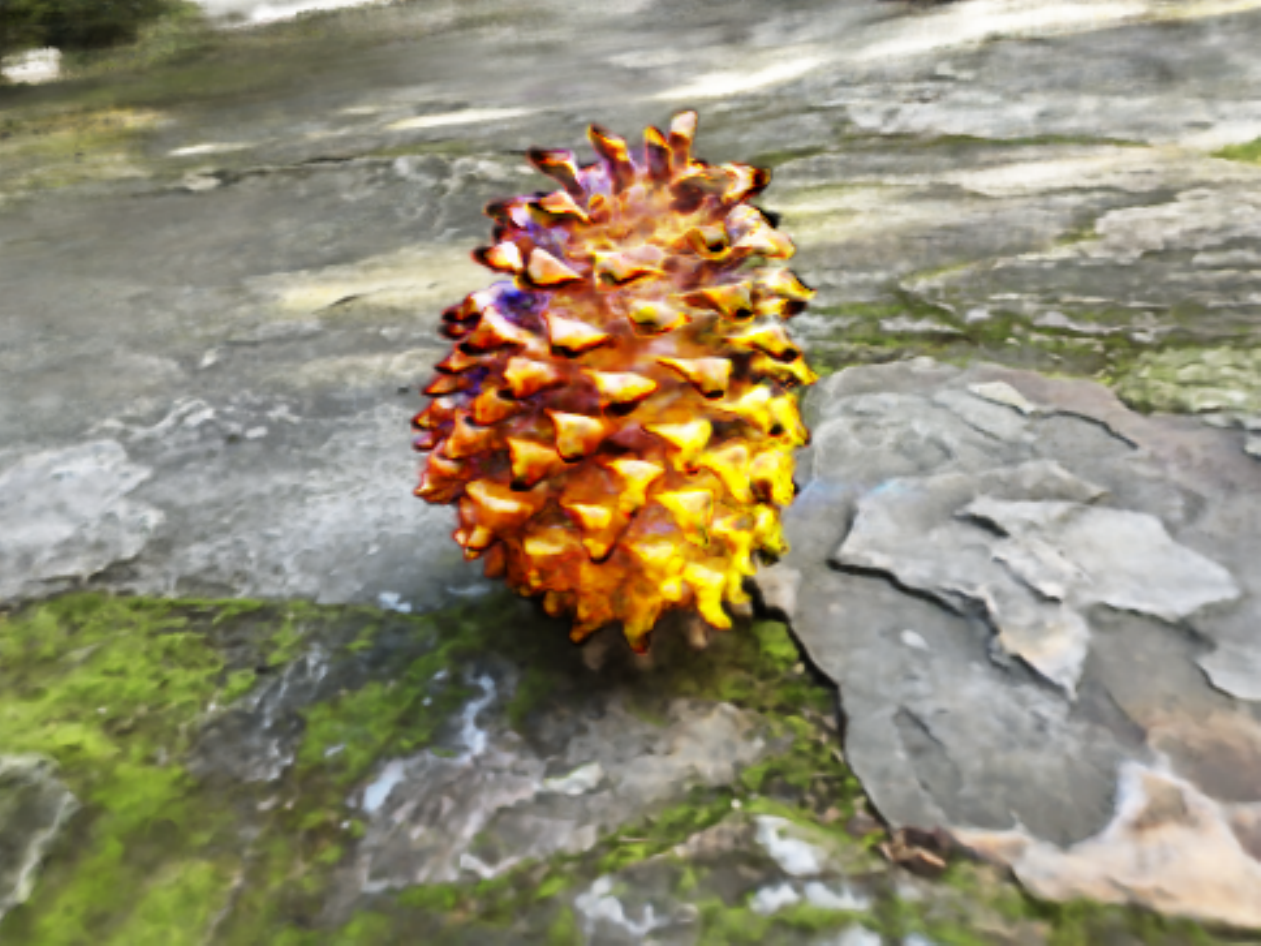} &
        \includegraphics[width=\ww,frame]{figures/pinecone_texture/wool_pinecone_009.pdf} \\
        
        \includegraphics[width=\ww,frame]{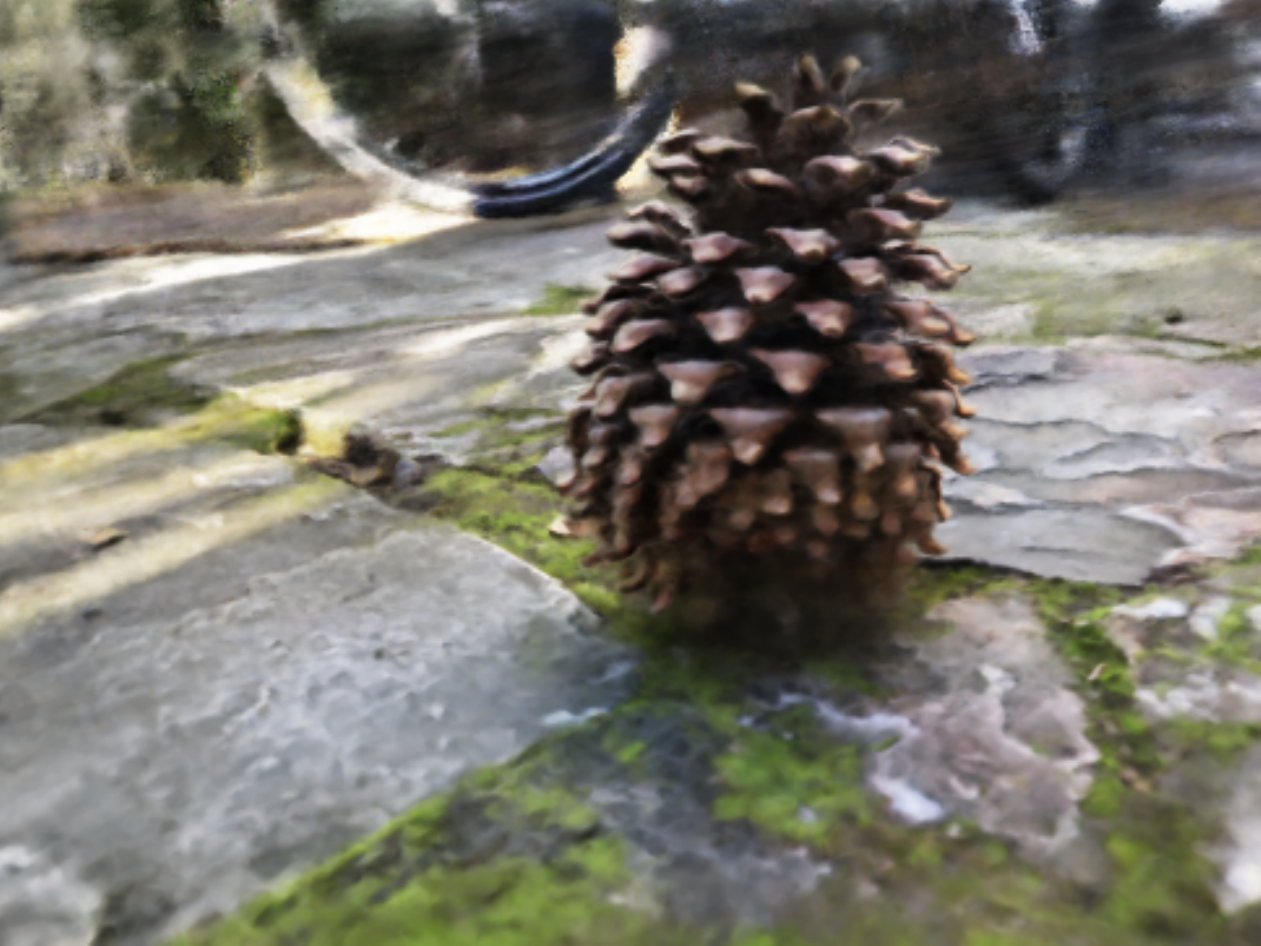} &
        \includegraphics[width=\ww,frame]{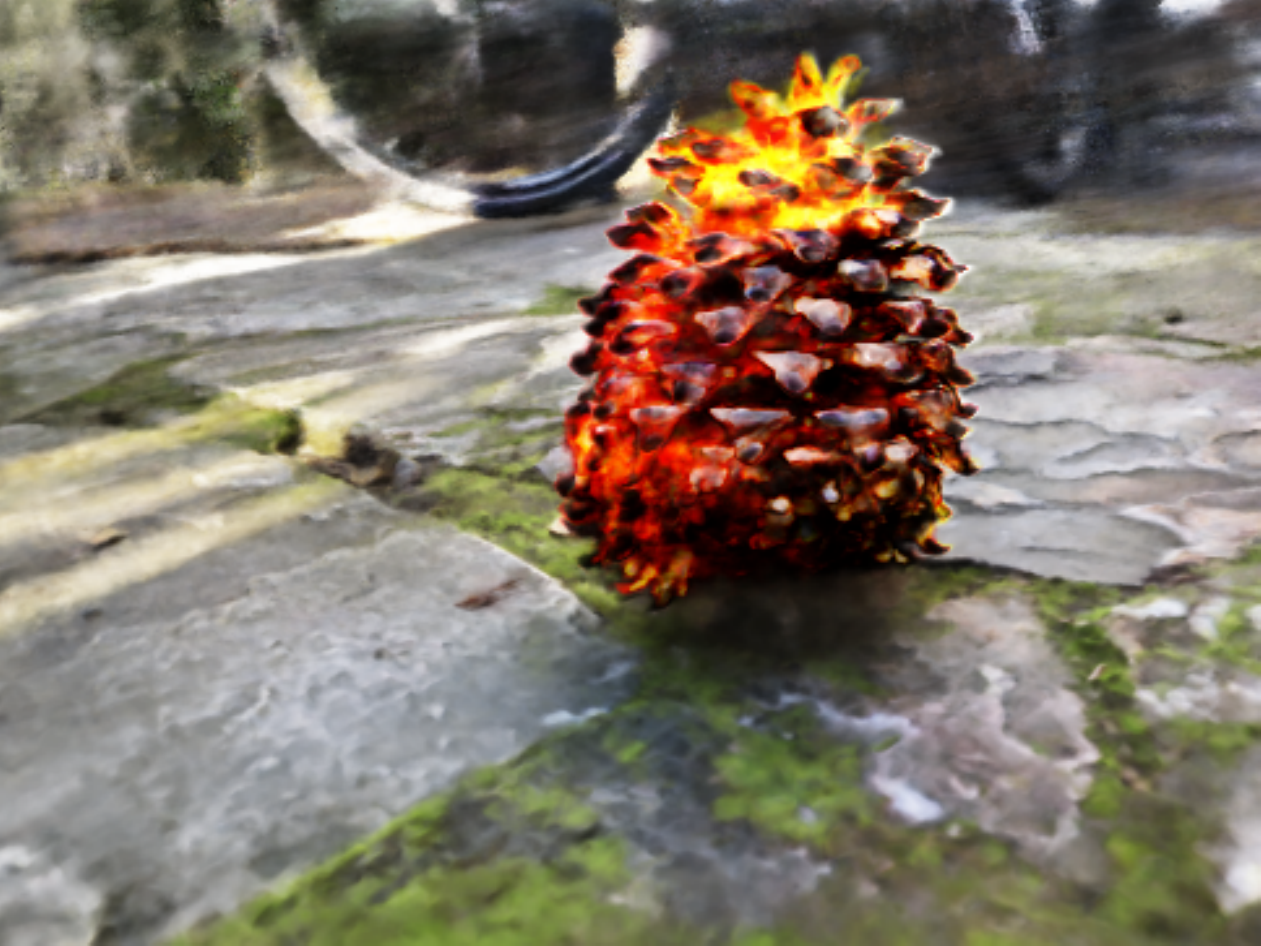} &
        \includegraphics[width=\ww,frame]{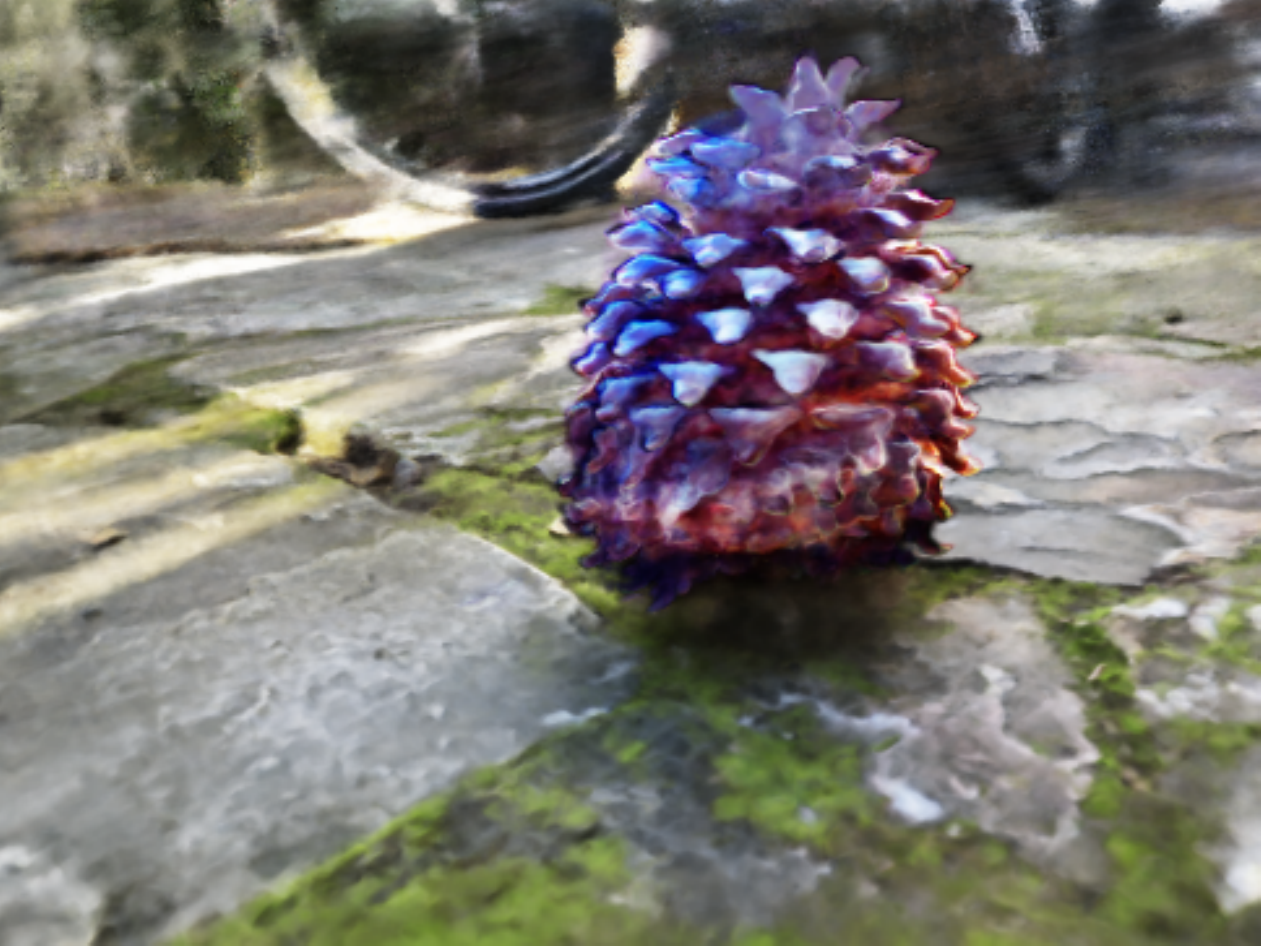} &
        \includegraphics[width=\ww,frame]{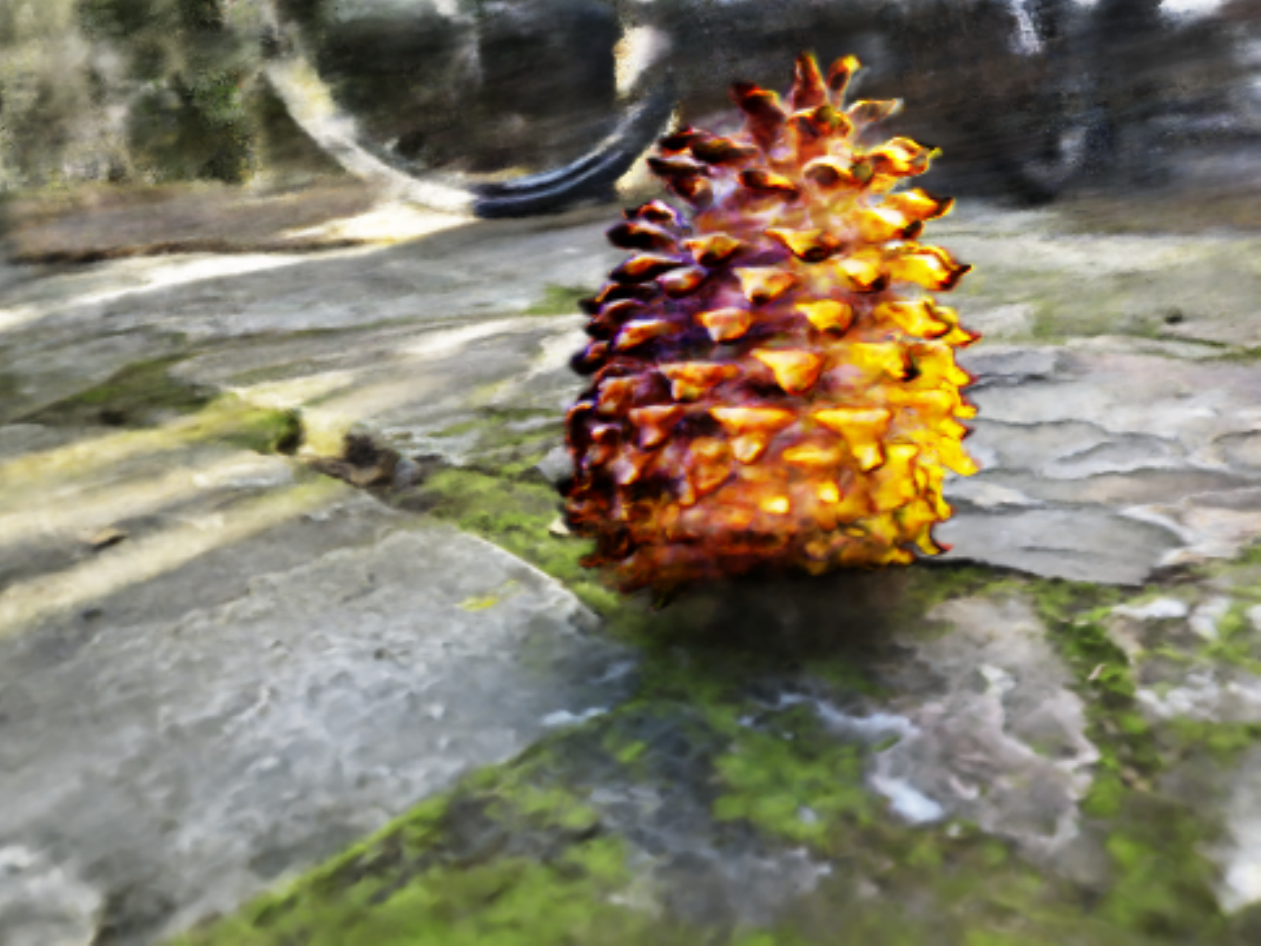} &
        \includegraphics[width=\ww,frame]{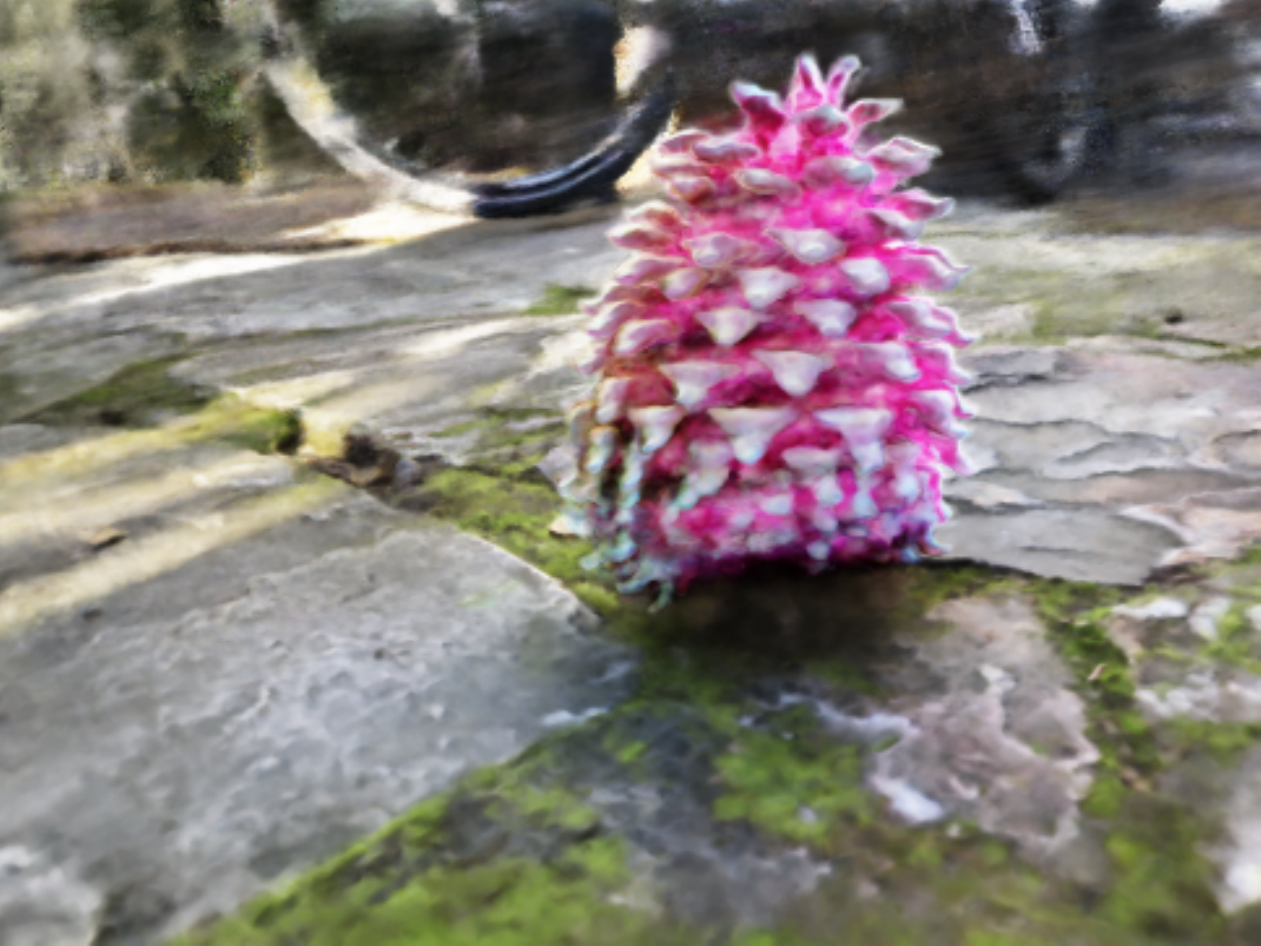} \\
		
		\scriptsize{Original Scene.} & 
		\scriptsize{"Burning pinecone".} & 
		\scriptsize{"Iced pinecone".} &
            \scriptsize{"Golden pinecone".} &
            \scriptsize{"Pinecone made of pink wool".} \\
    \end{tabular}
    
    \caption{\textbf{Texture conversion on 360 pinecone scene.}} \label{fig:pinecone_texture_sup}
\end{figure*}

\begin{figure*}[h] 
    \centering
    \setlength{\ww}{0.45\columnwidth}
    \subfloat[original Scene.]
    {
    \includegraphics[width=\ww]{figures/vasedeck_texture/vasedesk_originl_view0.pdf}
    \hspace{0.0001\textwidth}
    \includegraphics[width=\ww]{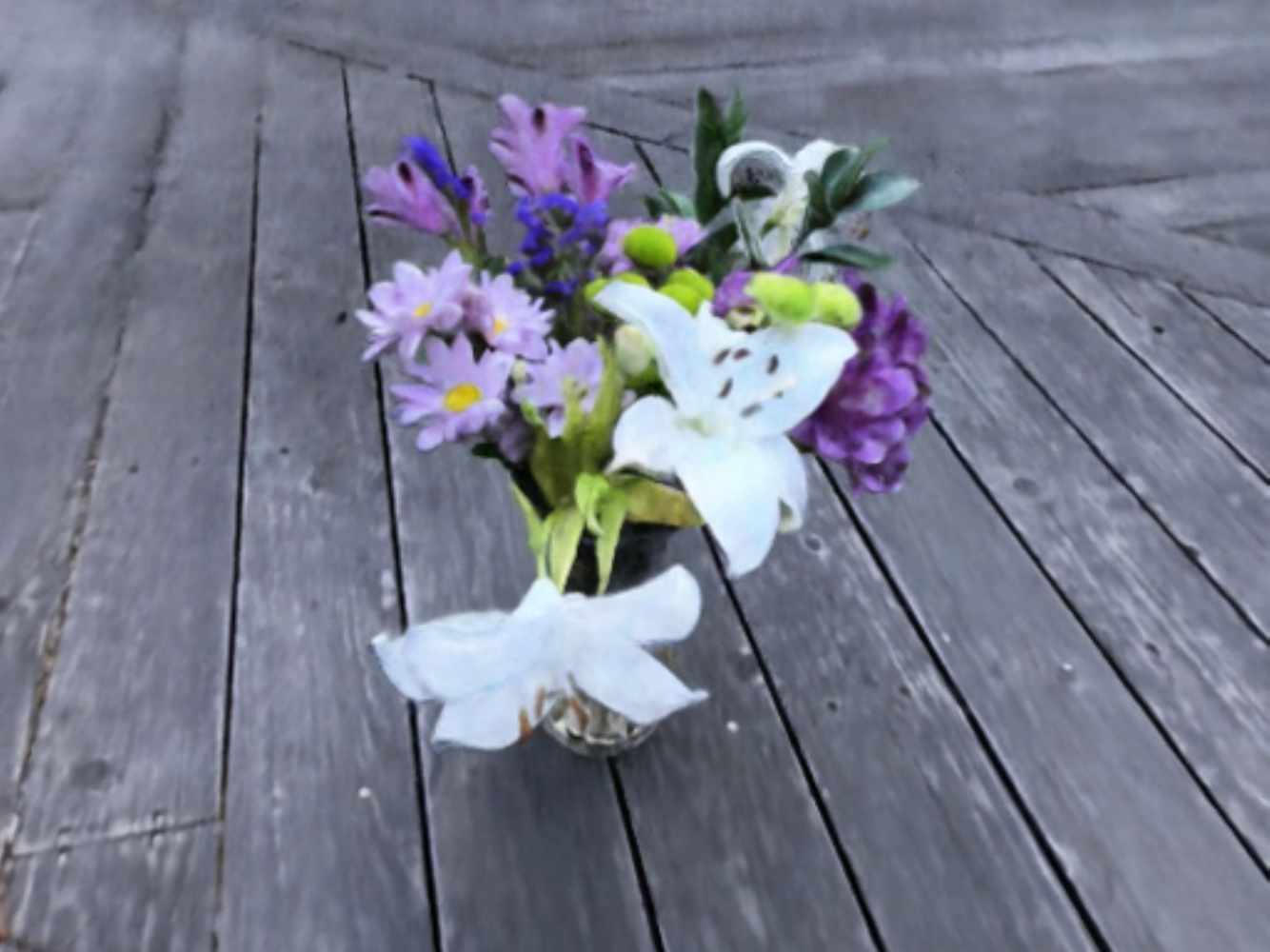}
    \hspace{0.0001\textwidth}
    \includegraphics[width=\ww]{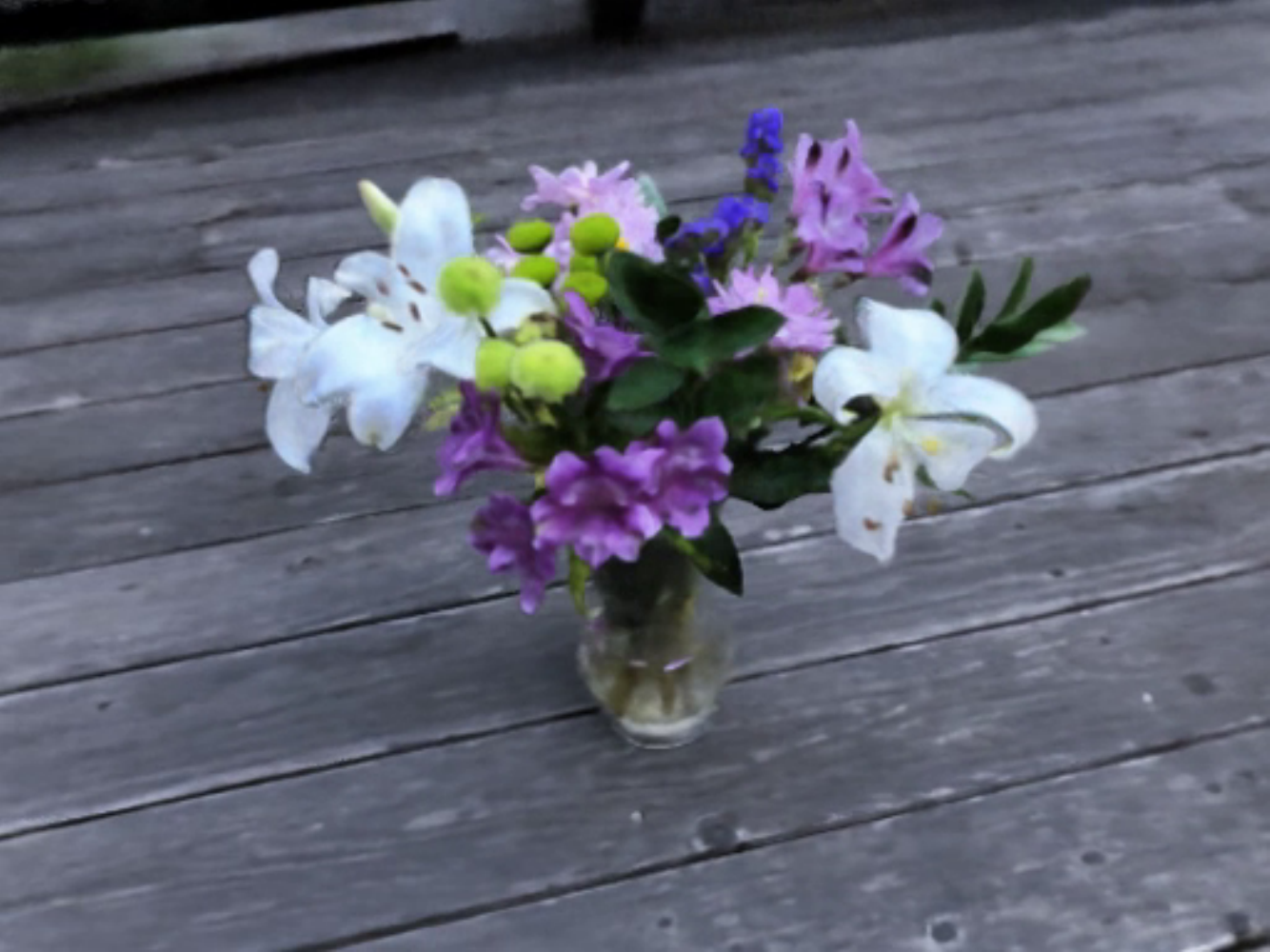}
    \hspace{0.0001\textwidth}
    \includegraphics[width=\ww]{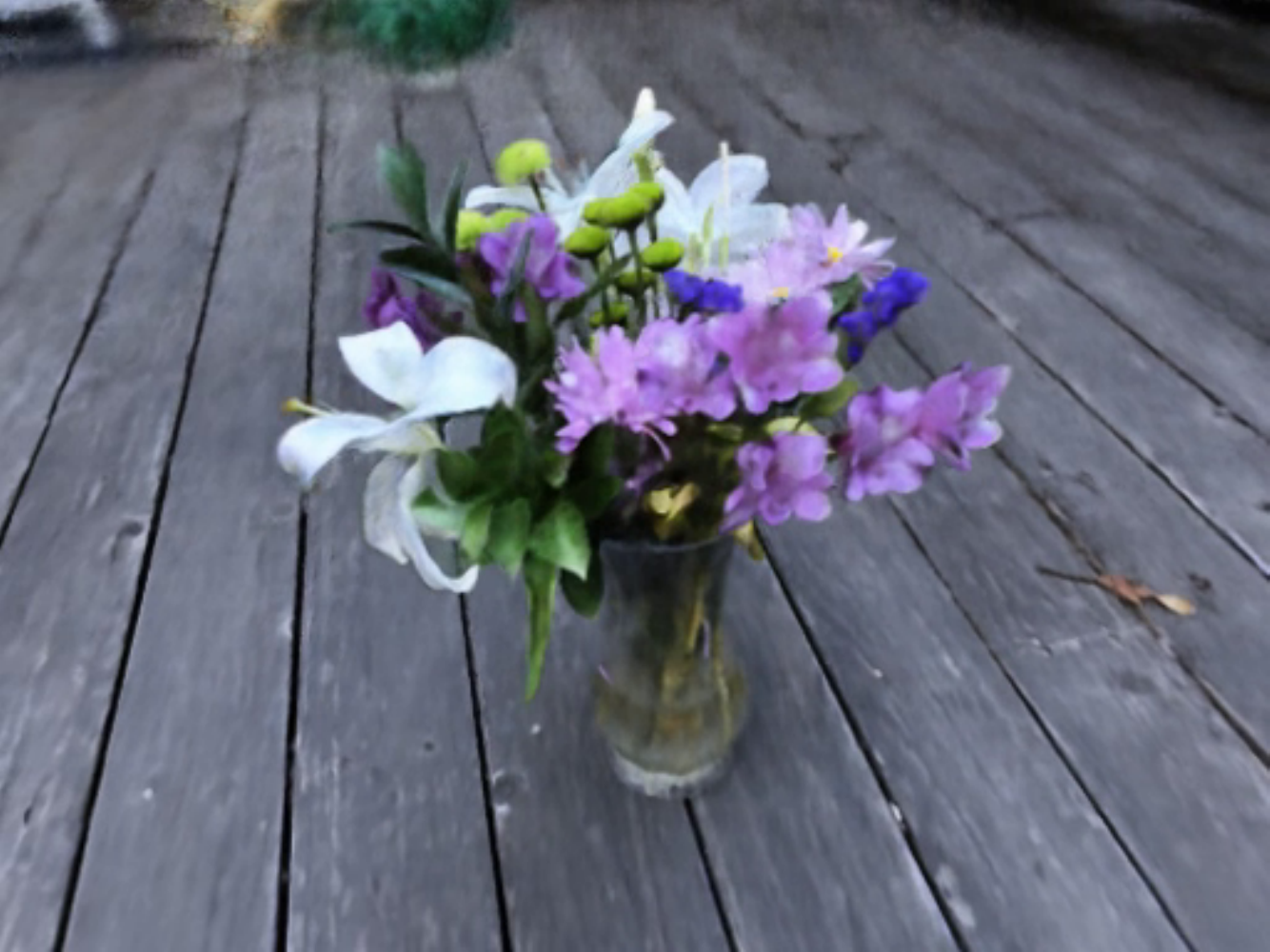}
    } 
    \vspace{0.0001\paperheight}

    \subfloat[``a vase made of glass.'']
    {
    \includegraphics[width=\ww,frame]{figures/vasedeck_texture/vasedesk_glass_edited_view0.pdf}
    \hspace{0.0001\textwidth}
    \includegraphics[width=\ww,frame]{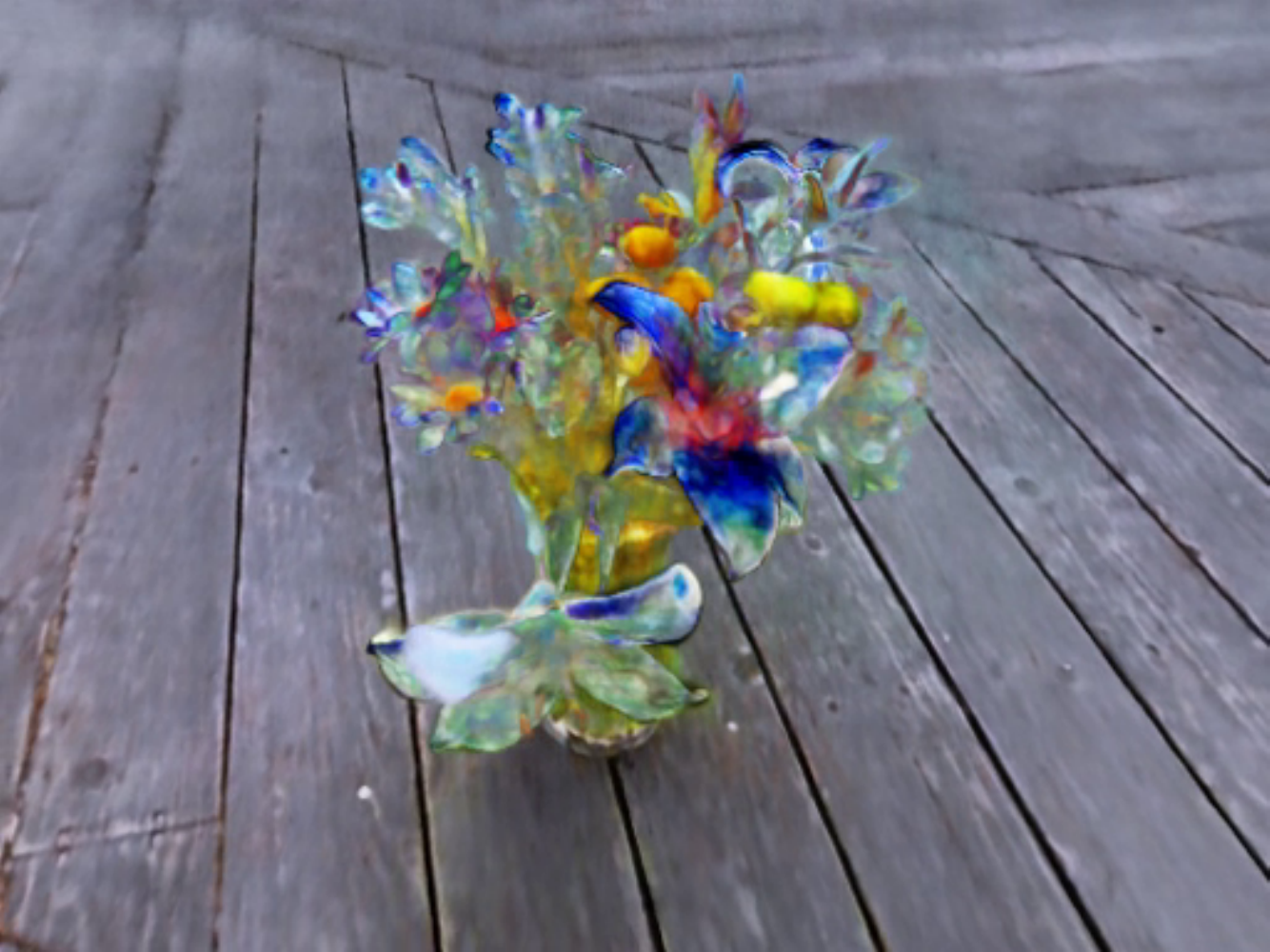}
    \hspace{0.0001\textwidth}
    \includegraphics[width=\ww,frame]{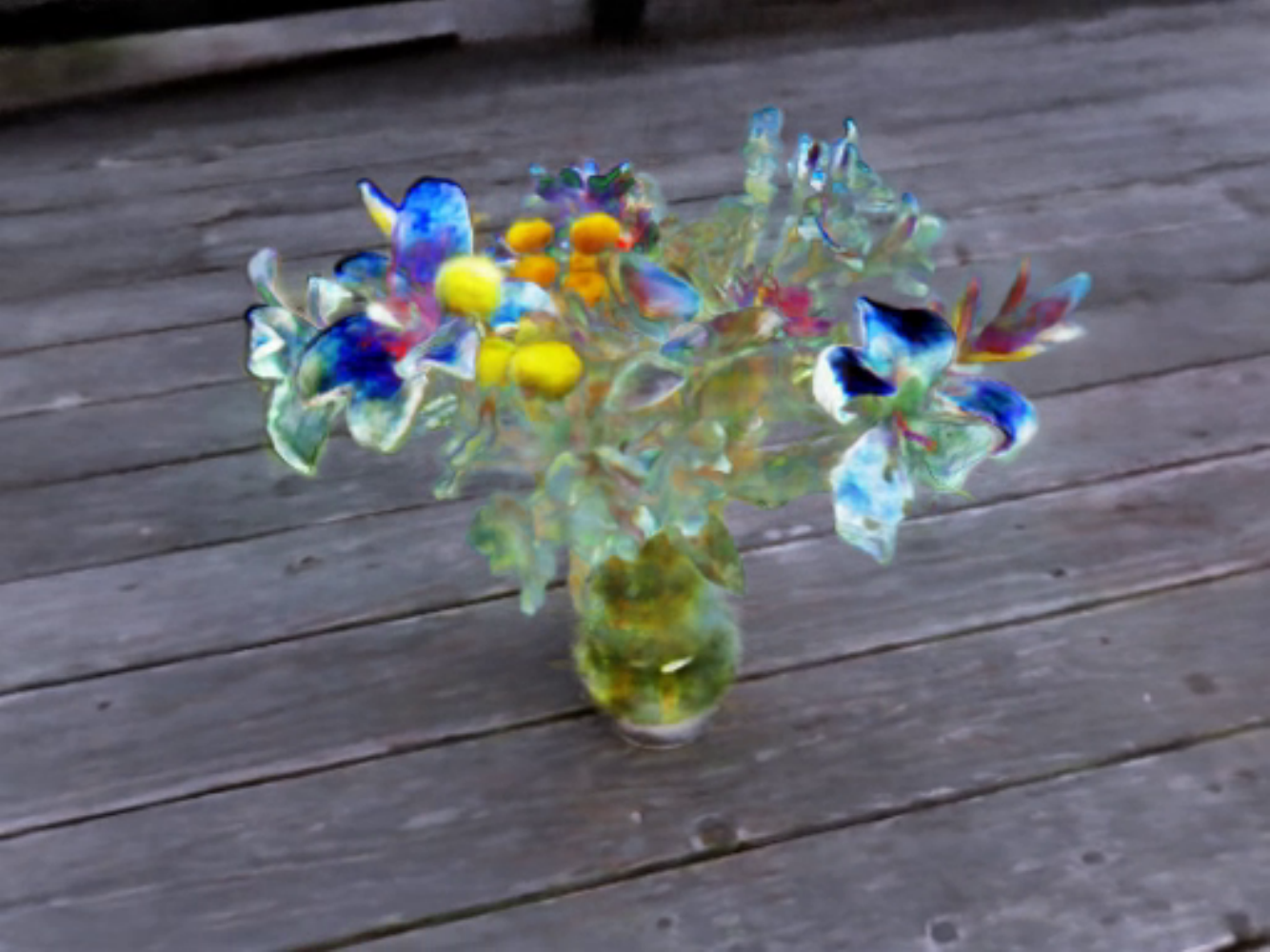}
    \hspace{0.0001\textwidth}
    \includegraphics[width=\ww,frame]{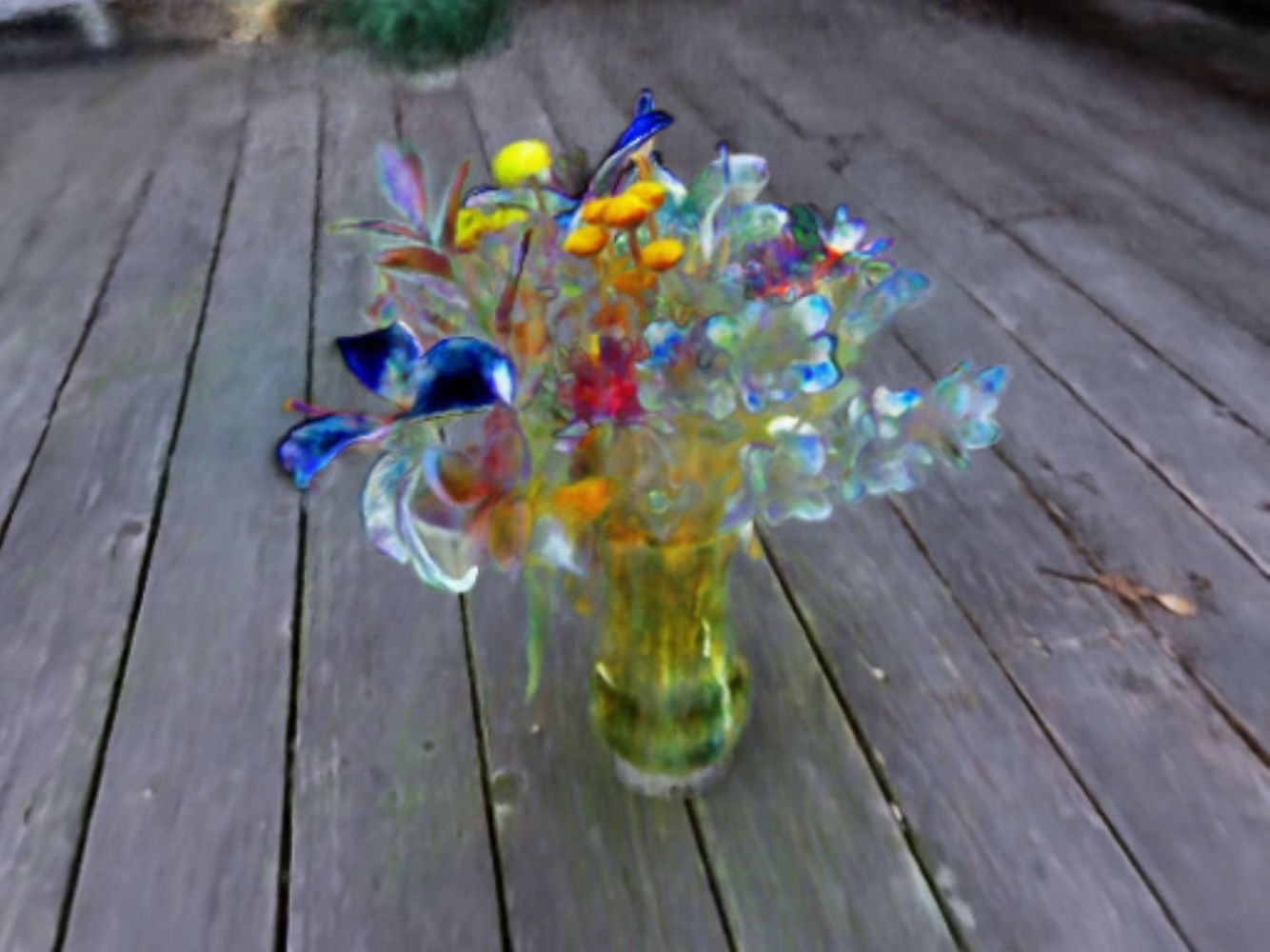}
    } 
    \vspace{0.0001\paperheight}

    \subfloat[``a vase made of stone.'']    
    {
    \includegraphics[width=\ww,frame]{figures/vasedeck_texture/vasedesk_stone_edited_view0.pdf}
    \hspace{0.0001\textwidth}
    \includegraphics[width=\ww,frame]{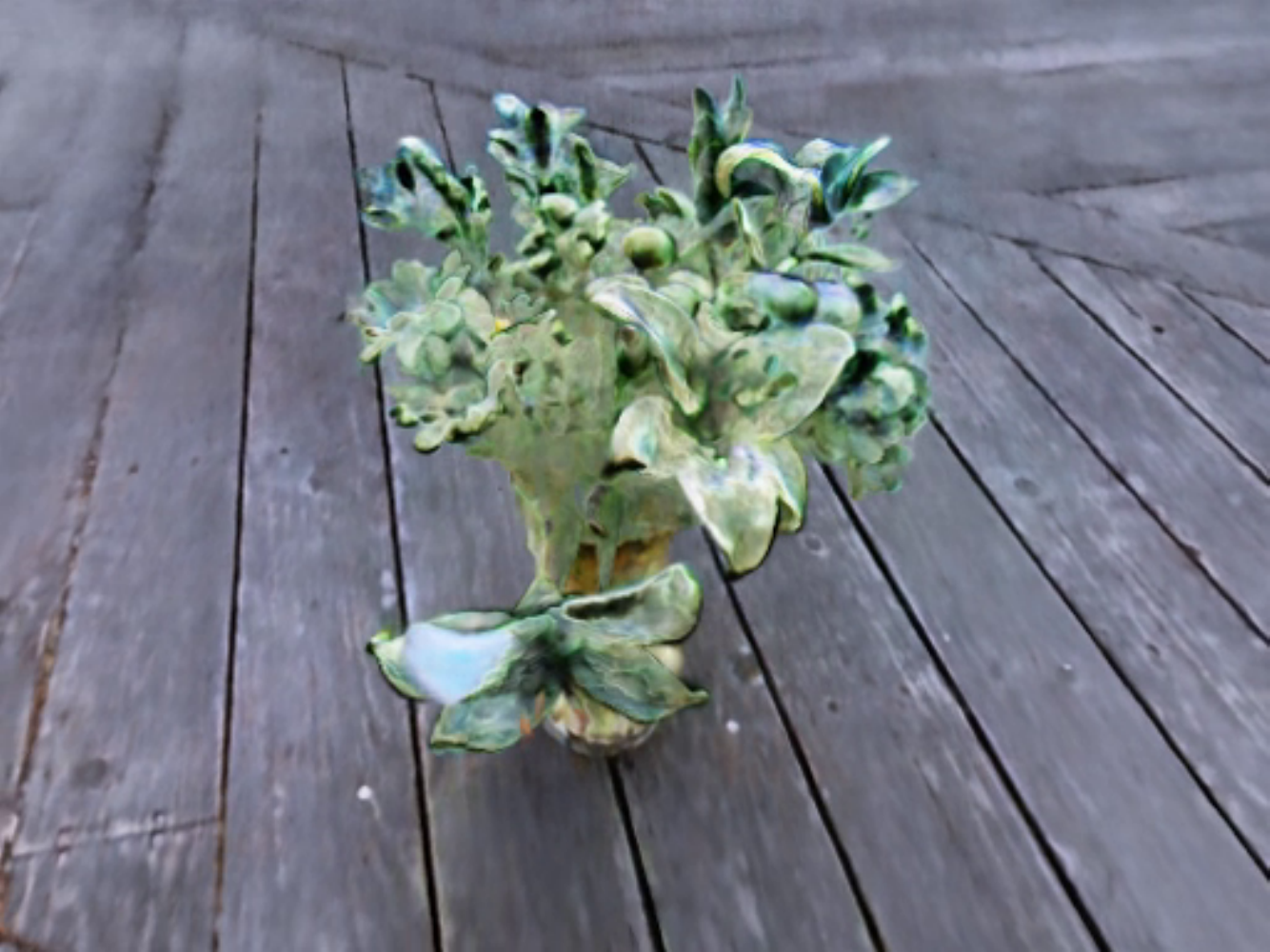}
    \hspace{0.0001\textwidth}
    \includegraphics[width=\ww,frame]{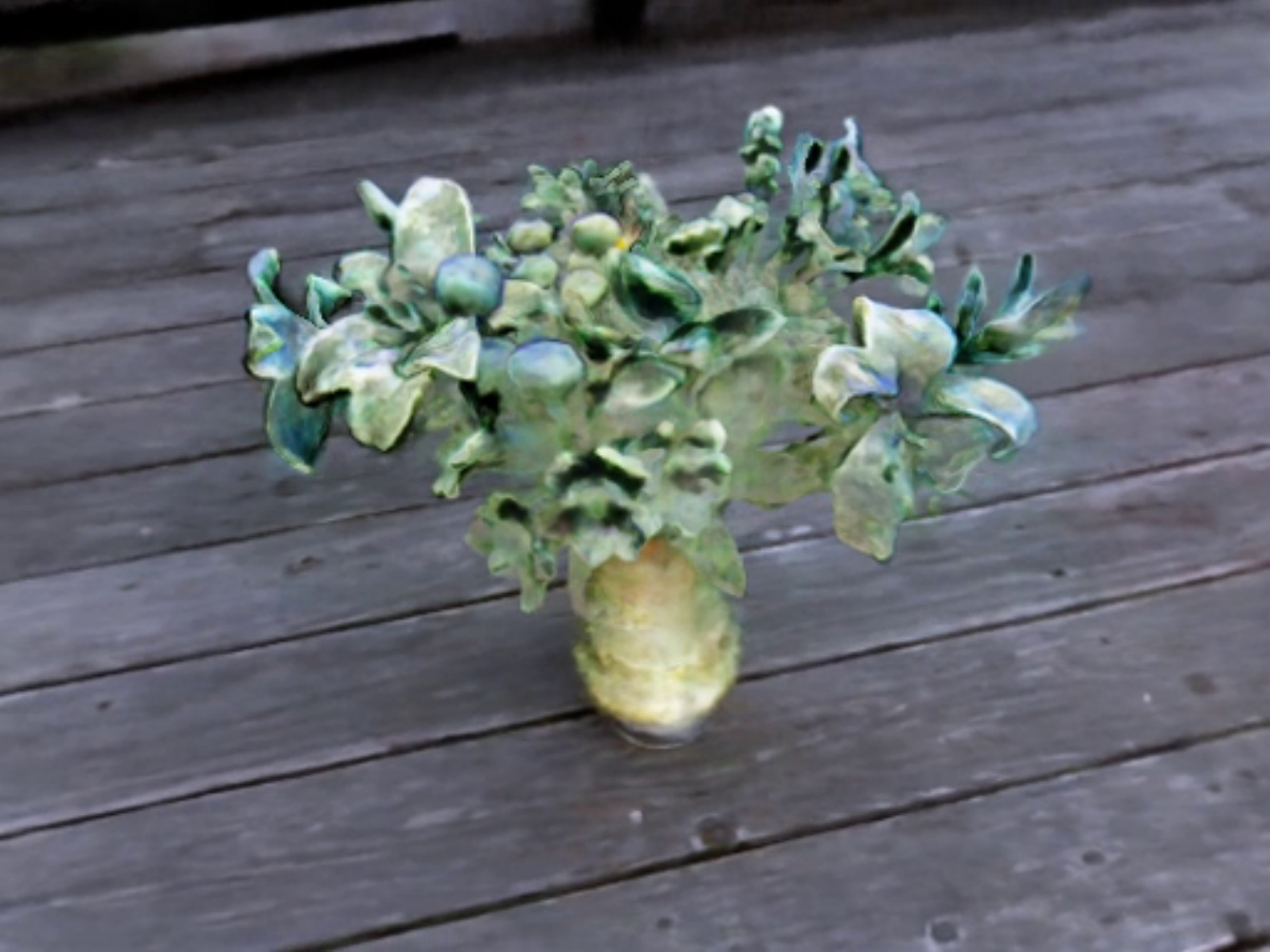}
    \hspace{0.0001\textwidth}
    \includegraphics[width=\ww,frame]{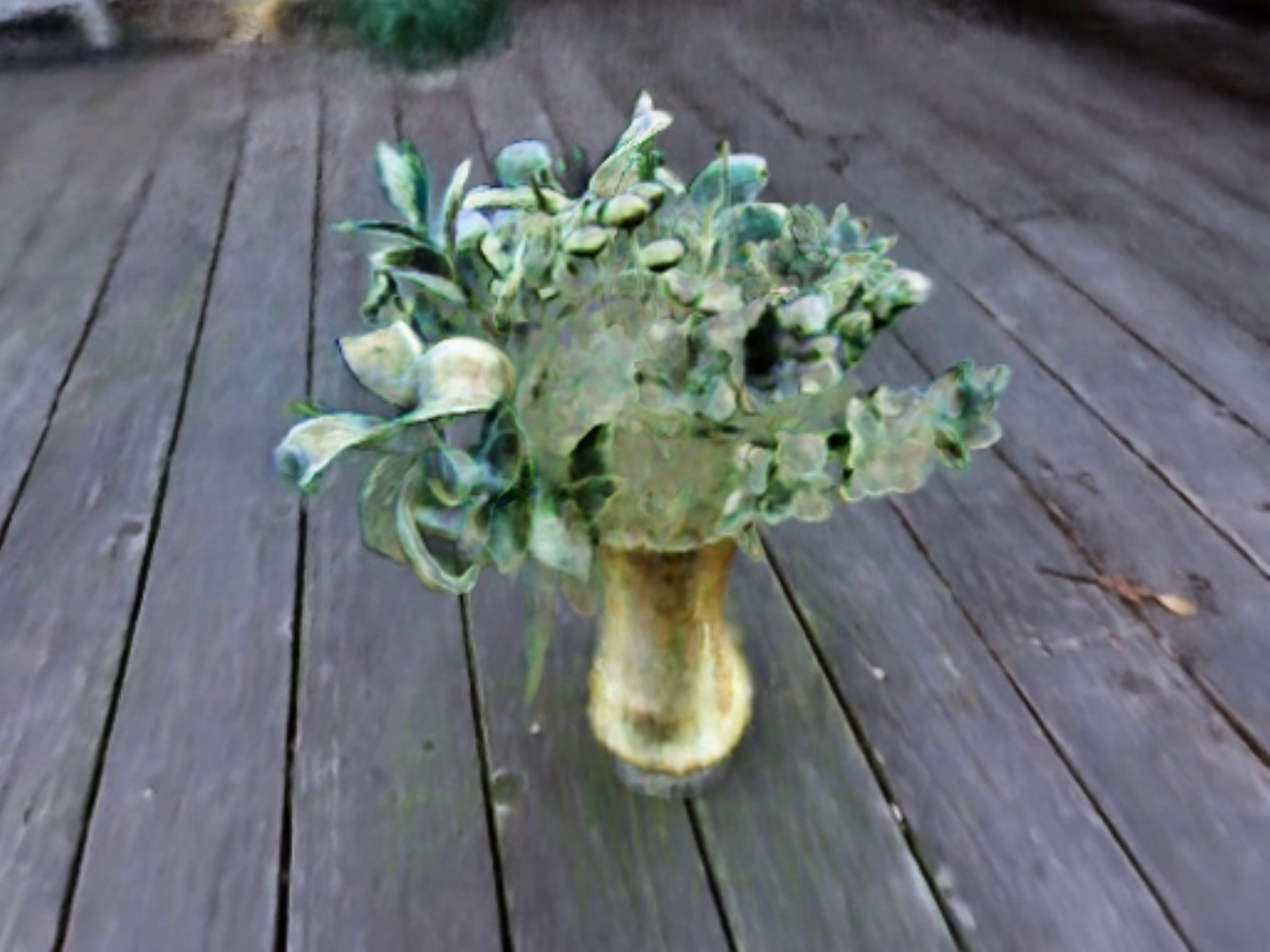}
    } 
    \vspace{0.0001\paperheight}

    \subfloat[``a water paint of a vase with flowers.'']    
    {
    \includegraphics[width=\ww,frame]{figures/vasedeck_texture/vasedesk_water_paint_view0.pdf}
    \hspace{0.0001\textwidth}
    \includegraphics[width=\ww,frame]{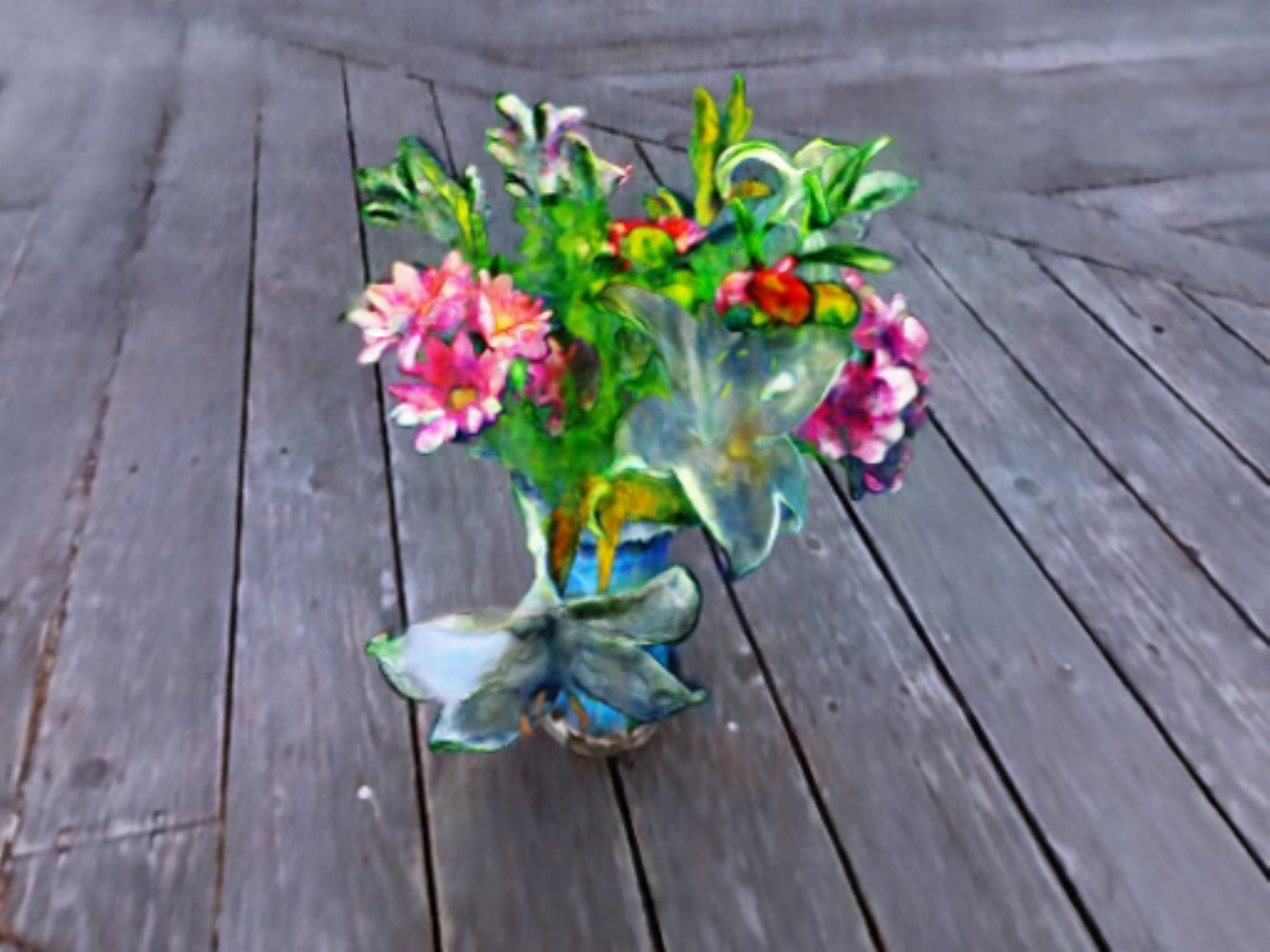}
    \hspace{0.0001\textwidth}
    \includegraphics[width=\ww,frame]{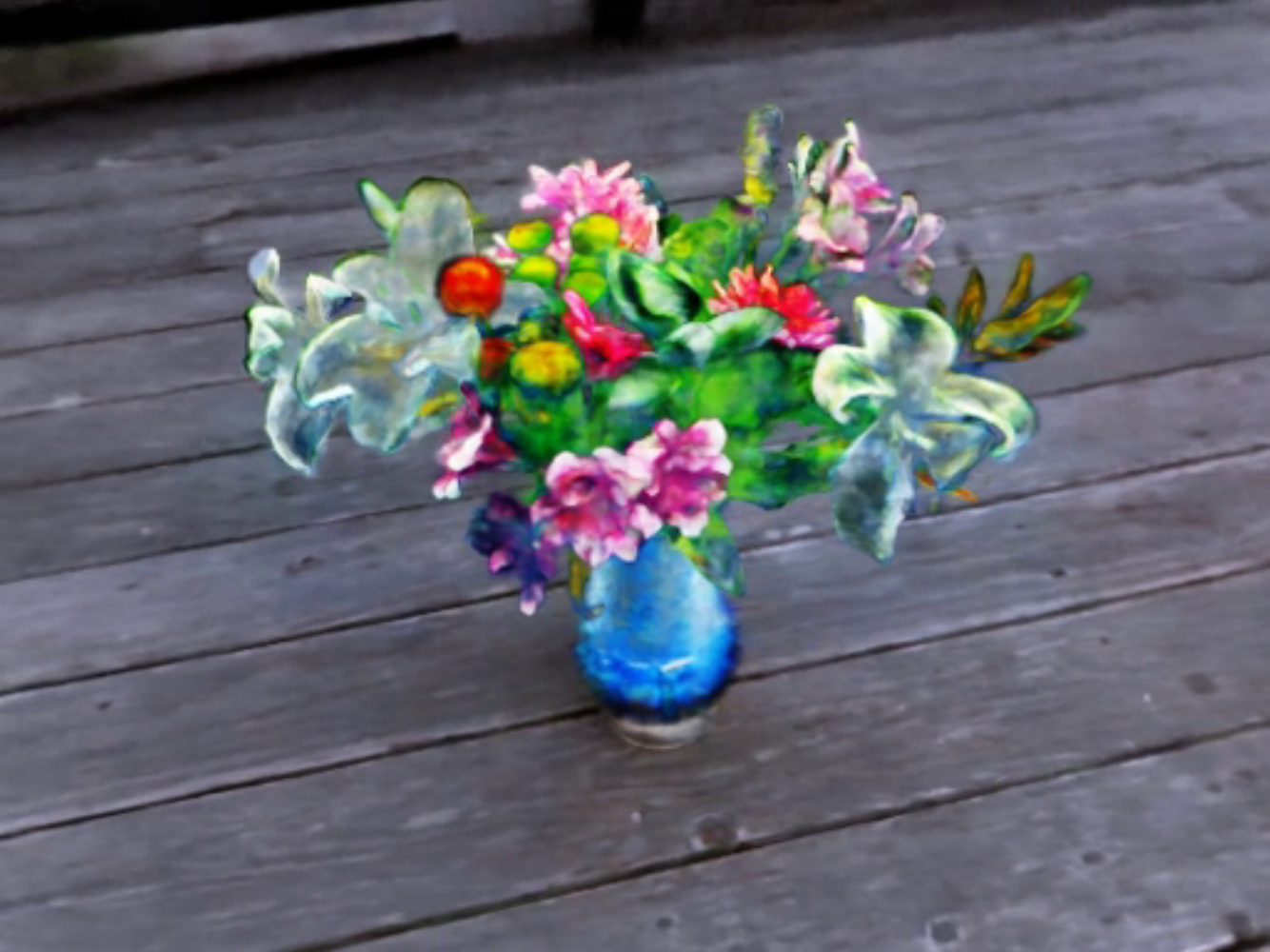}
    \hspace{0.0001\textwidth}
    \includegraphics[width=\ww,frame]{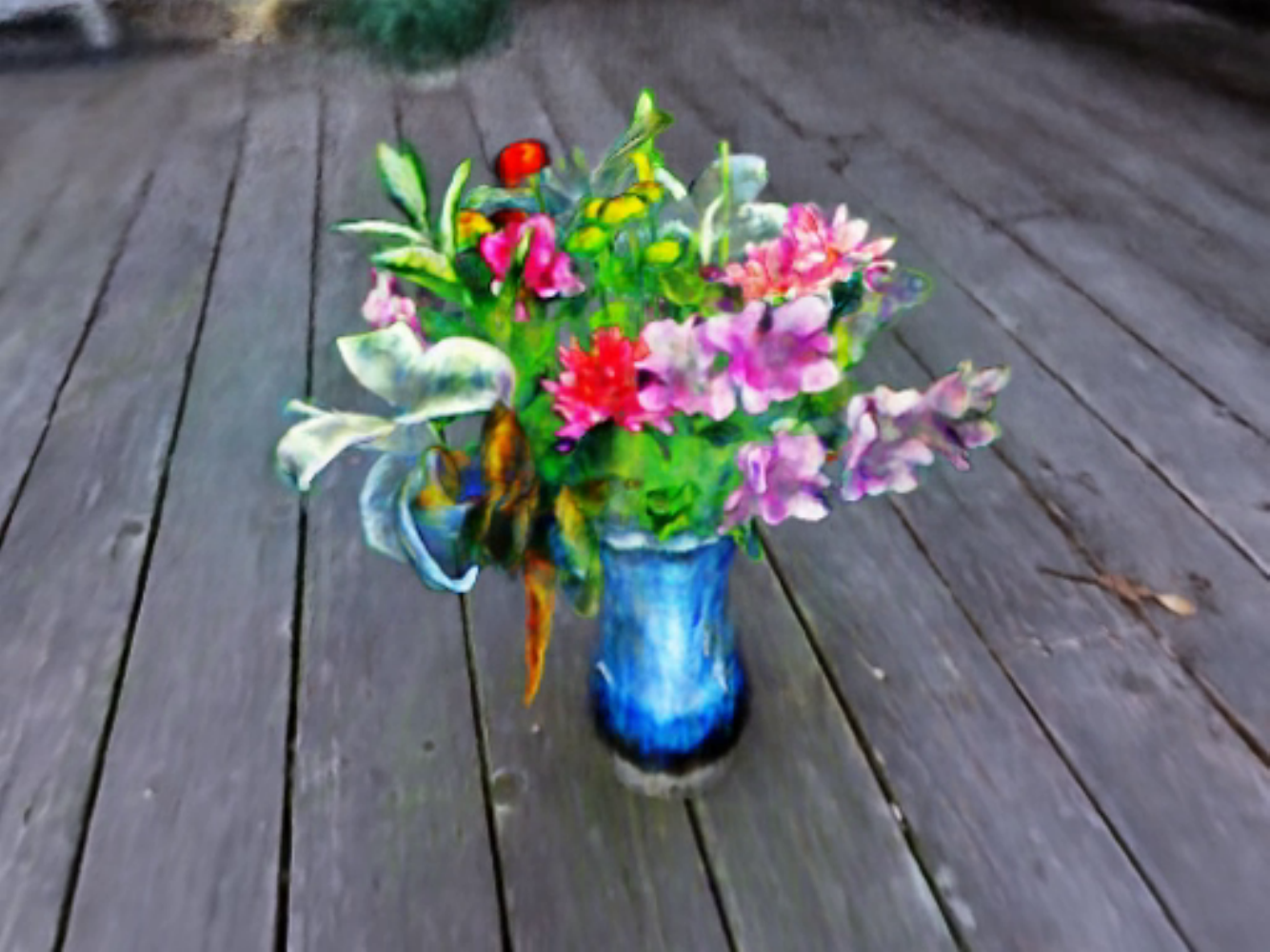}
    } 
    \vspace{0.0001\paperheight}

    \caption{ \textbf{Texture conversion on 360 vasedeck scene.}}  \label{fig:vasedeck_texture_sup}
\end{figure*}

We provide additional examples for the applications in the main paper. In \Cref{fig:volume_comp_views} we display additional views for the object replacement comparison with Volumetric Disentanglement for 3D Scene Manipulation \cite{VolumeDisentanglement2022}. In \Cref{fig:coco_object_insertion} we demonstrate new object insertion using several
captions from COCO dataset \cite{lin2014microsoftcoco}. In \Cref{fig:vasedeck_flower_petals} and \Cref{fig:pineapple} we show more examples for object replacement, and in \Cref{fig:pinecone_texture_sup} and \Cref{fig:vasedeck_texture_sup} we display more edits and views for texture conversion task on 360 scenes.

\end{document}